\pdfoutput=1
\documentclass[11pt]{article}

\usepackage{acl}

\usepackage{latexsym}
\usepackage{times}

\usepackage[T1]{fontenc}
\usepackage[utf8]{inputenc}
\usepackage{microtype}
\usepackage{url}

\usepackage{graphicx}
\usepackage{amsmath}
\usepackage{amsthm}
\usepackage{amssymb}
\usepackage{algorithm}
\usepackage{algorithmic}
\usepackage[tight,footnotesize]{subfigure}
\usepackage{bm}
\usepackage{caption}
\usepackage{enumitem}
\usepackage{natbib}
\usepackage{multirow}
\usepackage{color}
\usepackage{booktabs}
\usepackage{nicefrac}
\usepackage[font = it]{caption}
\usepackage{stackengine}
\usepackage[normalem]{ulem}
\usepackage{listings}

\newtheorem{definition}{Definition}

\usepackage{array}
\newcolumntype{L}[1]{>{\raggedright\let\newline\\\arraybackslash\hspace{0pt}}m{#1}}
\newcolumntype{C}[1]{>{\centering\let\newline  \\\arraybackslash\hspace{0pt}}m{#1}}
\newcolumntype{R}[1]{>{\raggedleft\let\newline \\\arraybackslash\hspace{0pt}}m{#1}}

\title{KGTuner: Efficient Hyper-parameter Search for \\ Knowledge Graph Learning}

\author{Yongqi Zhang$^1$ \quad  Zhanke Zhou$^2$ \quad Quanming Yao$^3$ \quad Yong Li$^3$ \\
	$^1$4Paradigm Inc., Beijing, China \\
	$^2$Hong Kong Baptist University, Hong Kong, China \\
	$^3$Department of Electronic Engineering, Tsinghua University, Beijing, China \\
	\texttt{\small zhangyongqi@4paradigm.com,}
	\texttt{\small cszkzhou@comp.hkbu.edu.hk,}
	\texttt{\small {qyaoaa/liyong07}@tsinghua.edu.cn}}

\begin{document}

\maketitle

\begin{abstract}
While
hyper-parameters (HPs) 
are important for knowledge graph (KG) learning,
existing methods 
fail to search them efficiently.
To solve this problem,
we first
analyze the properties of different HPs
and measure the transfer ability from small subgraph to the full graph.
Based on the analysis,
we propose an efficient two-stage search algorithm KGTuner,
which efficiently explores HP configurations
on small subgraph at the first stage
and transfers the top-performed configurations for fine-tuning on the large full graph
at the second stage.
Experiments 
show that
our method 
can consistently find better HPs than the baseline algorithms 
within the same time budget,
which achieves {9.1\%} average relative improvement
for four embedding models
on the large-scale KGs 
in open graph benchmark.
Our code is released in 
\url{https://github.com/AutoML-Research/KGTuner}.
\footnote{The work is performed when
	Z. Zhou was an intern in 4Paradigm,
	and correspondence is to Q. Yao.
	}
\end{abstract}

\section{Introduction}
\label{sec:intro}

\begin{figure*}[ht]
	\centering
	\vspace{-7px}
	\setlength{\abovecaptionskip}{-2pt}
	\subfigure[Conventional methods]{\includegraphics[height=3.6cm]{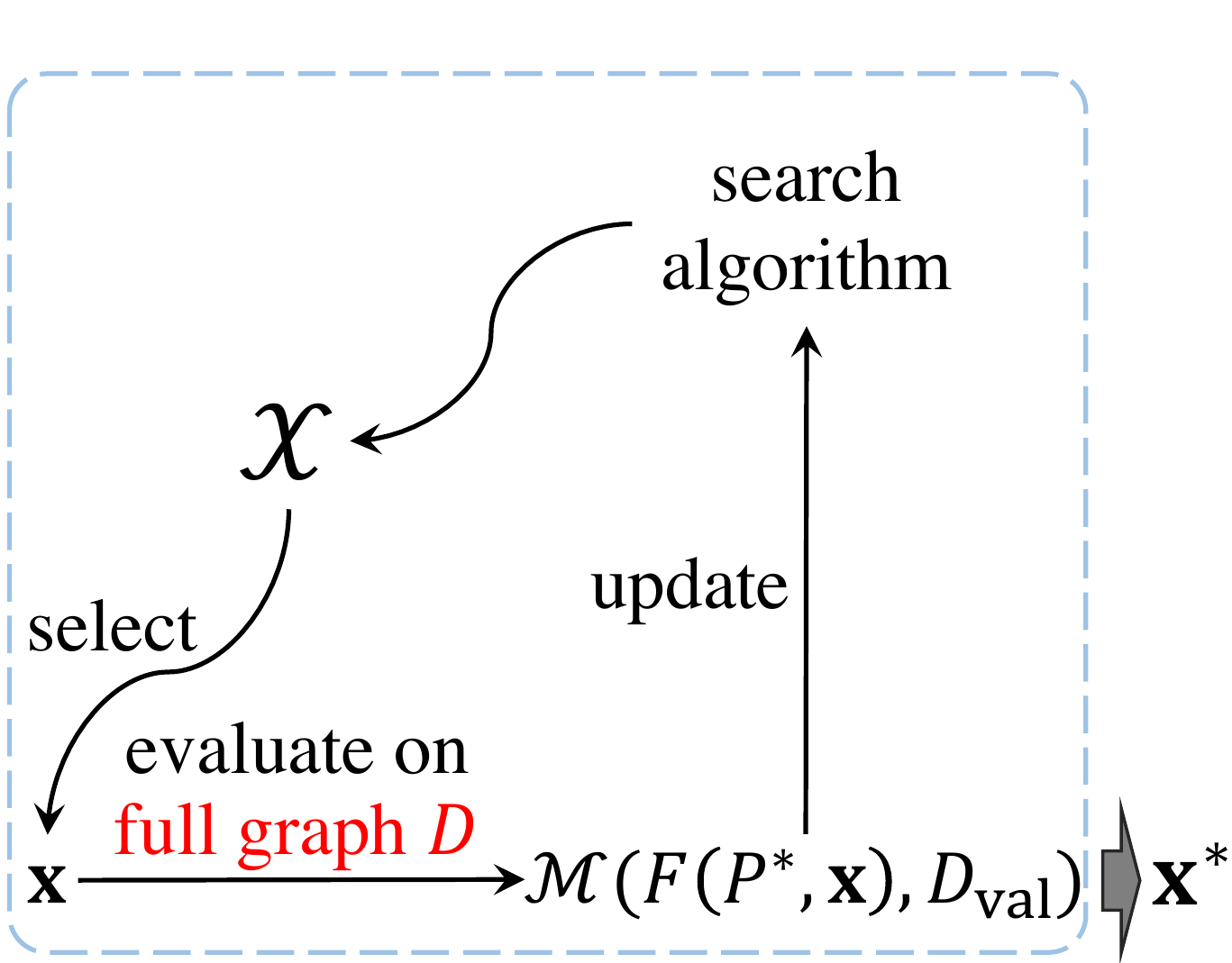} \label{fig:conventional}}
	\hfill
	\subfigure[KGTuner]{\includegraphics[height=3.6cm]{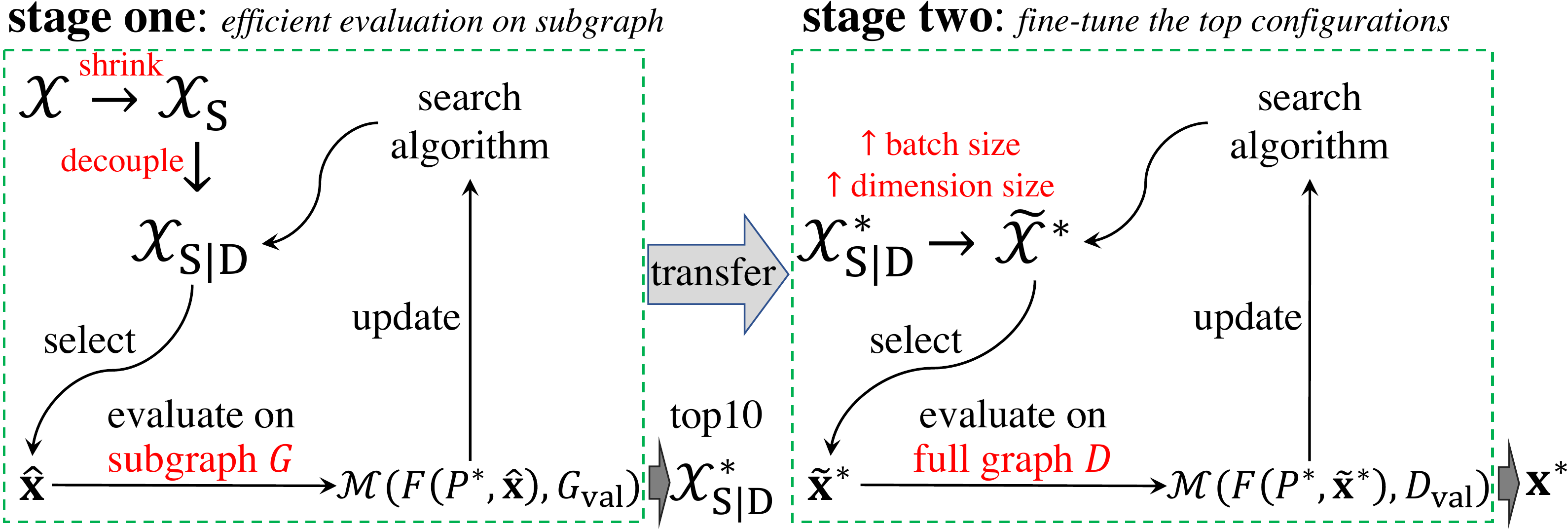} \label{fig:KGTuner}}
	\caption{The framework of conventional HP search algorithm and the proposed KGTuner.}
	\vspace{-12px}
\end{figure*}

Knowledge graph (KG) is a special kind of graph structured data
to represent knowledge 
through entities and  
relations between the entities
\citep{wang2017knowledge,ji2020survey}.
Learning from KG aims to discover the latent properties from KGs
to infer the existence of interactions among entities 
or the types of entities \cite{wang2017knowledge,zhang2022knowledge}.
KG 
embedding,
which encodes entities and relations as low dimensional vectors,
is an important technique to learn from KGs~\citep{wang2017knowledge,ji2020survey}.
The existing models range from
translational distance models
\citep{bordes2013translating},
tensor factorization models 
\citep{nickel2011three,trouillon2017knowledge,balavzevic2019tucker},
neural network models 
\citep{dettmers2017convolutional,guo2019learning},
to graph neural networks \citep{schlichtkrull2018modeling,vashishth2019composition}.

Hyper-parameter (HP) search~\citep{claesen2015hyperparameter}
is very essential for KG learning.
In this work,
we take KG embedding methods \cite{wang2017knowledge},
as a good example to study the impact of HPs
to KG learning.
As studied,
the HP configurations greatly
influence the model performance~\citep{ruffinelli2019you,ali2020bringing}.
An improper HP configuration
will impede the model 
from stable convergence,
while an appropriate one
can make considerable promotion to the model performance.
Indeed, studying the HP configurations
can help us make a more scientific understanding of
the contributions made by existing works \citep{rossi2020knowledge,sun2020re}.
In addition, 
it is also important
to search for an optimal HP configuration
when adopting KG embedding methods to 
the real-world applications \citep{bordes2014question,zhang2016collaborative,saxena2020improving}.

Algorithms for HP search 
on general machine learning problems
have been well-developed \citep{claesen2015hyperparameter}.
As shown in Figure~\ref{fig:conventional},
the search algorithm selects a HP configuration
from the search space in each iteration,
then the evaluation feedback obtained by full model training
is used to update the search algorithm.
The optimal HP is the one achieving
the best performance on validation data in the search process.
Representative HP search algorithms
are within sample-based methods
like grid search, random search \citep{bergstra2012random},
and sequential model-based Bayesian optimization (SMBO) methods
like Hyperopt \citep{bergstra2013hyperopt},
SMAC \citep{hutter2011sequential},
Spearmint \citep{snoek2012practical}
as well as BORE \citep{tiao2021bore}, etc.
Recently,
some 
subgraph-based methods~\citep{tu2019autone,wang2021explainable}
are proposed to learn a predictor with configurations efficiently evaluated on small subgraphs
The predictor is then transferred
to guide HP search
on the full graph.
However,
these methods fail to efficiently search 
a good configuration of HPs for KG embedding models
since the training cost of individual model is high 
and the correlation of HPs in the huge search space is very complex.


To address the limitations of 
existing 
HP search algorithms,
we carry a comprehensive understanding study
on the influence 
and correlation of HPs
as well as 
their transfer ability 
from small subgraph to full graph
in KG learning.
From the aspect of performance,
we classify the HPs into four different groups
including \textit{reduced options},
\textit{shrunken range},
\textit{monotonously related} 
and \textit{no obvious patterns}
based on their influence on the performance.
By analyzing the validation curvature of these HPs,
we find that the space is rather complex
such that only tree-based models can 
approximate it well.
In addition,
we observe that the  consistency
between evaluation on small subgraph
and that on the full graph is high,
while the evaluation cost is significantly smaller on the small subgraph.

Above understanding
motivates us to 
reduce the size of search space
and
design a two-stage search algorithm
named as KGTuner.
As shown in Figure~\ref{fig:KGTuner},
KGTuner 
explores
HP configurations in the shrunken and decoupled space
with the search algorithm RF+BORE \citep{tiao2021bore}
on a subgraph 
in the first stage,
where the evaluation cost of HPs are small.
Then in the second stage,
the configurations achieving the \textit{top10} performance at the first stage
are equipped with large batch size and dimension size
for fine-tuning on the full graph.

Within the same time budget,
KGTuner can consistently
search better configurations
than the baseline search algorithms 
for seven KG embedding models on WN18RR \citep{dettmers2017convolutional}
and FB15k-237 \citep{toutanova2015observed}.
By applying KGTuner
to the large-scale benchmarks
ogbl-biokg and ogbl-wikikg2 \citep{hu2020open},
the performances of embedding models
are improved compared with the reported results on 
OGB link prediction leaderboard.
Besides,
we justify the improvement of efficiency
via analyzing the design components in KGTuner.


\section{Background: HPs in KG embedding}
\label{sec:revisit}

We firstly revisit 
the important and common HPs in KG embedding.
Following the general framework~\citep{ruffinelli2019you,ali2020bringing},
the learning problem can be written as
\begin{align}
\vspace{-2px}
\!\!\!\!\bm P^* 
\!=\! \arg\min\nolimits_{\bm P} L(F(\cdot, \bm P), D^+, D^-)
\! + \! r(\bm P),
\label{eq:kge}
\vspace{-2px}
\end{align}
where $F$ is the form of an embedding model
with learnable parameters $\bm P$,
$D^+$ is the set of positive samples from the training data,
$D^-$ represents negative samples,
and $r(\cdot)$ is a regularization function.
There are four groups of hyper-parameters (Table~\ref{tab:spacefull}),
i.e., 
the size of \textit{negative sampling} for $D^-$,
the choice of \textit{loss function} $L$,
the form of \textit{regularization} $r(\cdot)$,
and the \textit{optimization} $\arg\min_{\bm P}$.

\begin{table*}[ht]
	\centering
	 \vspace{-7px}
	\caption{The HP space. 
		Conditioned HPs are in parenthesize.
		``adv.'' and ``reg.'' are short for ``adversarial'' and ``regularization'', respectively.
		Please refer to the Appendix~\ref{app:searchspace} for more details.}
	\label{tab:spacefull}
	\vspace{-8px}
\setlength\tabcolsep{4pt}
	\begin{tabular}{c|c|c|c }
		\toprule
		component &   name & type  & range  \\
		\midrule
		negative sampling	
		&	\# negative samples &  cat &  \{32, 128, 512, 2048, \texttt{1VsAll}, \texttt{kVsAll}\}     \\   \midrule
		\multirow{3}{*}{loss function}	&	loss function & cat & \{MR, BCE\_(mean, sum, adv), CE\}     \\
		&	gamma (MR) & float &  [1, 24]         \\ 
		&	adv. weight (BCE\_adv)& float &  [0.5, 2.0]       \\    \midrule
		\multirow{3}{*}{regularization}
		&	regularizer  & cat &   \{FRO, NUC, DURA, None\}   \\
		&	reg. weight (not None) & float  &  [$10^{-12}$, $10^{2}$]    \\
		&	dropout rate  & float   & $[0, 0.5]$    \\ \midrule
		\multirow{6}{*}{optimization}
		&	optimizer & cat &  \{Adam, Adagrad, SGD\}     \\
		&	learning rate &  float & [$10^{-5}$, $10^0$]       \\
		&	initializer & cat &   \{uniform, normal, xavier\_uniform, xavier\_norm\}       \\
		&	batch size & int & \{128, 256, 512, 1024\}  \\
		&	dimension size  & int  &   \{100, 200, 500, 1000, 2000\}   \\ 
		&  inverse relation  &  bool & \{True, False\} \\
		\bottomrule
	\end{tabular}
	\vspace{-4px}
\end{table*}

\vspace{1px}
\noindent
\textbf{Embedding model.}
While there are many existing embedding models,
we follow \citep{ruffinelli2019you} to focus on some representative models.
They are
translational distance models
TransE \citep{bordes2013translating} and RotatE \citep{sun2019rotate},
tensor factorization models
RESCAL \citep{nickel2011three}, DistMult \citep{yang2014embedding},
ComplEx \citep{trouillon2017knowledge} and TuckER \citep{balavzevic2019tucker},
and
neural network models
ConvE \citep{dettmers2017convolutional}.
Graph neural networks for KG embedding
\citep{schlichtkrull2018modeling,vashishth2019composition,zhang2022knowledge}
are not studied here
for their scalability issues on large-scale KGs \citep{ji2020survey}.

\vspace{1px}
\noindent
\textbf{Negative sampling.} 
Sampling negative triplets is important
as only positive triplets are contained in the KGs \citep{wang2017knowledge}.
We can pick up $m$ triplets 
by replacing the head or tail entity with uniform sampling~\citep{bordes2013translating}
or use a full set of negative triplets.
Using the full set
can be defined
as the \texttt{1VsAll} \citep{lacroix2018canonical}
or \texttt{kVsAll} 
 \citep{dettmers2017convolutional}
 according to the positive triplets used.
The methods \cite{cai2018kbgan,zhang2021simple}
requiring additional models for negative sampling
are not considered here.

\vspace{1px}
\noindent
\textbf{Loss function.}
There are three types of loss functions.
One can use margin ranking (MR) loss 
\citep{bordes2013translating} to rank the positive triplets higher over the negative ones,
or use binary cross entropy (BCE) loss,
with variants 
BCE\_mean, BCE\_adv \citep{sun2019rotate}
and
BCE\_sum \citep{trouillon2017knowledge},
to classify the positive and negative triplets as binary classes,
or use cross entropy (CE) loss \citep{lacroix2018canonical}
to classify the positive triplet
as the true label over the negative triplets.

\vspace{1px}
\noindent
\textbf{Regularization.}
To balance the expressiveness and complexity,
and to avoid unbounded embeddings,
the regularization techniques can be considered,
such as
regularizers like
Frobenius norm (FRO) \citep{yang2014embedding,trouillon2017knowledge},
Nuclear norm (NUC) \citep{lacroix2018canonical}
as well as
DURA \citep{zhang2020duality},
and dropout on the embeddings \citep{dettmers2017convolutional}.

\vspace{1px}
\noindent
\textbf{Optimization.}
To optimize the embeddings,
important optimization choices include
the optimizer, such as SGD, 
Adam \citep{kingma2014adam}
and Adagrad \citep{duchi2011adaptive},
learning rate, 
initializers,
batch size,
embedding dimension size,
and add inverse relation \citep{lacroix2018canonical}
or not.


\section{Defining the search problem}
\label{sec:understanding}

Denote an instance $\mathbf x = (x_1, x_2\dots, x_n)$,
which is called an HP configuration,
in the search space $\mathcal X$.
Let $F(\bm P, \mathbf x)$ be an embedding model
with model parameters $\bm P$
and HPs $\mathbf x$,
we define 
$\mathcal M(F(\bm P, \mathbf x), D_{\text{val}})$
as the performance measurement (the larger the better) on validation data $D_{\text{val}}$
and 
$\mathcal L(F(\bm P, \mathbf x), D_{\text{tra}})$
as the loss function (the smaller the better) on training data $D_{\text{tra}}$.
We define the problem 
of HP search for KG embedding models 
in Definition~\ref{def:evaluation}.
The objective is to search 
an optimal configuration $\mathbf x^*\in\mathcal X$
such that the embedding model $F$
can achieve the best performance on the validation data $D_{\text{val}}$.

\begin{definition}[Hyper-parameter search for KG embedding]
\label{def:evaluation}

The problem of HP search for KG embedding model
is formulated as
\begin{align}
\vspace{-2px}
\mathbf x^* &
= \arg\max\nolimits_{\mathbf x\in\mathcal X} \mathcal M\big(F(\bm P^*, \mathbf x), D_{\text{val}} \big),
\label{eq:opthp} 
\\
\bm P^* &
= \arg\min\nolimits_{\bm P} \mathcal L\big(F(\bm P, \mathbf x), D_{\text{tra}}\big).  
\label{eq:optemb}
\vspace{-2px}
\end{align}
\end{definition}

Definition~\ref{def:evaluation}
is a bilevel optimization problem \citep{colson2007overview},
which can be solved by many conventional HP search algorithms.
The most common and widely used approaches are sample-based methods
like grid search and random search \citep{bergstra2012random},
where the HP configurations are independently sampled.
To guide the sampling of HP configurations by historical experience,
SMBO-based methods \citep{bergstra2011algorithms,hutter2011sequential}
learn a surrogate model to 
select configurations based on the results that have been evaluated.
Then,
the model parameters $\bm P$ 
are optimized
by minimizing the loss function $\mathcal L$
on $D_{\text{tra}}$ in Eq.~\eqref{eq:optemb}.
The evaluation feedback $\mathcal M$ of $\mathbf x$ on the validation data $D_{\text{val}}$
is used to update the surrogate.

\begin{figure*}[ht]
	
	\centering
	\vspace{-7px}
	\setlength{\abovecaptionskip}{4pt}
	
	\includegraphics[height=2.9cm]{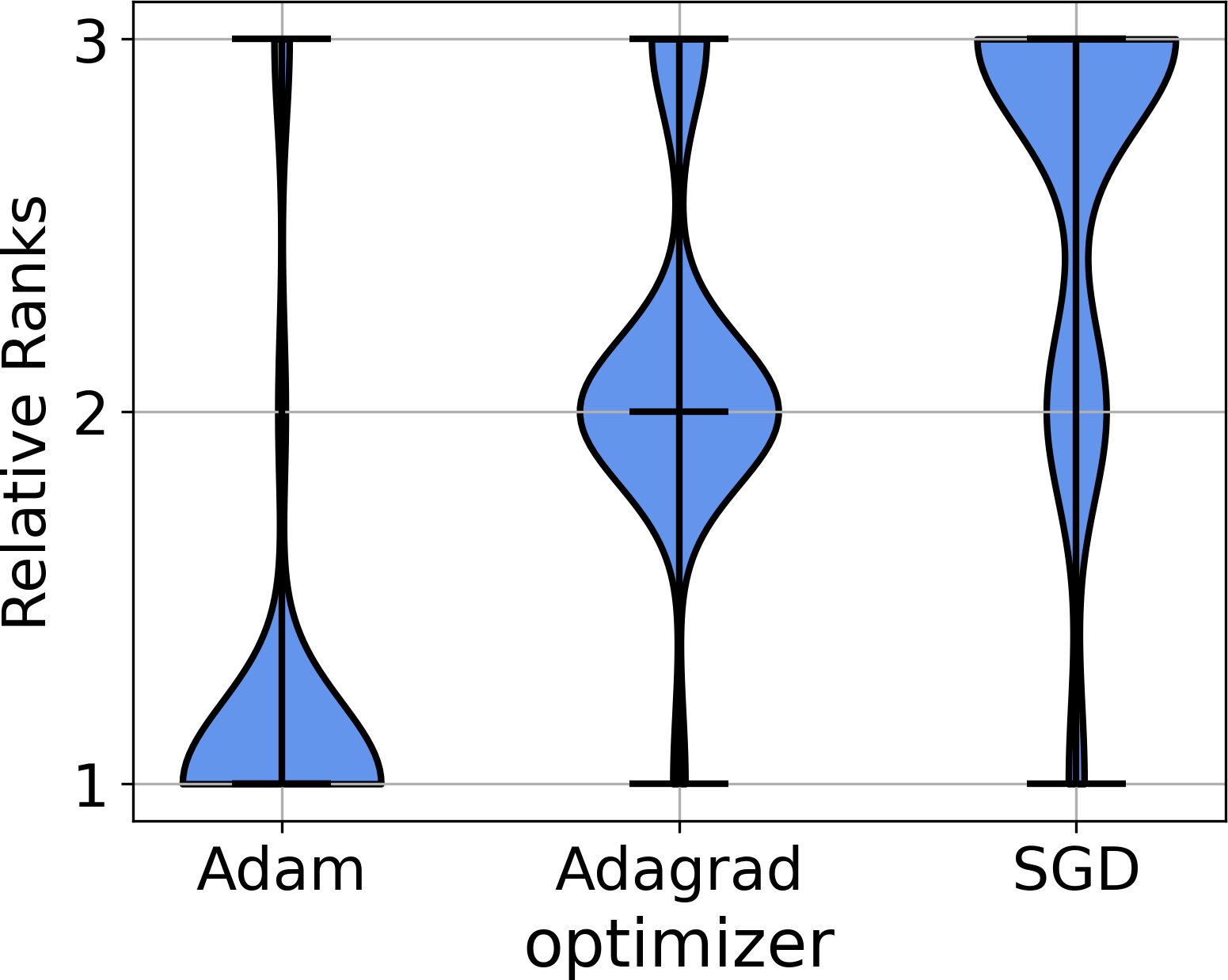}\quad
	\includegraphics[height=2.9cm]{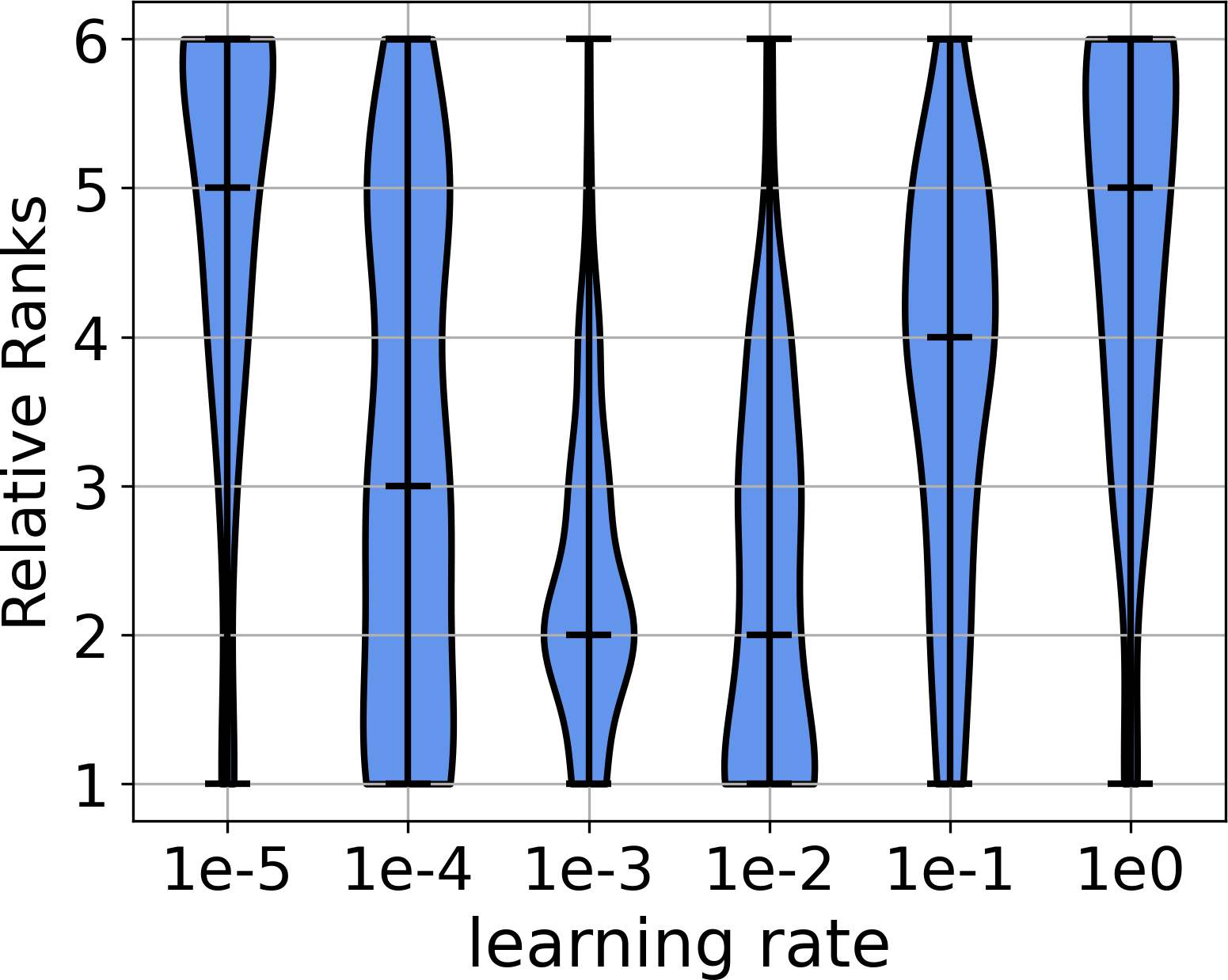}\quad
	\includegraphics[height=2.9cm]{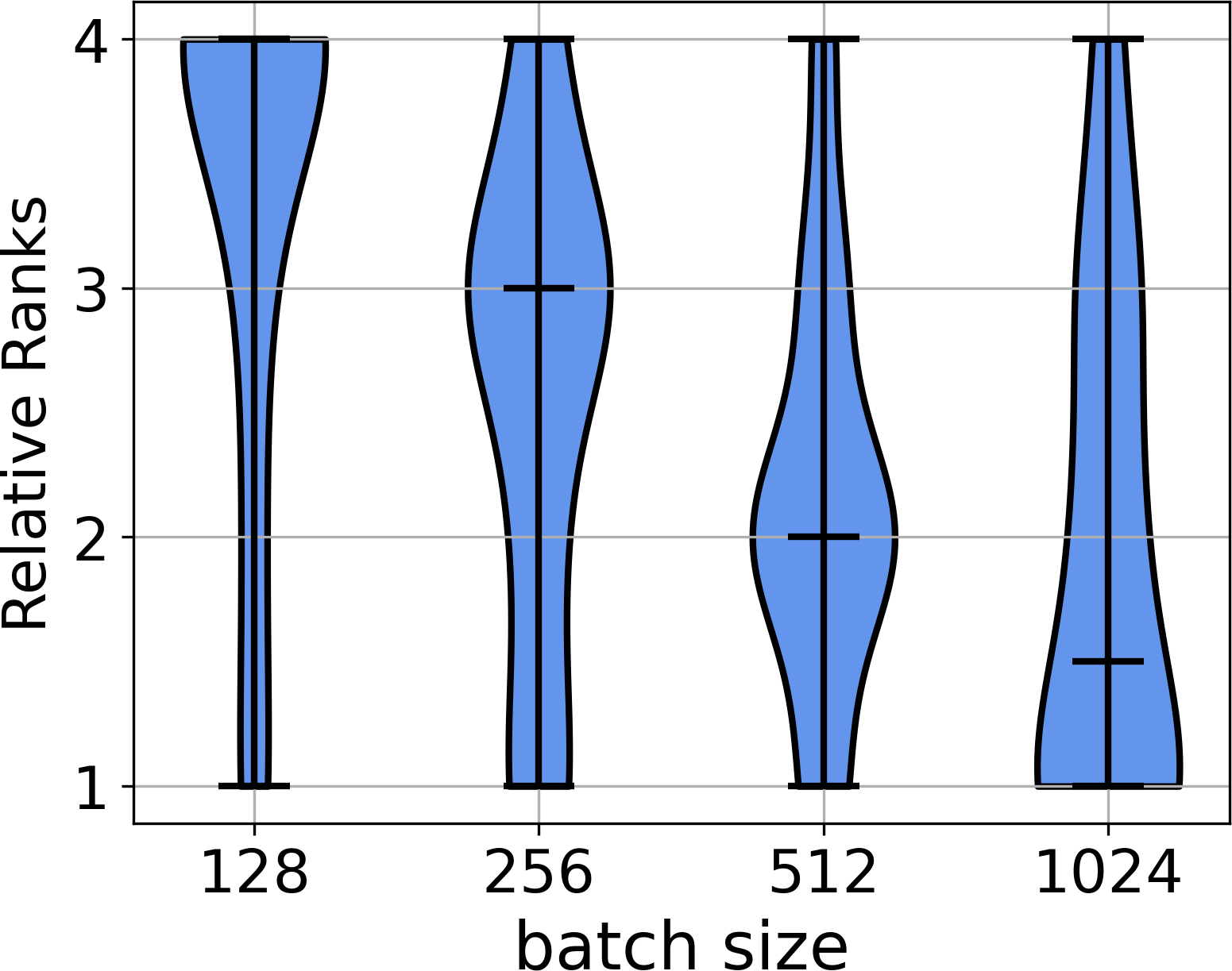}\quad
	\includegraphics[height=2.9cm]{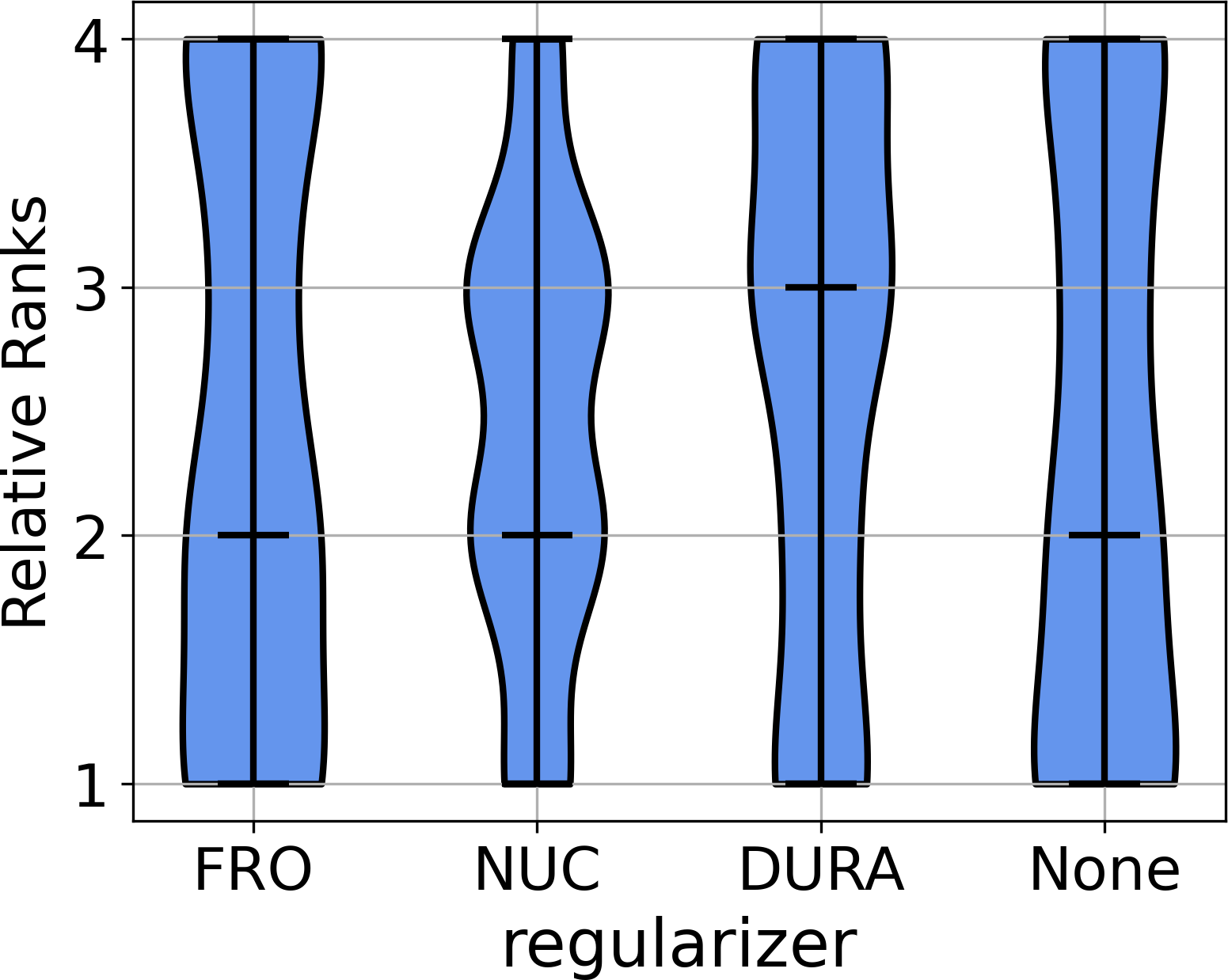}
	
	\caption{Ranking distribution of selected HPs. 
		A value with larger area in the bottom indicates 
		the higher ranking of this value.
		The four figures correspond to the four groups:
		reduced options, shrunken range, monotonously related, no obvious patterns.
		Full results are in the Appendix~\ref{app:under:space}.}	
	\label{fig:hyper-range}
	\vspace{-12px}
\end{figure*}

There are three major aspects determining the efficiency of Definition~\ref{def:evaluation}:
(i)
the size of search space $\mathcal X$,
(ii)
the validation curvature of $\mathcal M(\cdot, \cdot)$ in Eq.~\eqref{eq:opthp},
and (iii)
the evaluation cost in solving
$\arg\min_{\bm P}\mathcal L$ in Eq.~\eqref{eq:optemb}.
However,
the existing methods \cite{ruffinelli2019you,ali2020bringing}
directly search on a huge space
with commonly used surrogate models
and slow evaluation feedback from the full KG
due to the lack of understanding on the search problem,
leading to low efficiency.

\section{Understanding the search problem}
\label{sec:underpro}

To address the mentioned limitations,
we measure the significance 
and correlation of 
each HP
to determine the feasibility of the search space $\mathcal X$ 
in Section~\ref{ssec:indivudial}.
In Section~\ref{ssec:surrogate},
we visualize the HPs 
that determine the curvature of Eq.~\eqref{eq:opthp}.
To reduce the evaluation cost in Eq.~ \eqref{eq:optemb},
we analyze the approximation methods in Section~\ref{ssec:fasteval}.
Following \citep{ruffinelli2019you},
the experiments run on the seven embedding models
in Section~\ref{sec:revisit}
and two widely used datasets
WN18RR \citep{dettmers2017convolutional}
and FB15k-237 \citep{toutanova2015observed}.
The experiments are implemented 
with PyTorch framework \citep{paszke2017automatic},
on a machine with
two Intel Xeon 6230R CPUs
and
eight 
RTX 3090 GPUs with 24 GB memories each.
We provide the implementation details
in the Appendix~\ref{app:implementation}.

\subsection{Search space: $\mathbf{x} \in \mathcal X$}
\label{ssec:indivudial}

Considering such large amount of HP configurations in $\mathcal X$,
we take the simple and efficient approach
where HPs are evaluated under control variate
\citep{hutter2014efficient,you2020design},
which varies the $i$-th HP
while fixing the other HPs.
First, we discretize the continuous HPs
according to their ranges.
Then 
the feasibility of the search space $\mathcal X$
is analyzed by  
checking the ranking distribution and consistency
of individual HPs.
These can help us 
shrink
and
decouple the search space.
The detailed setting for this part is in the Appendix~\ref{app:configgen}.

\vspace{2px}

\noindent
\textbf{Ranking distribution.}
To shrink the search space,
we 
use the ranking distribution
to indicate what HP values perform consistently.
Given an anchor configuration $\mathbf x$,
we obtain the ranking of different values $\theta\in X_i$ by fixing the other HPs,
where $X_i$ is the range of the $i$-th HP.
The ranking distribution is then collected
over the different anchor configurations in $\mathcal X_i$,
different models and datasets.
According to the violin plots of ranking distribution
shown in Figure~\ref{fig:hyper-range},
the HPs can be classified into four groups:
\begin{itemize}[leftmargin=18px, itemsep=0.6pt,topsep=0pt,parsep=0pt,partopsep=0pt]
\item[(a)]
\textit{reduced options}, e.g., Adam is the best optimizer and inverse relation should not be introduced;

\item[(b)] 
\textit{shrunken range}, e.g., learning rate, reg. weight and dropout rate are better in certain ranges;

\item[(c)] 
\textit{monotonously related}: e.g., larger batch size and dimension size tend to be better;

\item[(d)] 
\textit{no obvious patterns}: e.g., the remaining HPs.
\end{itemize}

\vspace{2px}
\noindent
\textbf{Consistency.}
To decouple the search space,
we measure the consistency of configurations' rankings
when only a specific HP changes.
For the $i$-th HP,
if 
the ranking of configurations' performance
is consistent
with different values of $\theta\in X_i$,
we can 
decouple the search procedure of the $i$-th HP
with the others.
We measure such consistency 
with the spearman's ranking correlation coefficient (\texttt{SRCC})
\citep{schober2018correlation}.

\begin{figure*}[ht]
	\centering
	\vspace{-7px}
	\setlength{\abovecaptionskip}{-2pt}
	\subfigure[Ground truth]
	{\includegraphics[height=2.3cm]{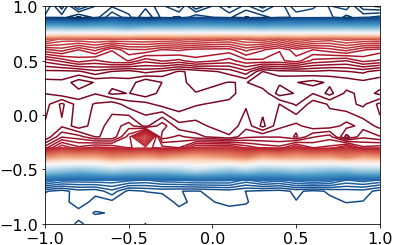} \label{fig:hyperspace_curvature}}	
	\hfill
	\subfigure[GP prediction]
	{\includegraphics[height=2.3cm]{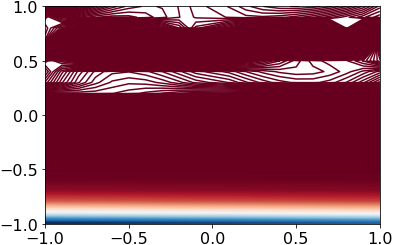}
		\label{fig:gp_curvature}}
	\subfigure[MLP prediction]
	{\includegraphics[height=2.3cm]{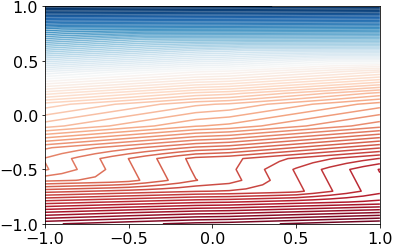}
		\label{fig:mlp_curvature}}
	\subfigure[RF prediction]
	{\includegraphics[height=2.3cm]{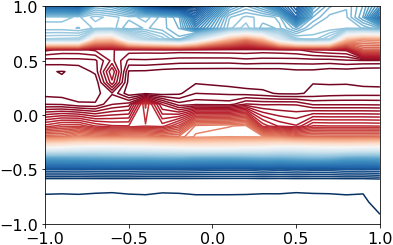}
		\label{fig:rf_curvature}}
	
	\caption{Curvature of the search space and three surrogate models. 
		The search space curvature is quite complex
		with many local maximum areas.
		The curvature of RF approximate the ground truth best.}
	\vspace{-12px}
	\label{fig:hyperspace_curvature-range}
\end{figure*}

Given a value $\theta\in X_i$,
we obtain the ranking $r(\mathbf x, \theta)$ of  the anchor configurations
$\mathbf x \in \mathcal X_i$
by fixing the $i$-th HP as $\theta$.
Then,
the \texttt{SRCC} between the two HP values $\theta_1, \theta_2\in X_i$ 
is computed as
\begin{align}
	1 -
	\frac{\sum_{\mathbf x\in\mathcal X_i}\!|r(\mathbf x,\theta_1)\!-\!r(\mathbf x,\theta_2)|^2}
	{|\mathcal X_i|\cdot (|\mathcal X_i|^2-1)},
	\vspace{-1px}
	\label{eq:spearman}
\end{align}
where $|\mathcal X_i|$
means the number of anchor configurations in $\mathcal X_i$.
\texttt{SRCC} indicates the matching rate of rankings 
for the anchor configurations in $\mathcal X_i$
with respect to $x_i=\theta_1$ and $x_i=\theta_2$.
Then
the consistency of the $i$-th HP is evaluated by 
averaging the \texttt{SRCC} over 
the different pairs of $(\theta_1, \theta_2)$ for $X_i$,
the different models and different datasets.
The larger consistency (in the range $[-1,1]$) indicates
that changing the value of the $i$-th HP
does not influence much on the other configurations' ranking.

\begin{figure}[ht]
	\centering
	\setlength{\abovecaptionskip}{-2pt}
	\includegraphics[width=0.95\columnwidth]{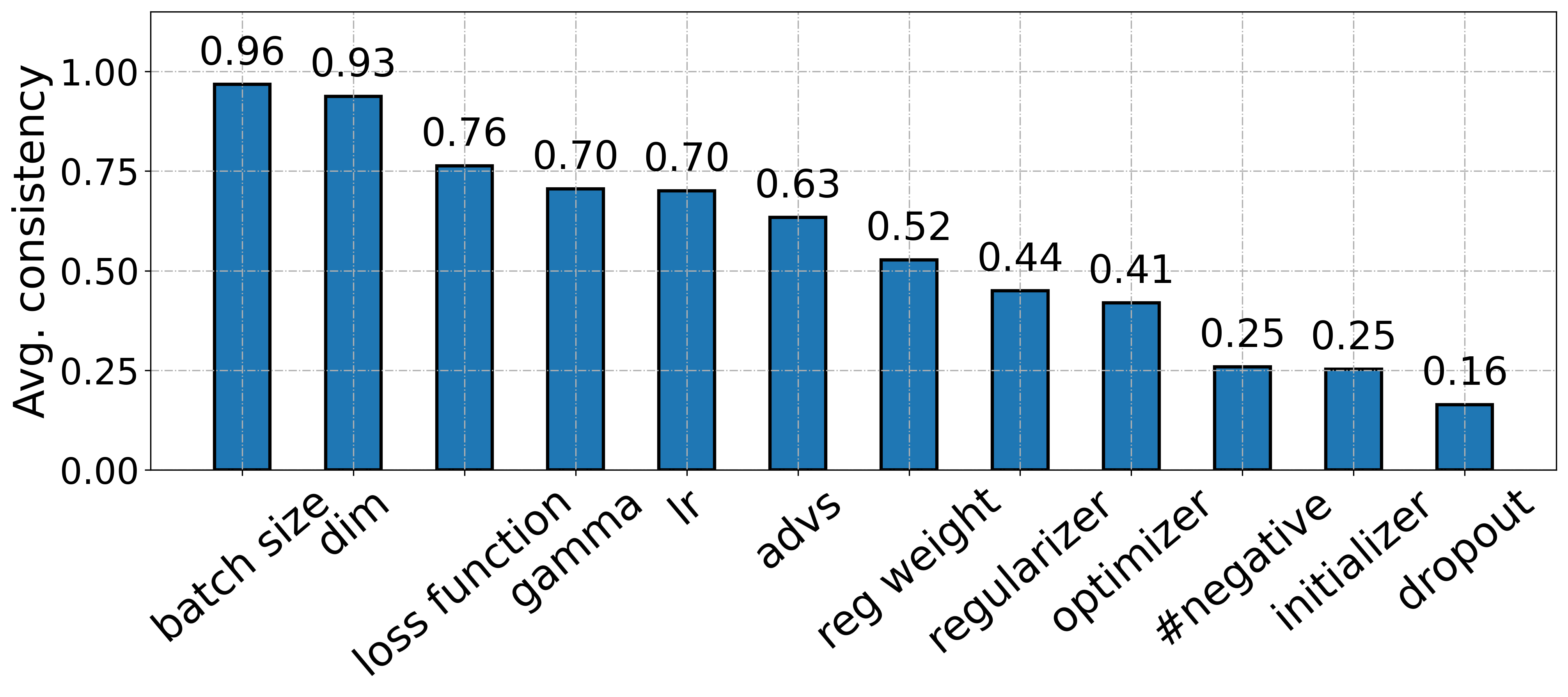}
	\caption{Consistency of each HP.}	
	\label{fig:HP_consistency}
	\vspace{-12px}
\end{figure}

As in Figure~\ref{fig:HP_consistency},
the batch size and dimension size show higher consistency than the other HPs.
Hence,
the evaluation of the configurations
can be consistent
with different choices of  the two HPs.
This indicates that
we can decouple the search of batch size and dimension size with the other HPs.

\subsection{Validation curvature: $\mathcal{M}(\cdot, \cdot)$}
\label{ssec:surrogate}


We analyze the curvature of the validation performance
$\mathcal{M}(\cdot, \cdot)$ w.r.t $\mathbf x \in \mathcal{X}$.
Specifically,
we follow \citep{li2017visualizing} 
to visualize the validation loss' landscape 
by uniformly varying the numerical HPs 
in two directions (20 configurations  in each direction)
on the model ComplEx and dataset WN18RR.
From Figure~\ref{fig:hyperspace_curvature},
we observe that the curvature is quite complex
with many local maximum areas.

To gain insights from evaluating these configurations
and guide the next configuration sampling,
we learn a surrogate model 
as the predictor
to approximate the validation curvature.
The curvatures of
three common
surrogates,
i.e.,
Gaussian process (GP)~\citep{williams1996gaussian},
multi-layer perceptron (MLP)~\citep{gardner1998artificial}
and random forest (RF)~\citep{breiman2001random},
are in Figures~\ref{fig:gp_curvature}-\ref{fig:rf_curvature}.
The surrogate models are trained with the evaluations of $100$ random configurations 
in the search space.
As shown,
both GP and MLP fail to capture the complex local surface in Figure~\ref{fig:hyperspace_curvature}
as they tend to learn a flat and smooth distribution in the search space.
In comparison,
RF is better in capturing the local distributions.
Hence,
we regard RF as a better choice in the search space.
A more detailed comparison on the approximation ability
of different surrogates is in the Appendix~\ref{app:fitting}.

\subsection{Evaluation cost: $\arg\min_{\bm{P}} \mathcal{L}$}
\label{ssec:fasteval}

The evaluation cost of the HP configuration on an embedding model  is the majority computation cost
in HP search.
Thus,
we firstly evaluate the HPs
that have influence on the evaluation cost,
including 
batch size, dimension size,
number of negative samples
loss function and regularizer.
Then,
we analyze the evaluation transfer ability
from small subgraph to the full graph.

\vspace{1px}

\noindent
\textbf{Cost of different HPs.}
The cost of each HP value
$\theta \in X_i$
is averaged
over the different anchor configurations in $\mathcal X_i$, 
different models and datasets.
For fair comparsion,
the time cost is counted
per thousand iterations.
We find that 
the evaluation cost increases significantly with larger batch size and dimension size,
while the number of negative samples
and the choice of loss function or regularizer
do not have much influence on the cost.
We provide two exemplar curves in Figure~\ref{fig:hyper-cost}
and put the remaining results in the Appendix~\ref{app:under:cost}.

\begin{figure}[ht]
	\centering
	\vspace{-7px}
	\setlength{\abovecaptionskip}{-2pt}
	\subfigure{\includegraphics[width=0.44\columnwidth]{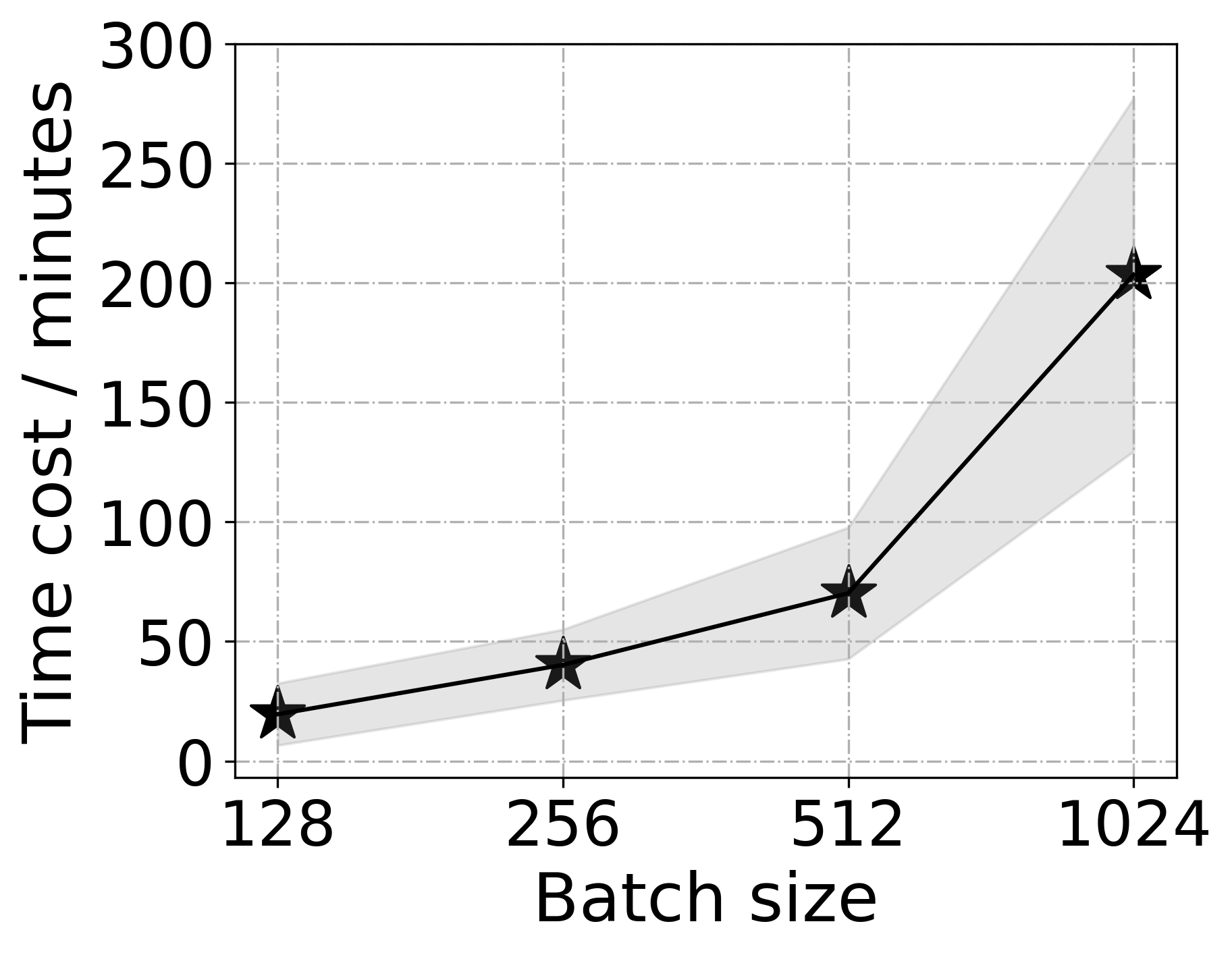}} \quad\ 
	\subfigure{\includegraphics[width=0.44\columnwidth]{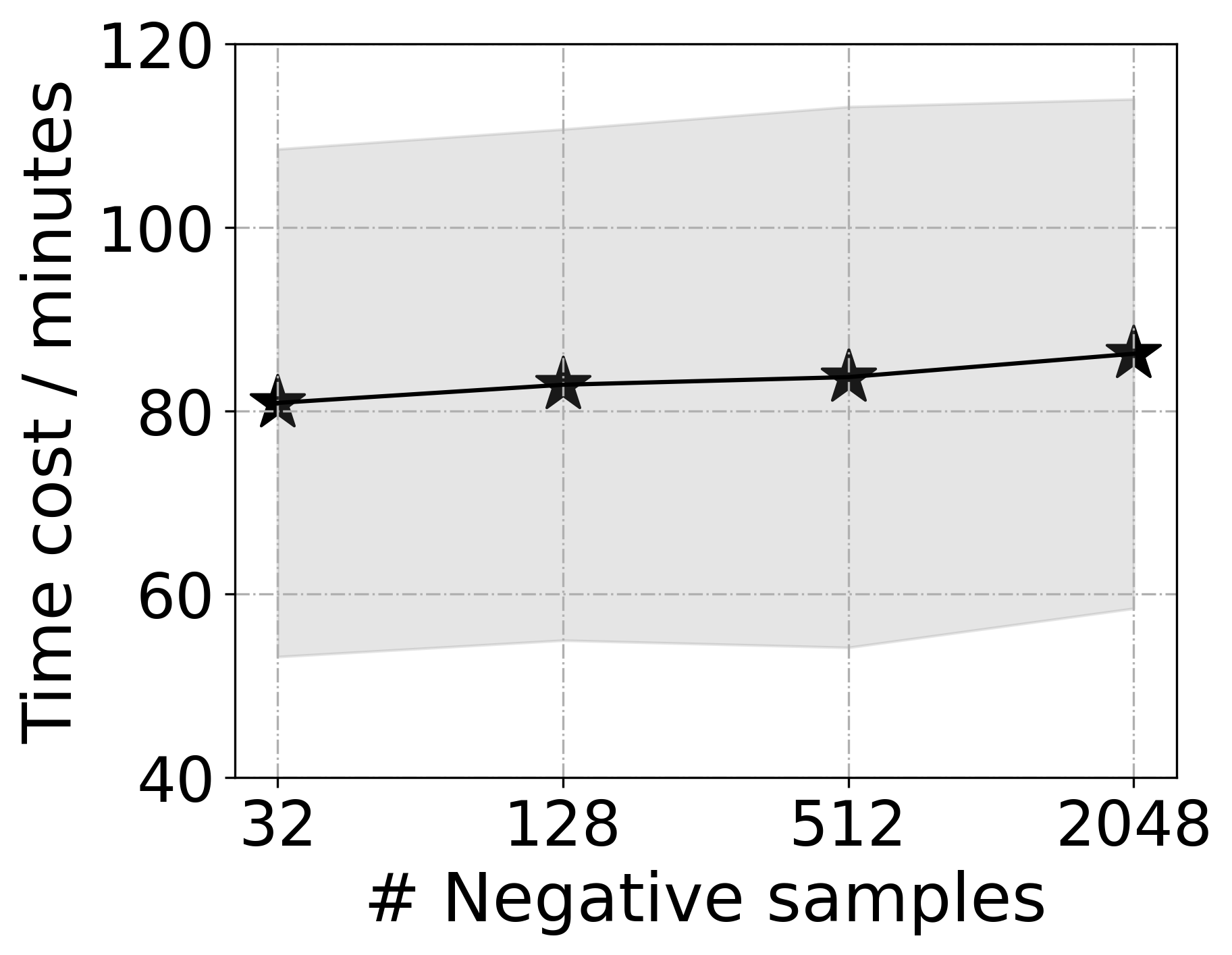}}
	\caption{Computing time cost. The dots are the average and the shades are the standard deviation.}	
	\label{fig:hyper-cost}
	\vspace{-7px}
\end{figure}

\noindent
\textbf{Transfer ability of subgraphs.}
Subgraphs can efficiently approximate the properties of the full graph
\citep{hamilton2017inductive,teru2020inductive}.
We evaluate the impact of subgraph sampling
on HP search by 
checking the 
consistency
between evaluations results on small subgraph and 
those on the full graph.

\begin{figure}[ht]
	\centering
	\setlength{\abovecaptionskip}{-1pt}
	\includegraphics[width=0.95\columnwidth]{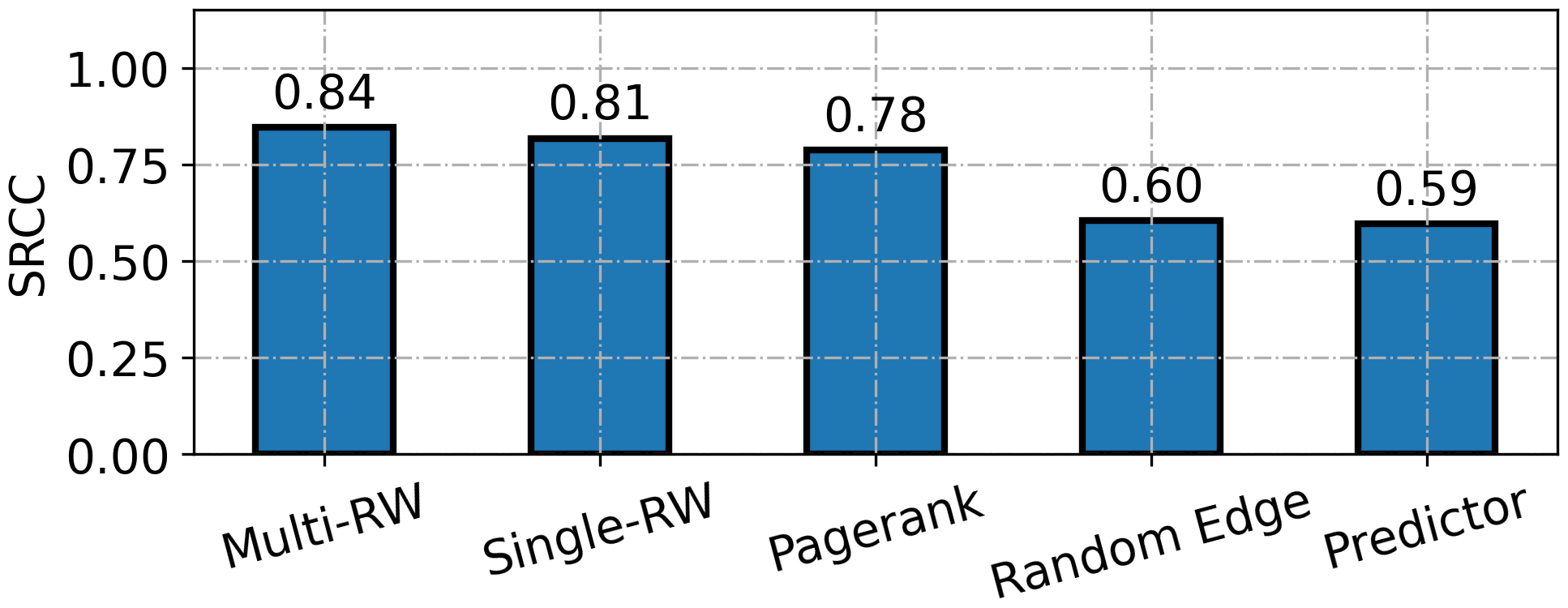}
	\caption{Comparison of the sampling methods.}	
	\vspace{-8px}
	\label{fig:sampling}
\end{figure}

First,
we study how to sample subgraphs.
There are several approaches to sample small subgraphs
from a large graph~\cite{leskovec2006sampling}.
We compare four representative approaches in 
Figure~\ref{fig:sampling},
i.e.,
Pagerank node sampling (Pagerank),
random edge sampling (Random Edge),
single-start random walk (Single-RW)
and multi-start random walk (Multi-RW).
For a fair comparison,
we constrain the subgraphs with about $20\%$ of the full graph.
The consistency between the sampled subgraph with the full graph is evaluated
by the \texttt{SRCC} in (\ref{eq:spearman}).
We observe that
multi-start random walk is the best among the different sampling methods.


Apart from directly transferring the evaluation from subgraph to full graph,
we can alternatively 
train a predictor with 
observations on subgraphs 
and then
transfers the model to predict the configuration performance on the full graph.
From Figure~\ref{fig:sampling},
we find that
directly transferring evaluations from subgraphs to the full graph is much better than
transferring the predictor model.

\begin{figure}[ht]
	\centering
	\setlength{\abovecaptionskip}{2pt}
	\includegraphics[width=0.95\columnwidth]{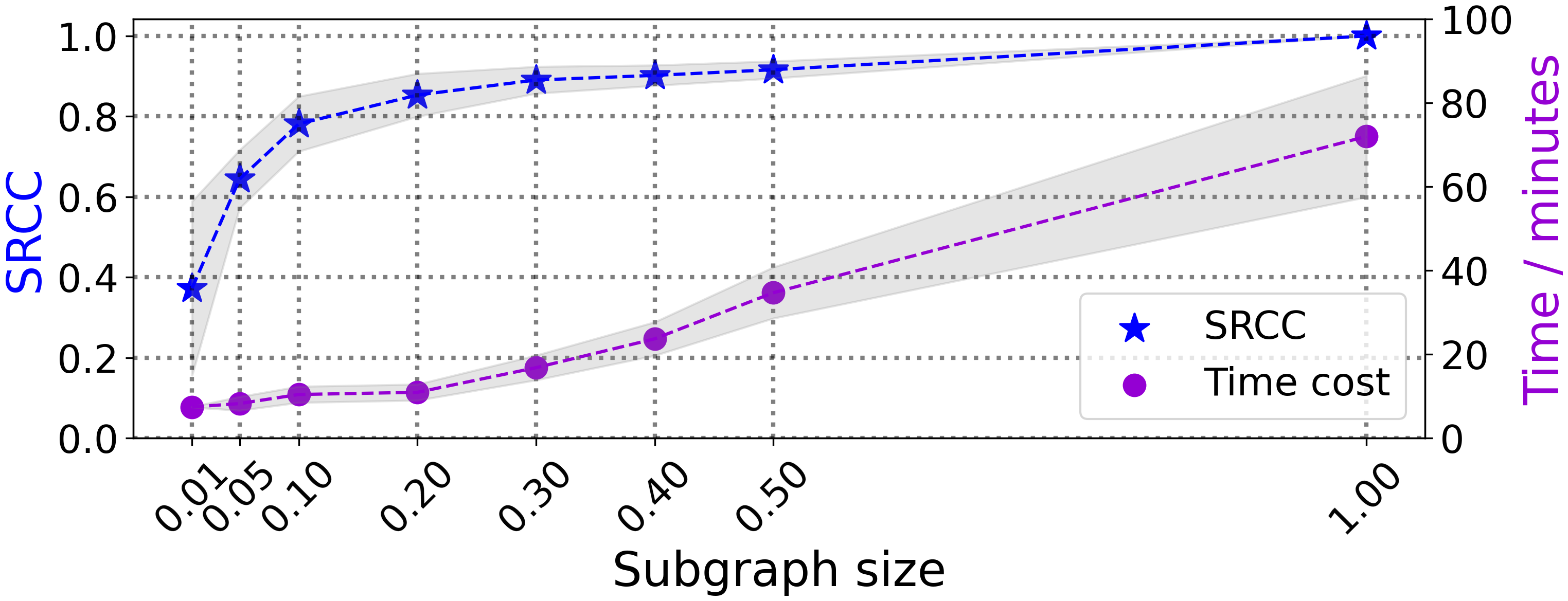}
	\caption{Consistency and cost of different subgraph sizes, where the shades are the standard deviation.}
	\label{fig:subgraph_correlation}
	\vspace{-10px}
\end{figure}

In addition,
we show the consistency and cost in terms of different subgraph sizes 
(percentage of entities compared to the full graph)
in Figure~\ref{fig:subgraph_correlation}.
As shown,
evaluation on subgraphs can significantly improve the efficiency.
When the scale increases,
the consistency increases
but the cost also increases.
To balance the consistency and cost,
the subgraphs with $20\%$
entities are the better choices.

\section{Efficient search algorithm}
\label{sec:algorithm}

By analyzing the ranking distribution 
and consistency of HPs in Section~\ref{ssec:indivudial},
we observe that not all the HP values are 
equivalently good,
and some HPs can be decoupled.
These observations
motivate us to revise the search space 
in Section~\ref{ssec:reduce}.
Based on the
analysis in Section~\ref{ssec:surrogate} and~\ref{ssec:fasteval},
we then propose an efficient two-stage search algorithm in Section~\ref{ssec:transfer}.



\subsection{Shrink and decouple the search space}
\label{ssec:reduce}

To shrink the search space,
we mainly consider groups (a) and (b)
of HPs in Section~\ref{ssec:indivudial}.
From the full results in the Appendix~\ref{app:under:space},
we observe that Adam can consistently perform better than the other two optimizers,
the learning rate is better in the range of $[10^{-4}, 10^{-1}]$,
the regularization weight is better in $[10^{-8}, 10^{-2}]$,
dropout rate is better in $[0, 0.3]$,
and add inverse relation is not a good choice.

To decouple the search space,
we consider batch size and dimension size
that have larger consistency values 
than the other HPs,
and are monotonously related to the performance as in group (c).
However,
the computation costs of batch size and dimension size
increase prominently as shown in Figure~\ref{fig:hyper-cost}.
Hence,
we can set batch size as 128 and dimension size as 100
to search the other HPs with low evaluation cost
and increase their values in a fine-tuning stage.

Given the full search space $\mathcal X$,
we denote the shrunken space as ${\mathcal X}_\text{S}$
and the further decoupled space as ${\mathcal X}_\text{S|D}$.
We achieve hundreds of times size reduction from  ${\mathcal X}_\text{S}$ to ${\mathcal X}_\text{S|D}$
and we show the details of 
changes
in the Appendix \ref{app:algorithm}.

\begin{figure*}[ht]
	\centering
	\vspace{-5px}
	\setlength{\abovecaptionskip}{-2pt}
	{\includegraphics[width=1.9\columnwidth]{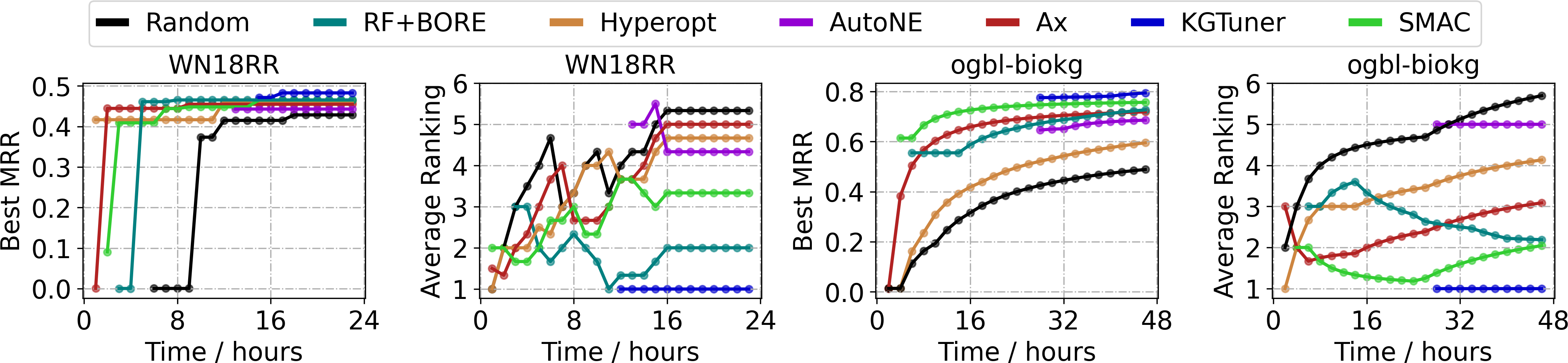}}
	\vspace{5px}
	\caption{Search algorithm comparison (viewed in color).
		The dots are the results collected per hour.}
	\label{fig:HPO_comparison_mrr}
	\vspace{-12px}
\end{figure*}

\subsection{Two-stage search algorithm (KGTuner)}
\label{ssec:transfer}

As discussed in Section~\ref{ssec:fasteval},
the evaluation cost can be significantly reduced
with small batch size, 
dimension size
and subgraph.
This
motivates us to design a two-stage search algorithm,
named KGTuner,
as in
Figure~\ref{fig:KGTuner}
and
Algorithm~\ref{alg:search}.

\begin{algorithm}[ht]
	\centering
	\caption{KGTuner: two-stage search algorithm}
	\label{alg:search}
	\small
	\begin{algorithmic}[1]
		\REQUIRE KG embedding model $F$, dataset $D$, and budget $B$;
		
		\STATE shrink the search space $\mathcal{X}$ to ${\mathcal X}_\text{S}$ and decouple ${\mathcal X}_\text{S}$ to ${\mathcal X}_{\text{S|D}}$;
		\\ \textbf{\# state one}: \textit{efficient evaluation on subgraph}
		
		\STATE sample a subgraph (with $20\%$ entities) $G$ from $D_{\text{tra}}$ by multi-start random walk;
		\label{step:samplesub}
		
		\REPEAT   \label{step:stage1:start}
		\STATE sample a configuration $\hat{\mathbf x}$ from ${\mathcal X}_{S|D}$ by RF+BORE;
		\STATE evaluate $\hat{\mathbf x}$ on the subgraph $G$ to get the performance;
		\STATE update the RF with record $\big(\hat{\mathbf x}, \mathcal M(F(P^*,\hat{\mathbf x}), G_{\text{val}})\big)$;
		\UNTIL{$\nicefrac{B}{2}$ budget exhausted;}\label{step:stage1:end}
		\STATE save the \textit{top10} configurations in ${\mathcal X}_\text{S|D}^*$; \\
		\textbf{\# state two}: \textit{fine-tune the top configurations}
		\STATE increase the batch/dimension size in ${\mathcal X}_\text{S|D}^*$ to get $\tilde{\mathcal X}^*$;
		\STATE set $y^*=0$ and re-initialize the RF surrogate;
		\REPEAT  \label{step:stage2start}
		\STATE select a configuration $\tilde{\mathbf x}^*$ from $\tilde{\mathcal X}^*$ by RF+BORE;
		\STATE evaluate on full graph $G$ to get the performance;
		\STATE update the RF with record $\big(\tilde{\mathbf x}^*\!\!, \mathcal M(F(P^*\!,\tilde{\mathbf x}^*), \!D_{\text{val}})\!\big)$;
		\STATE \textbf{if} $\mathcal M(F(P^*,\tilde{\mathbf x}^*), D_{\text{val}})>y^*$ \textbf{then}
		\\
		$y^*\!\leftarrow \mathcal M(F(P^*,\tilde{\mathbf x}^*), D_{\text{val}})$ 
		and $\mathbf x^* \leftarrow \tilde{\mathbf x}^*$; 
		\textbf{end if}
		\UNTIL{the remaining $\nicefrac{B}{2}$ budget exhausted;}  \label{step:stage2end}
		\RETURN $\mathbf x^*$.
	\end{algorithmic}
\end{algorithm}

\begin{itemize}[leftmargin=*,noitemsep,topsep=0pt,parsep=0pt,partopsep=0pt]
\item In the first stage,
we sample a subgraph $G$ with $20\%$ entities from the full graph $D_{\text{tra}}$ 
by multi-start random walk.
Based on the understanding of curvature  in Section~\ref{ssec:surrogate},
we use the surrogate model random forest (RF)  
under the state-of-the art framework BORE \cite{tiao2021bore},
denoted as RF+BORE,
to explore HPs in ${\mathcal X}_\text{S|D}$
on the subgraph $G$
in steps~\ref{step:stage1:start}-\ref{step:stage1:end}.
The \textit{top10} configurations evaluated in this stage are saved in a set ${\mathcal X}^*_\text{S|D}$.

\item In the second stage,
we increase batch size and dimension size for configurations in ${\mathcal X}^*_\text{S|D}$ 
to generate a new set $\tilde{\mathcal X}^*$.
Then, the configurations in $\tilde{\mathcal X}^*$ are searched 
by the
RF+BORE again in steps~\ref{step:stage2start}-\ref{step:stage2end}
until the remaining $\nicefrac{B}{2}$ budget is exhausted.

\item Finally,
the configuration $\mathbf x^*$ achieving the best performance
on the full validation data $D_{\text{val}}$ is returned for testing.
\end{itemize}

\subsection{Discussion}

e now 
summarize the main differences of KGTuner
with the existing HP search algorithms,
i.e. Random (random search) \citep{bergstra2012random},
Hyperopt \cite{bergstra2013hyperopt},
SMAC \citep{hutter2011sequential},
RF+BORE \citep{tiao2021bore},
and AutoNE \citep{tu2019autone}.

\begin{table}[ht]
	\centering
	\vspace{-3px}
	\caption{Comparison of HP search algorithms.}
	\vspace{-7px}
	\setlength\tabcolsep{4.5pt}
	\small
	\label{tab:methods_compare}
	\begin{tabular}{c | c | c | c | c}
		\toprule
		& \multicolumn{2}{c|}{search space} & surrogate & fast       \\ 
		& shrink & decouple                & model     & evaluation \\ \midrule
		Random  &   $\times$  &  $\times$   &   $\times$  & $\times$ \\
		Hyperopt  &  $\times$  &  $\times$   &   TPE  & $\times$ \\
		Ax  &  $\times$  &  $\times$   &   GP  & $\times$ \\
		SMAC  &  $\times$  &  $\times$   &   RF  & $\times$ \\
		RF+BORE  &  $\times$  &  $\times$   &   RF  & $\times$ \\
		AutoNE  & $\times$  &  $\times$   &   GP  &  $\surd$ \\
		KGTuner & $\surd$  &   $\surd$    &   RF   &	$\surd$  \\
		\bottomrule
	\end{tabular}
	\vspace{-3px}
\end{table}

The comparison is based on three aspects,
i.e.,
search space,
surrogate model
and fast evaluation,
in Table~\ref{tab:methods_compare}.
KGTuner shrinks and decouples the search space
based on the understanding of HPs' properties,
and uses the surrogate RF based on the understanding on validation curvature.
The fast evaluation on subgraph in KGTuner selects the \textit{top10} configurations 
to directly transfer for fine-tuning,
while AutoNE~\citep{tu2019autone} just uses fast evaluation on subgraphs to train the surrogate model,
and transfers the surrogate model for HP search on the full graph.
In Figure~\ref{fig:sampling},
the transfer ability of the surrogate model is shown to be much worse
than direct transferring.

\section{Empirical evaluation}

\subsection{Overall performance}
\label{ssec:exp:alg}

In this part,
we compare the proposed algorithm KGTuner
with six 
HP search algorithms in Table~\ref{tab:methods_compare}.
For AutoNE,
we allocate half budget for it to search on the subgraph
and another half budget on the full graph with the transferred surrogate model.
The baselines search in the full search space (in Table~\ref{tab:spacefull}) 
with the 
same amount of budget as KGTuner.
We use the mean reciprocal ranking (MRR, the larger the better) \citep{bordes2013translating}
to indicate the performance.

\begin{figure*}[ht]
	\centering
	\vspace{-5px}
	\subfigure[Full search processes]{
		\includegraphics[height=2.7cm]{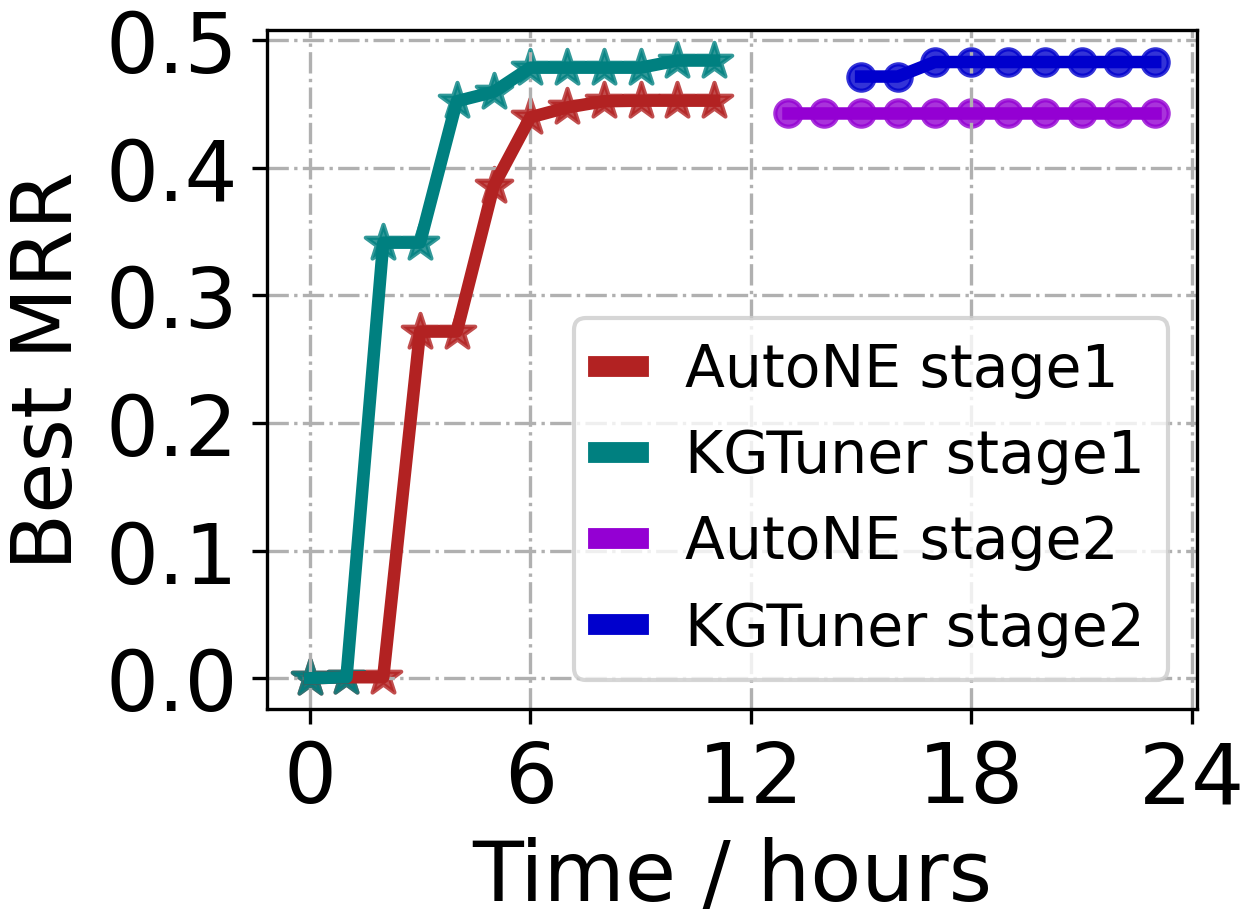}
		\label{fig:two_stage_process}}
	\hfill
	\subfigure[Search space]{
		\includegraphics[height=2.7cm]{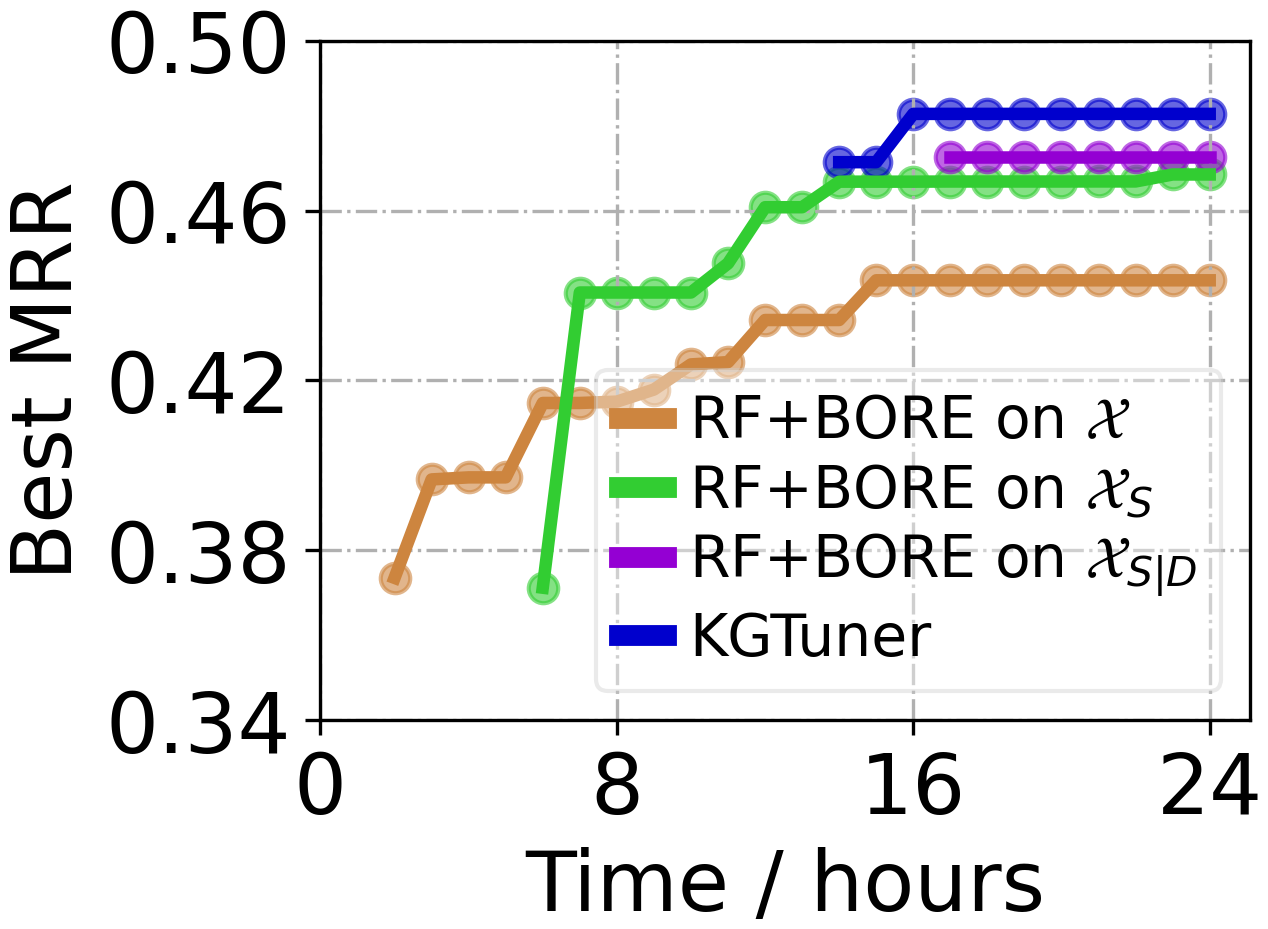}
		\label{fig:ablation_space}}
	\subfigure[Subgraph size]{
		\includegraphics[height=2.7cm]{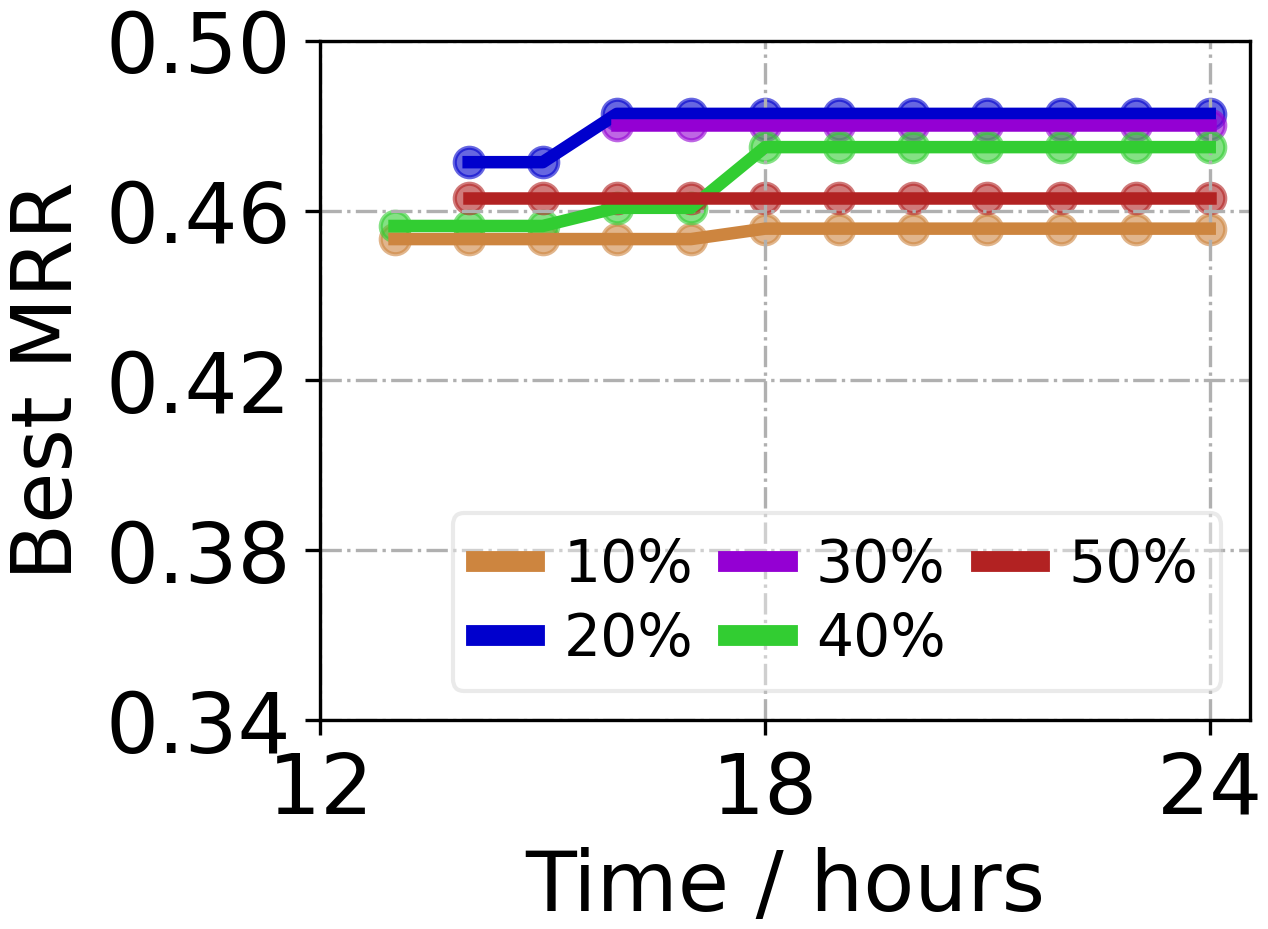}
		\label{fig:ablation_subgraph_size}}
	\subfigure[First-stage budget]{
		\includegraphics[height=2.7cm]{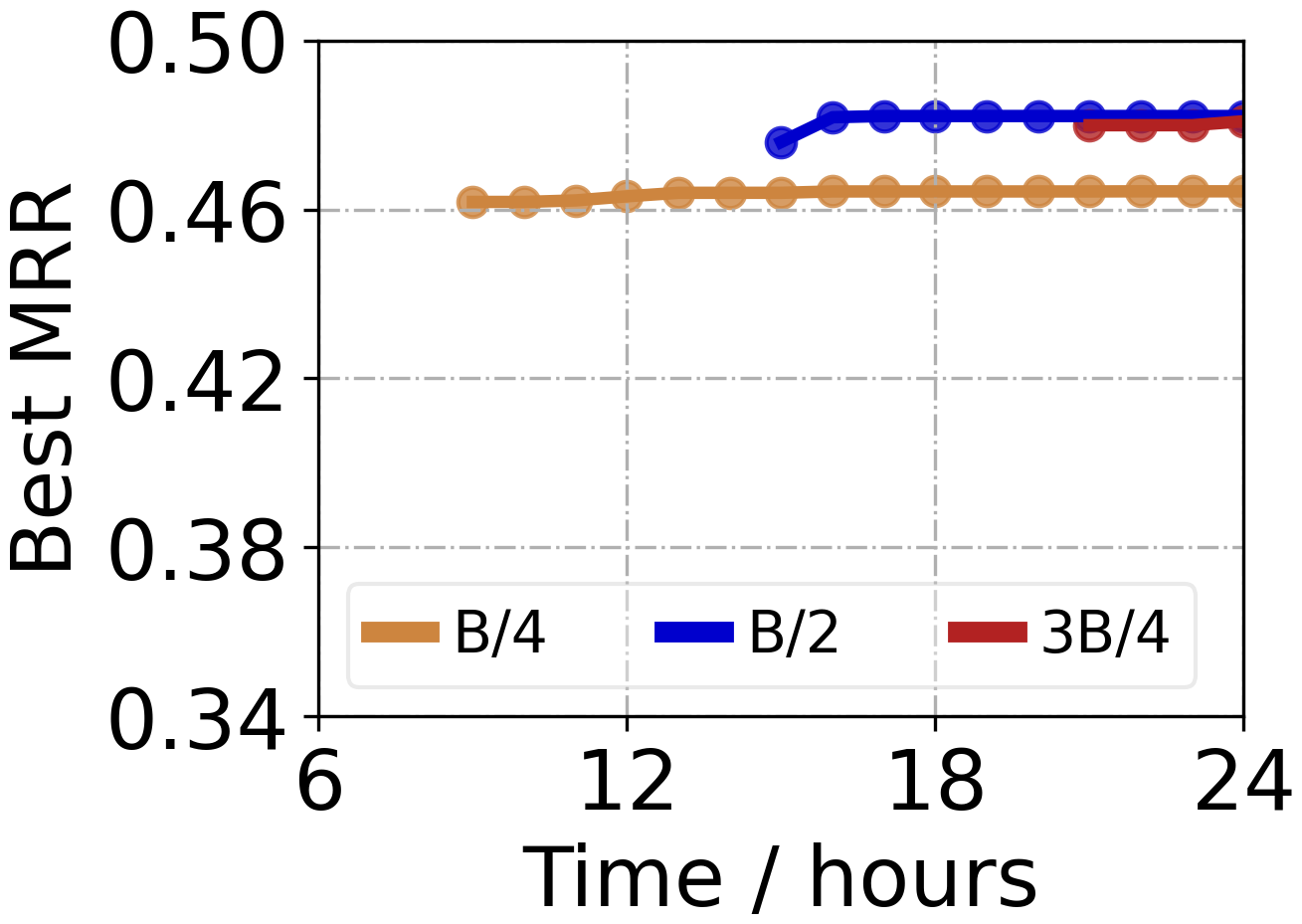}
		\label{fig:ablation_tradeoff}}
	\vspace{-12px}
	\caption{(a): full search processes of the two-stage algorithms. 
		(b-d): ablation studies on KGTuner. Model ComplEx and dataset WN18RR are used in these experiments.}
	\label{fig:HPO_ablation}
	\vspace{-10px}
\end{figure*}

\vspace{2px}
\noindent
\textbf{Efficiency.}
We compare the different search algorithms in Figure~\ref{fig:HPO_comparison_mrr}
on an in-sample dataset WN18RR 
and an out-of-sample dataset ogbl-biokg.
The time budget we set for WN18RR is one day's clock time,
while that for ogbl-biokg is two days' clock time.
For each dataset we show two kinds of figures.
First, the best performance achieved along the clock time
in one experiment on a specific model ComplEx.
Second, we plot the 
the ranking of each algorithm averaged over all the models and datasets.
Since AutoNE and KGTuner run on the subgraphs in the first stage,
the starting points of them locate after 12 hours.
The starting point of KGTuner is a bit later than AutoNE
since it constrains to use large batch size and dimension size in the second stage,
which is more expensive.
As shown,
random search is the worst.
SMAC and RF+BORE
achieve better performance 
than Hyperopt and Ax
since
RF can fit the space better
than TPE and GP as in Section~\ref{ssec:surrogate}.
Due to the weak transfer ability of the predictor (see Figure~\ref{fig:sampling})
and the weak approximation ability of GP (see Figure~\ref{fig:hyperspace_curvature-range}),
AutoNE also performs bad.
KGTuner is much better
than all the baselines.
We show the full search process of the two-stage algorithms
AutoNE and KGTuner on WN18RR in Figure~\ref{fig:two_stage_process}.
By exploring sufficient number of configurations in the first stage,
the configurations fine-tuned in the second stage
can consistently achieve the best performance.



\vspace{2px}
\noindent
\textbf{Effectiveness.}
For WN18RR and FB15k-237,
we provide the reproduced results on TransE, ComplEx and ConvE
with the original HPs, HPs searched by LibKGE
and HPs searched by KGTuner
in Table~\ref{tab:mrr:part}.
The full results on the remaining four embedding models
RotatE, RESCAL, DistMult and TuckER are in the Appendix~\ref{app:general-benchmark}.
Overall, KGTuner achieves better performance
compared with both
the original reported results
and the reproduced results in \citep{ruffinelli2019you}.
We observe that
the tensor factorization models such as RESCAL,
ComplEx and TuckER
have better performance than the translational distance models TransE, RotatE
and neural network model ConvE.
This conforms with the theoretical analysis
that tensor factorization models are more expressive \citep{wang2018multi}.

\begin{table}[ht]
	\centering
	\caption{MRR of models with HPs tuned in different methods. 
		The bold numbers mean the best performance of the same model.}
	\label{tab:mrr:part}
	\small
	\vspace{-8px}
	\begin{tabular}{cc|c|c}
		\toprule
		source                          & models  & WN18RR & FB15k-237 \\
		\midrule
		\multirow{3}{*}{original}       & TransE  & 0.226  & 0.296     \\
		& ComplEx & 0.440  & 0.247     \\
		& ConvE   & 0.430  & 0.325     \\
		\midrule
		\multirow{3}{*}{LibKGE}      & TransE  & 0.228  & 0.313     \\
		& ComplEx & 0.475  & 0.348     \\
		& ConvE   & \textbf{0.442}  & \textbf{0.339}     \\
		\midrule
		\multirow{3}{*}{KGTuner} & TransE  & \textbf{0.233}  & \textbf{0.327}     \\
		& ComplEx & \textbf{0.484}  & \textbf{0.352}     \\
		& ConvE   & 0.437  & 0.335    	\\
		\bottomrule
	\end{tabular}
		\vspace{-2px}
\end{table}

To further demonstrate the advantage of KGTuner,
we apply it to the  Open Graph Benchmark (OGB) \citep{hu2020open},
which is a collection of realistic and large-scale 
benchmark datasets for machine learning on graphs.
Many embedding models
have been tested there
by two large-scale KGs
for link prediction,
i.e., 
ogbl-biokg and ogbl-wikikg2.
Due to their scale,
the evaluation cost of a HP configuration
is very expensive.
We use KGTuner to search HPs for embedding models,
i.e.,
TransE, RotatE, DistMult, ComplEx and AutoSF \cite{zhang2020autosf},
on OGB.
Since the computation costs of
the two datasets are much higher,
we set the time budget as 2 days for ogbl-biokg and 5 days for ogbl-wikikg2.
All the embedding models
evaluated here are constrained to
have the same (or lower) number of model parameters\footnote{
	We run all models on ogbl-wikikg2 with 100 dimension size
	to avoid out-of-memory, instead of 500 on OGB board.}.
More details on model parameters, 
standard derivation,
and validation performance are in the Appendix~\ref{app:ogb}.
As shown in 
Table~\ref{tab:ogb},
KGTuner consistently improves the performance
of the four embedding models
with the same or fewer parameters
compared with the results on the OGB board.

\begin{table}[ht]
	\centering
	\caption{Performance in MRR
		in OGB link prediction board {\small \url{https://ogb.stanford.edu/docs/leader_linkprop/}}
		and those reproduced by KGTuner on ogbl-biokg and ogbl-wikikg2.
		Relative improvements are in parenthesize.
	}
	\vspace{-7px}
	\setlength\tabcolsep{2.5pt}
	\small
	\label{tab:ogb}
	\begin{tabular}{cc|c|c}
		\toprule
		\multicolumn{2}{c|}{{models}} &{ogbl-biokg} & {ogbl-wikikg2} \\ \midrule
		&        TransE        & 0.7452      & 0.4256          \\
		&        RotatE        & 0.7989 &       0.2530       \\
		original         &       DistMult       & 0.8043     & 0.3729     \\ 
		&       ComplEx        & 0.8095  & 0.4027    \\   
		&       AutoSF        &  0.8320   &   0.5186  \\   \midrule
		&        TransE        & 0.7781 (4.41\%$\uparrow$) &   0.4739 (11.34\%$\uparrow$)     \\
		&        RotatE        & 0.8013 (0.30\%$\uparrow$)   & 0.2944  (16.36\%$\uparrow$)      \\
		KGTuner &       DistMult       & 0.8241 (2.46\%$\uparrow$)   & 0.4837 (29.71\%$\uparrow$)  \\
		&       ComplEx        & 0.8385 (3.58\%$\uparrow$)  & 0.4942 (22.72\%$\uparrow$)    \\ 
		&       AutoSF        &   0.8354 (0.41\%$\uparrow$)  &  0.5222 (0.69\%$\uparrow$)  \\   
		\midrule
		\multicolumn{2}{c|}{average improvement} &  2.23\%  &  16.16\%  \\
		\bottomrule
	\end{tabular}
	\vspace{-4px}
\end{table}

\subsection{Ablation study}
\label{ssec:exp:abla}
In this subsection, we probe into how important and 
sensitive
the various components of KGTuner are.

\vspace{2px}
\noindent
\textbf{Space comparison.}
To demonstrate the effectiveness gained by shrinking and decoupling the search space,
we compare the following variants:
(i) RF+BORE on the full space $\mathcal X$;
(ii) RF+BORE on the shrunken space ${\mathcal X}_\text{S}$;
(iii) RF+BORE on the decoupled space ${\mathcal X}_\text{S|D}$,
which differs from KGTuner by searching on the full graph in the first stage;
and 
(iv) KGTuner in Algorithm~\ref{alg:search}.
All the variants, 
i.e., RF+BORE,
have one day's time budget.
As in Figure~\ref{fig:ablation_space},
the size of search space matters
for the search efficiency.
The three components,
i.e., space shrinkage,
space decoupling,
and fast evaluation on subgraph,
are all important to the success of KGTuner.

\vspace{2px}
\noindent
\textbf{Size of subgraphs.}
We show the influence of subgraph sizes
with
different ratios of entities 
(10\%, 20\%, 30\%, 40\%, 50\%)
from the full graph
in 
Figure~\ref{fig:ablation_subgraph_size}.
Using subgraphs
with too large or too small size
is not guaranteed
to find good configurations.
Based on the understanding in Figure~\ref{fig:subgraph_correlation},
the subgraph with small size have poor transfer ability
and 
those with large size are expensive to evaluate.
Hence,
we should balance the transfer ability and evaluation cost
by sampling subgraphs with $20\%\sim 30\%$ entities.

\vspace{2px}
\noindent
\textbf{Budget allocation.}
In Algorithm~\ref{alg:search},
we allocate $\nicefrac{B}{2}$ budget for both the first and second stage.
Here,
we show the performance of different allocation ratios,
i.e.,
$\nicefrac{B}{4}$, $\nicefrac{B}{2}$, and $\nicefrac{3B}{4}$ in the first stage
and the remaining budget in the second stage.
As in Figure~\ref{fig:ablation_tradeoff},
allocating too many or too few budgets
to the first stage is not good.
It either fails to explore sufficient configurations in the first stage
or only fine-tunes a few configurations in the second stage.
Allocating the same budget to the two stages
is in a better trade-off.

%

\section{Related works}
\label{sec:relworks}

In analyzing the performance of KG embedding models,
\citet{ruffinelli2019you} pointed out that the earlier works in KG embedding
only search HPs in small grids.
By searching hundreds of HPs in a unified framework,
the reproduced performance can be significantly improved.
Similarly,
\citet{ali2020bringing} proposed another unified framework
to evaluate different models.
\citet{rossi2020knowledge} evaluated 16 different models
and analyzed their properties on different datasets.
All of these works emphasize the importance of HP search,
but none of them provide efficient algorithms
to search HPs for KG learning.
AutoSF \citep{zhang2020autosf} evaluates
the bilinear scoring functions
and set up a search problem to design
bilinear scoring functions,
which can be complementary to KGTuner.

Understanding the HPs in a large search space
is non-trivial since many HPs only have moderate impact on the model performance
\cite{ruffinelli2019you}
and jointly evaluating them requires
a large number of experiments \citep{fawcett2016analysing,probst2019tunability}.
Considering the huge amount of HP configurations
(with $10^{5}$ categorical choices and $5$ continuous values),
it is extremely expensive to 
exhaustively evaluate most of them.
Hence, we adopt control variate experiments
to efficiently evaluate HPs' properties
instead of the quasi-random search in \citep{ruffinelli2019you,ali2020bringing}.

Technically,
we are similar to
AutoNE \citep{tu2019autone} and e-AutoGR \citep{wang2021explainable}
by leveraging subgraphs to improve search efficiency on graph learning.
Since they do not target at KG embedding methods,
directly adopt them is not a good choice.
Besides,
based on the understanding in this paper,
we demonstrate that
transferring the surrogate model from subgraph evaluation to the full graph 
is inferior to
directly transferring the top configurations for KG embedding models.

\section{Conclusion}

In this paper,
we 
analyze the HPs' properties in KG embedding models
with search space size,
validation curvature and evaluation cost.
Based on the observations,
we propose an efficient search algorithm KGTuner
that 
efficiently explores configurations
in a reduced space on small subgraph
and then fine-tunes the top configurations
with increased batch size and dimension size
on the full graph.
Empirical evaluations show
that KGTuner is robuster and more efficient
than the existing HP search algorithms
and achieves competing performance 
on large-scale KGs in open graph benchmarks.
In the future work,
we will 
understand
the HPs in 
graph neural network based models
and apply KGTuner on them 
to solve the scaling limitations
in HP search.


\section*{Acknowledgements}

This work was supported in part by The National Key Research and Development Program of China under grant 2020AAA0106000.

\clearpage
\bibliographystyle{custom}
\bibliography{bib}

\clearpage
\appendix

\onecolumn

\section{Details of the search space}
\label{app:searchspace}

Denote a knowledge graph as $\mathcal G = \{E, R, D \}$,
where $E$ is the set of entities,
$R$ is the set of relations,
and $D$ is the set of triplets
with training/validation/test splits 
$D = D_{\text{tra}}\cup D_{\text{val}}\cup D_{\text{tst}}$.

Basically,
the KG embedding models
use a scoring function $f$ and the model parameters $\bm P$
to measure the plausibility of triplets.
We learn the embeddings such that the positive and negative triplets
can be separated by $f$ and $\bm P$.
In Table~\ref{tab:sfdefinition},
we provide the forms $f$ of the embedding model we used to evaluate the search space $\mathcal X$
in Section~\ref{sec:understanding}.

\begin{table*}[ht]
	\centering
	\caption{Definitions of the embedding models.
	 $\circ$ is a rotation operation in the complex value space;
 	$\otimes$ is the Hermitian dot product in the complex value space;
 	$\text{Re}(\cdot)$ returns the real part of a complex value;
 	$\mathcal W_{i,j,k}$ is the $ijk$-th element in a core tensor $\mathcal W\in\mathbb R^{d\times d\times d}$;
 	and conv is a convolution operator on the head and relation embeddings.
 	For more details, please refer to the corresponding references.}
	\vspace{-8px}
	\label{tab:sfdefinition}
	\small
	\begin{tabular}{cc|c|c}
		\toprule
		model type  &  model  &  $f(h,r,t)$  & embeddings  \\  \midrule
		\multirow{2}{*}{translational distance}&TransE \citep{bordes2013translating}  &  $-\|\bm h+\bm r-\bm t\|_1$  & $\bm h, \bm r, \bm t \in\mathbb R^d$ \\
		&RotatE \citep{sun2019rotate}   &  $ -\|\bm h \circ \bm r-\bm t\|_{c1}$  &  $\bm h, \bm r, \bm t \in\mathbb C^d$ \\ \midrule
		\multirow{4}{*}{tensor factorization} &RESCAL \citep{nickel2011three}  & $\bm h^\top \cdot\bm R_r\cdot\bm t$  & $\bm h, \bm t\in\mathbb R^d, \bm R_r\in\mathbb R^{d\times d}$ \\
		&DistMult \citep{yang2014embedding}  &  $\bm h^\top \cdot\text{diag}(\bm r)\cdot\bm t$  & $\bm h, \bm t, \bm r\in\mathbb R^d$ \\
		&ComplEx \citep{trouillon2017knowledge} &  $\bm h^\top \otimes \text{diag}({\bm r})\otimes \bm t$  & $\bm h, \bm t, \bm r\in\mathbb C^d$ \\
		&TuckER \citep{balavzevic2019tucker} &  $\sum_{i}^d\sum_j^d\sum_k^d \mathcal W_{i,j,k}h_i\cdot r_j \cdot t_k$ & $\bm h, \bm t, \bm r\in\mathbb R^d$ \\	\midrule
		neural network &ConvE \citep{dettmers2017convolutional} & $\text{ReLU}(\text{conv}(\bm h, \bm r))^\top \cdot \bm t$  &  $\bm h, \bm t, \bm r\in\mathbb R^d$ \\
		\bottomrule
	\end{tabular}
\end{table*}

\subsection{Negative sampling}
Since KG only contains positive triplets in $D_{\text{tra}}$
\citep{wang2017knowledge},
we should rely on the negative sampling to avoid trivial solutions of the embeddings.
Given a positive triplet $(h,r,t)\in D_{\text{tra}}$,
the corresponding set of negative triplets
is represented as 
\begin{align*}
D^-_{(h,r,t)} = \big\{(\tilde{h}, r, t)\notin D_{\text{tra}}\!:\! (h,r,t)\in D_{\text{tra}}, \tilde{h}\in E \big\}\cup \big\{({h}, r, \tilde{t})\notin D_{\text{tra}}\!:\! (h,r,t)\in D_{\text{tra}}, \tilde{t}\in E \big\}. 
\end{align*}
A common practice is to sample $m$ negative triplets from $D^-_{(h,r,t)}$.
The value of $m$ can be any integer smaller than the number of entities.
We follow \citep{sun2019rotate} to sample from the range of $m$ in $\{32,128,512,2048\}$
for simplicity.

An alternative choice is to use all the negative triplets in $D^-_{(h,r,t)}$,
leading to the \texttt{1VsAll} \citep{lacroix2018canonical} and \texttt{kVsAll} \citep{dettmers2017convolutional}
settings.
\begin{itemize}[leftmargin=*]
	\item In \texttt{1VsAll}, $(h,r,t)$ is in the positive part and all the triplets in the set $\{(\tilde{h}, r, t)\notin D_{\text{tra}}\!:\! (h,r,t)\in D_{\text{tra}}, \tilde{h}\in E \}$
	or $\{({h}, r, \tilde{t})\notin D_{\text{tra}}\!:\! (h,r,t)\in D_{\text{tra}}, \tilde{t}\in E \}$
	are in the negative part;
	\item In \texttt{kVsAll}, the positive part contains all the triplets sharing the same head-relation pair or tail-relation part,
	i.e. $\{(h,r,t')\in D_{\text{tra}} \}$	or $\{(h',r,t)\in D_{\text{tra}} \}$,
	with the corresponding negative part $\{({h}, r, \tilde{t})\notin D_{\text{tra}}\!:\! (h,r,t)\in D_{\text{tra}}, \tilde{t}\in E \}$
	or $\{(\tilde{h}, r, t)\notin D_{\text{tra}}\!:\! (h,r,t)\in D_{\text{tra}}, \tilde{h}\in E \}$.
\end{itemize}

Hence, the choice of negative sampling
can be set in the range $\{32, 128, 512, 2048, \text{\texttt{1VsAll}}, \text{\texttt{kVsAll}} \}$.

\subsection{Loss function}
For simplicity,
we denote $D^+$ and $D^-$ as the sets of positive and negative triplets, respectively.
Then,
we summarize the commonly used loss functions as follows:
\begin{itemize}[leftmargin=*]
	\item Margin ranking (MR) loss. 
	This loss ranks the positive triplets to have larger score than the negative triplets. 
	Hence, the ranking loss is defined as 
	\[ \mathcal L 
	= \sum\nolimits_{(h,r,t)\in D^+}
	\sum\nolimits_{(\tilde{h},r, \tilde{t})\in D^-} 
	-\big|\gamma-f(h,r,t) + f(\tilde{h}, r, \tilde{t})\big|_+, \]
	where $\gamma>0$ is the margin value and $|a|_+ = \max(a, 0)$.
	The MR loss is widely used in early developed models, 
	like TransE \citep{bordes2013translating} and DistMult \citep{yang2014embedding}.
	The value of  $\gamma$, conditioned on MR loss, is another HP to search.
	\item Binary cross entropy (BCE) loss.
	It is typical to set the positive and negative triplets as a binary classification problem.
	Let the labels for the positive and negative triplets as $+1$ and $-1$ respectively,
	the BCE loss is defined as
	\begin{align*}
	\mathcal L = 
	\sum\nolimits_{(h,r,t)\in D^+} \log\big(\sigma(f(h,r,t))\big) + 
	\sum\nolimits_{(\tilde{h},r, \tilde{t})\in D^-}\!\! w_{(\tilde{h},r, \tilde{t})}  \log\big(1-\sigma(f(\tilde{h},r,\tilde{t}))\big),
	\end{align*}
	where $\sigma(x)=\frac{1}{1+\exp(-x)}$ is the sigmoid function.
	The choice of  $w_{(\tilde{h},r, \tilde{t})}$ leads to three different loss functions
	\begin{itemize}[leftmargin=*]
		\item BCE\_mean \citep{sun2019rotate}, with $w_{(\tilde{h},r, \tilde{t})}= \nicefrac{1}{|D^-_{(h,r,t)}|}$.
		\item BCE\_sum \citep{dettmers2017convolutional}, with $w_{(\tilde{h},r, \tilde{t})}= 1$.
		\item BCE\_adv \citep{sun2019rotate}, with 
		\[w_{(\tilde{h},r, \tilde{t})}= \frac{\exp(\alpha\cdot f(\tilde{h},r,\tilde{t}))}{\sum_{(h',r,t')\in D^-}\exp(\alpha \cdot f({h'},r,{t'}))},\]
		where $\alpha>0$ is the adversarial weight conditioned on BCE\_adv loss.
	\end{itemize}
	\item Cross entropy (CE) loss. 
		Since the number of negative triplets is fixed, 
		we can also regard the $(h,r,t)$ as the true label over the negative ones.
		The loss can be written as
		\begin{align*}
		\mathcal L =  
		\sum\nolimits_{(h,r,t)\in D^+} 
		- f(h,r,t) 
		+ 
		\log
		\left( 
		\sum\nolimits_{(h',r,t')\in \{(h,r,t)\cup  D^-\}} \exp (f(h',r,t'))
		\right) ,
		\end{align*}
		where the left part is the score of positive triplet and the right is the log sum scores of the joint set of positive and negative triplets.
\end{itemize}

\subsection{Regularization}
To avoid the embeddings increasing to unlimited values
and reduce the model complexity,
regularization techniques are often used.
Denote $\bm P'$
as the embeddings participated in one iteration,
\begin{itemize}[leftmargin=*]
	\item the Frobenius norm is defined as the sum of L2 norms 
	$r_{\text{FRO}}= \|\bm P'\|_2^2 = \sum_{ij}{P'}_{ij}^2$ \citep{yang2014embedding};
	\item the NUC norm is defined as sum of L3 norms 
	$r_{\text{FRO}}= \|\bm P'\|_3^3 = \sum_{ij}|P_{ij}|^3$ \citep{lacroix2018canonical};
	\item DURA operates on triplets \citep{zhang2020duality}.
	Denote $\bm h, \bm r, \bm t$ as the embeddings for the triplet $(h,r,t)$,
	DURA constrains the composition of $\bm h$ and $\bm r$ to approximate $\bm t$
	with $r_{\text{DURA}} = \|c(\bm h,\bm r)- \bm t\|_2^2$,
	where the composition function $c(\bm h,\bm r)$ depends on corresponding scoring functions.
\end{itemize}
The regularization functions are then weighted by the regularization weight in the range $[10^{-12}, 10^2]$.

Apart from using explicit forms of regularization,
we can also add dropout on the embeddings \citep{dettmers2017convolutional}.
Specifically,
each dimension in the embeddings $\bm h, \bm r, \bm t$
will have a probability to be deactivated as $0$ in each iteration.
The probability is controlled by the dropout rate in the range $[0, 0.5]$.
In some cases,
working without regularization can also achieve good performance
\citep{ali2020bringing}.

\subsection{Optimization}
To solve the learning problem,
we should setup an appropriate optimization procedure.
First, we can directly use the training set
or add inverse relations to augment the data \citep{kazemi2018simple,lacroix2018canonical}.
This will not influence the training data,
but will introduce additional parameters for the inverse relations.
Second,
we should choose the dimension of embeddings
in small sizes $[100, 200]$ or large sizes $[500, 1000, 2000]$.
Then,
the embeddings are initialized by the 
initialization methods such as
uniform, normal, xavier\_norm, and xavier\_uniform \citep{goodfellow2016deep}.
The optimization is conducted
with optimizers like
standard SGD, Adam \citep{kingma2014adam} and Adagrad \citep{duchi2011adaptive}
with learning rate in the range $[10^{-5}, 0]$
Since the training is conducted on mini-batch,
a batch size is determined in the range $\{128, 256, 512, 1024\}$.

\section{Details of HP understanding}
In this part,
we provide the details of configuration generation
and the full results related to the HP understanding.

\subsection{Configure generation}
\label{app:configgen}

Since there are infinite numbers of values
for a continuous HP,
it is intractable to fully evaluate their ranges.
To better analyze the continuous HPs,
we discretize them in Table~\ref{tab:discrete}
according to their ranges.
Then,
for each HP $i=1\dots n$ with range $X_i$,
we sample a set ${\mathcal X}_{i} \subset \mathcal X$ of
$s$ anchor configurations  
through quasi random search \citep{bergstra2012random}
and uniformly distribute them to evaluate the different embedding models
and datasets.

\begin{table}[ht]
	\centering
	\caption{Discretized HP values.}
	\label{tab:discrete}
	\vspace{-8px}
	\small
	\begin{tabular}{c|c|c}
		\toprule
		name &  original range & discretized range   \\
		\midrule
		gamma &[1, 24]  &   $\{1, 6,12,24\}$      \\ 
		adv. weight  &   [0.5, 2.0]    &  \{0.5, 1, 2\}\\
		reg. weight &   [$10^{-12}\!, 10^{2}$]    &
		$10^2$ in log scale     \\
		dropout rate  &   $[0, 0.5]$  &   $0.1$ in linear scale      \\
		learning rate &  [$10^{-5}\!, 10^0$]   & $10^1$ in log scale         \\
		\bottomrule
	\end{tabular}
\end{table}

We use the control variate experiments to evaluate each HP.
For the $i$-th HP,
we enumerate the values $\theta\in X_i$ 
for each anchor configuration $\mathbf x\in{\mathcal X}_i$,
while fix the other HPs.
In this way,
we can observe the influence of $x_i$ without the influence of the other HPs.
For example, when evaluating the optimizers,
we  enumerate the optimizers Adam, Adagrad and SGD
for the anchor 
configurations in ${\mathcal X}_i$.
This generates a set of
$|\mathcal X_i| \cdot |X_i|$
configurations.
In this paper,
the number of anchor configurations $|\mathcal X_i|$
is 175 for each HP.

\subsection{Details for search space understanding}
\label{app:under:space}

In this part,
we add the ranking distribution of all the HPs.
In addition,
we also show the normalized MRR of each HP as a complementary.
The normalization is conducted on each dataset 
with $\frac{y-y_{\min}}{y_{\max}-y_{\min}}$
such that the results of the HPs
can be evaluated in the same value range.

\begin{figure}[ht]
	\subfigure[optimizer]{\includegraphics[width=0.24\textwidth]{figures/HP_understanding_optimizer_rank}
		\includegraphics[width=0.24\textwidth]{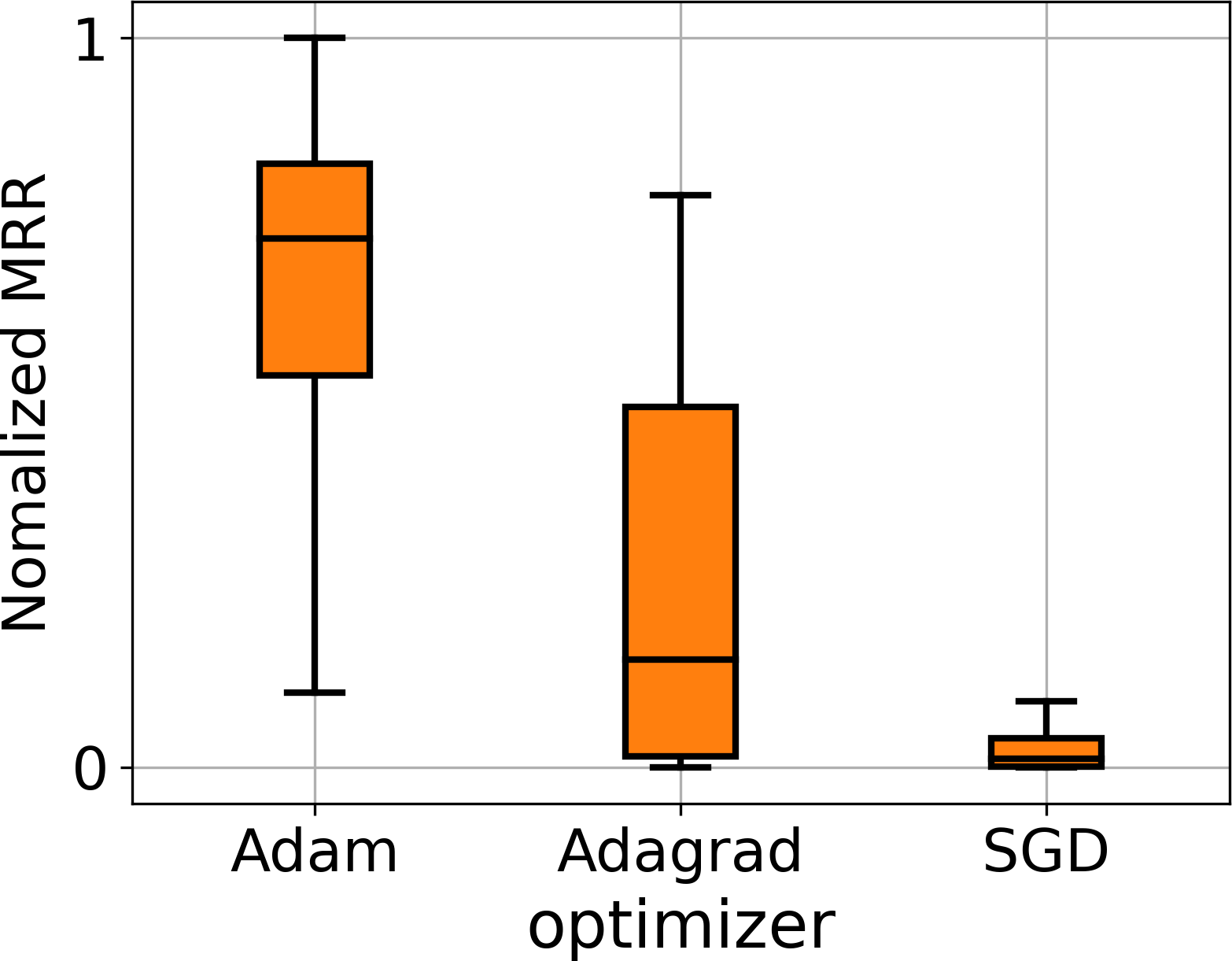}}\hfill
		\subfigure[add inverse relation]{\includegraphics[width=0.24\textwidth]{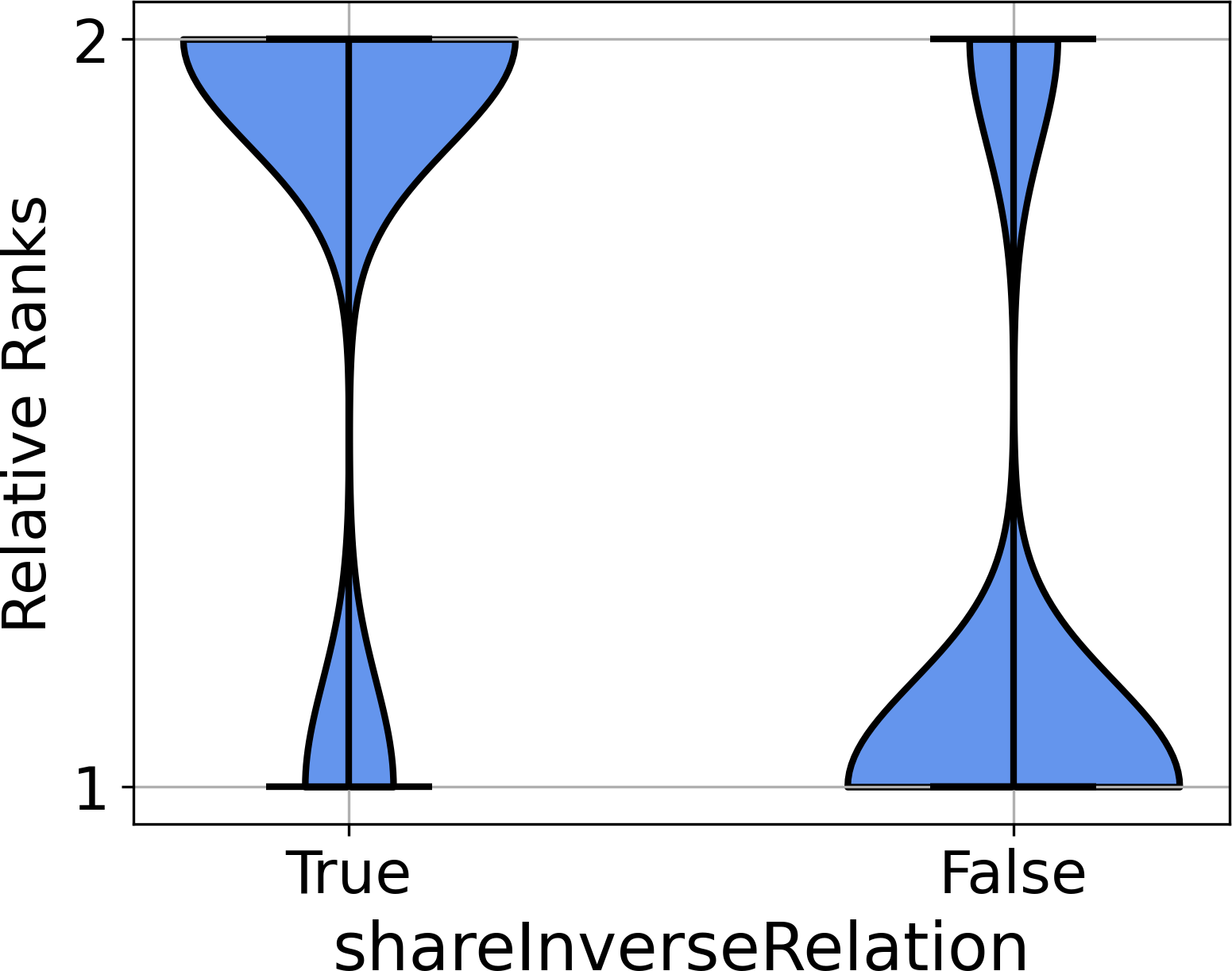}
		\includegraphics[width=0.24\textwidth]{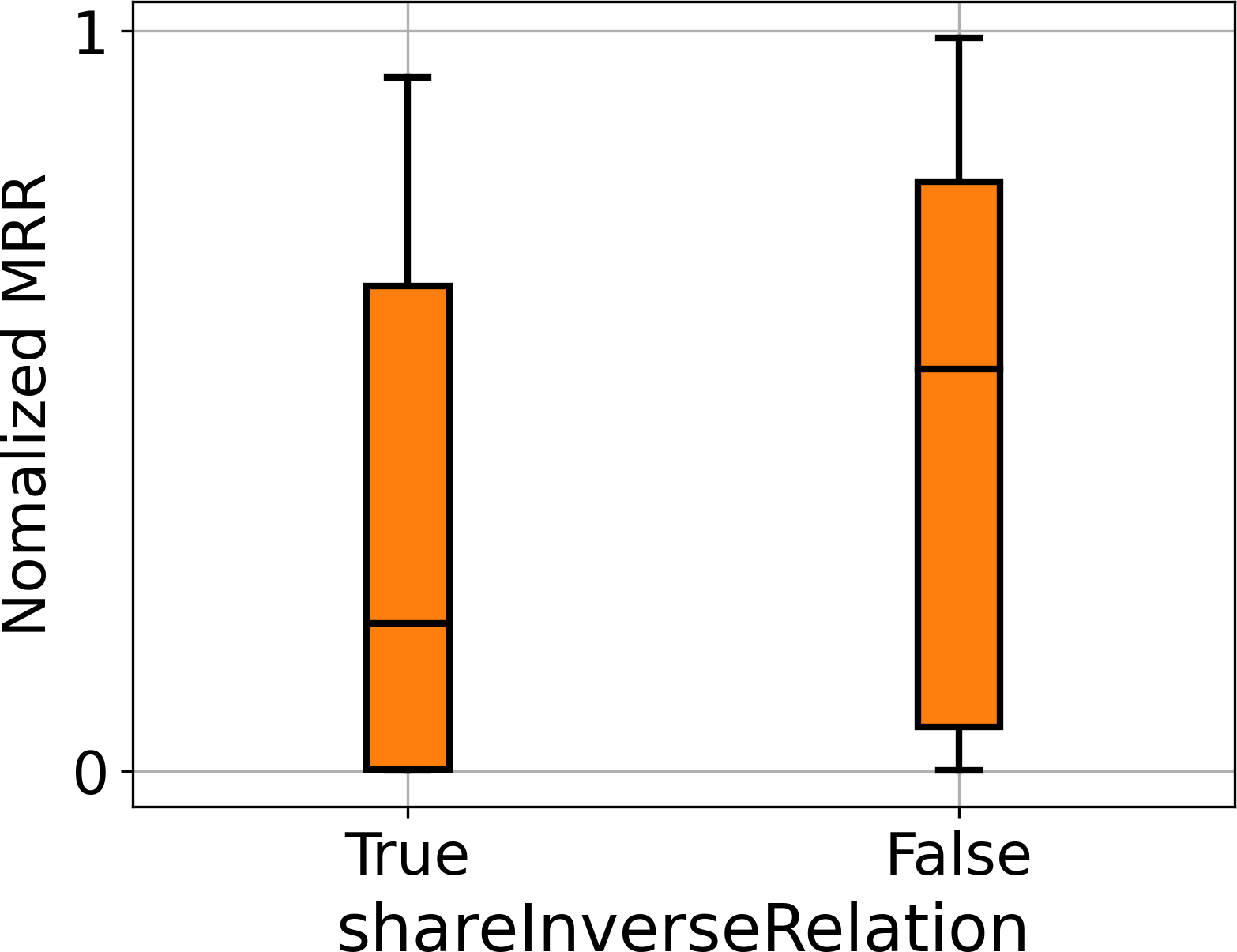}}
	\vspace{-11px}
	\caption{HPs that have fixed choice since one configure has significant advantage.}
	\label{fig:hp:fixed}
\end{figure}

\begin{figure}[ht]
		\subfigure[learning rate]{\includegraphics[width=0.24\textwidth]{figures/HP_understanding_lr_rank}
		\includegraphics[width=0.24\textwidth]{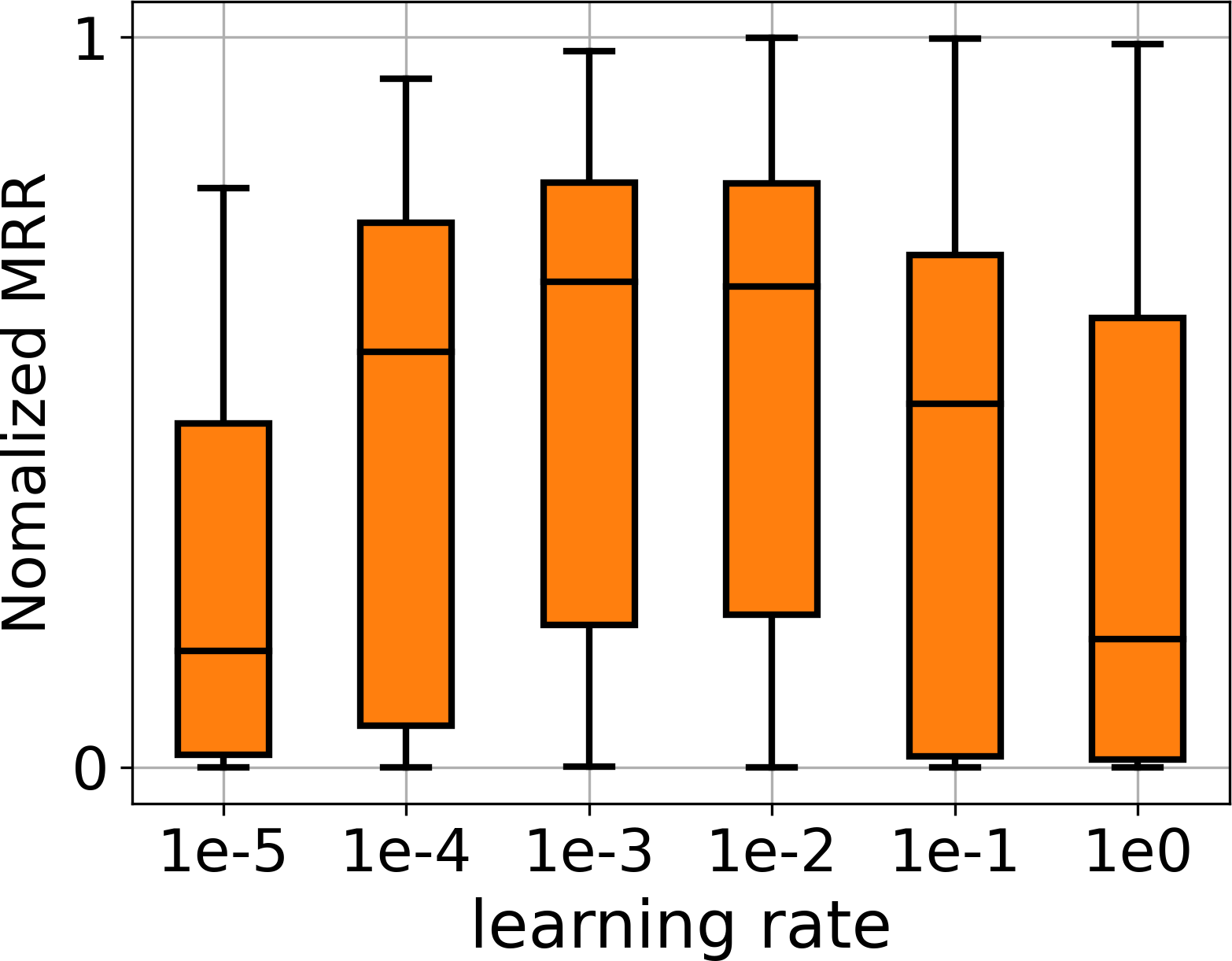}}\hfill
		\subfigure[reg. weight]{\includegraphics[width=0.24\textwidth]{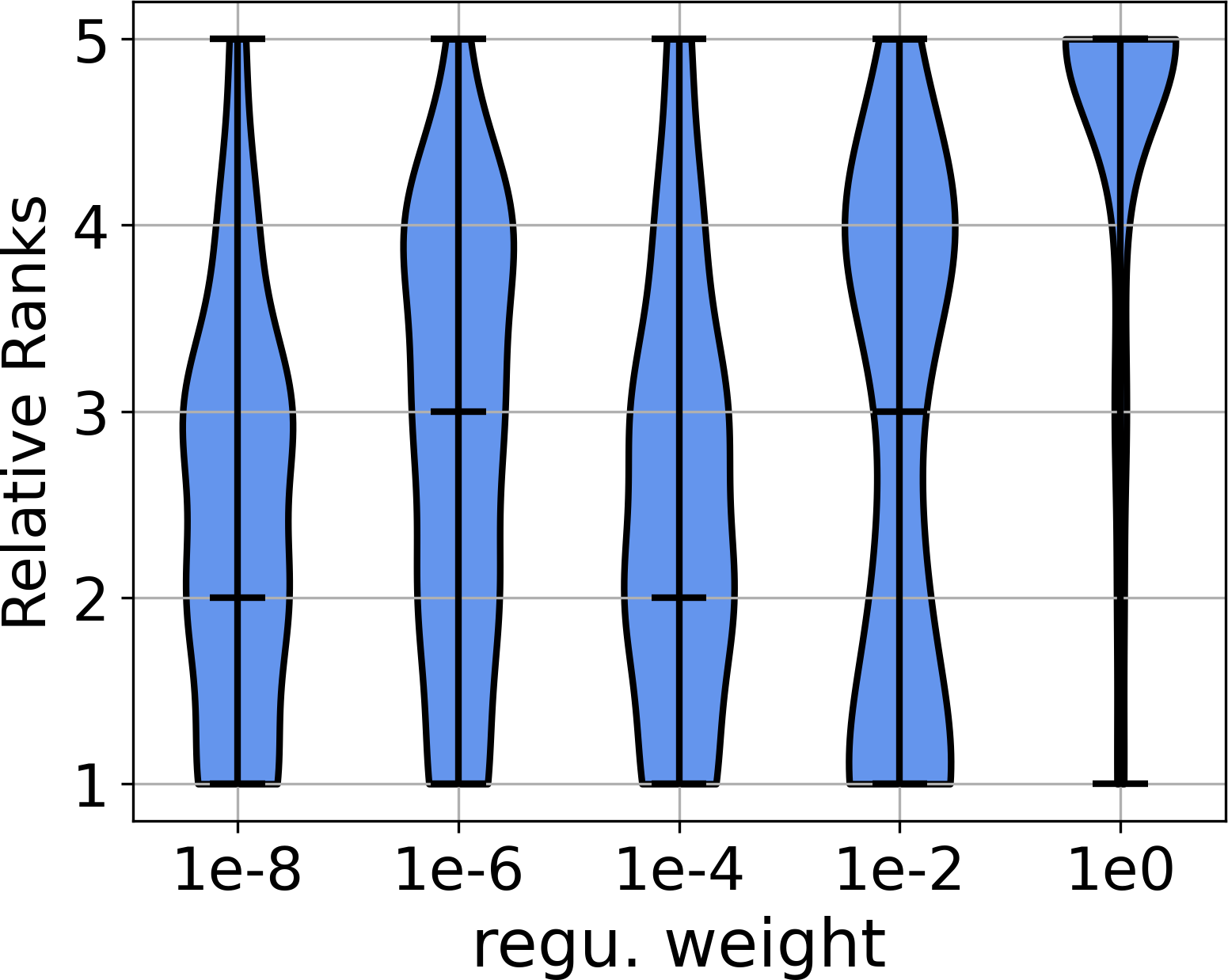}
		\includegraphics[width=0.24\textwidth]{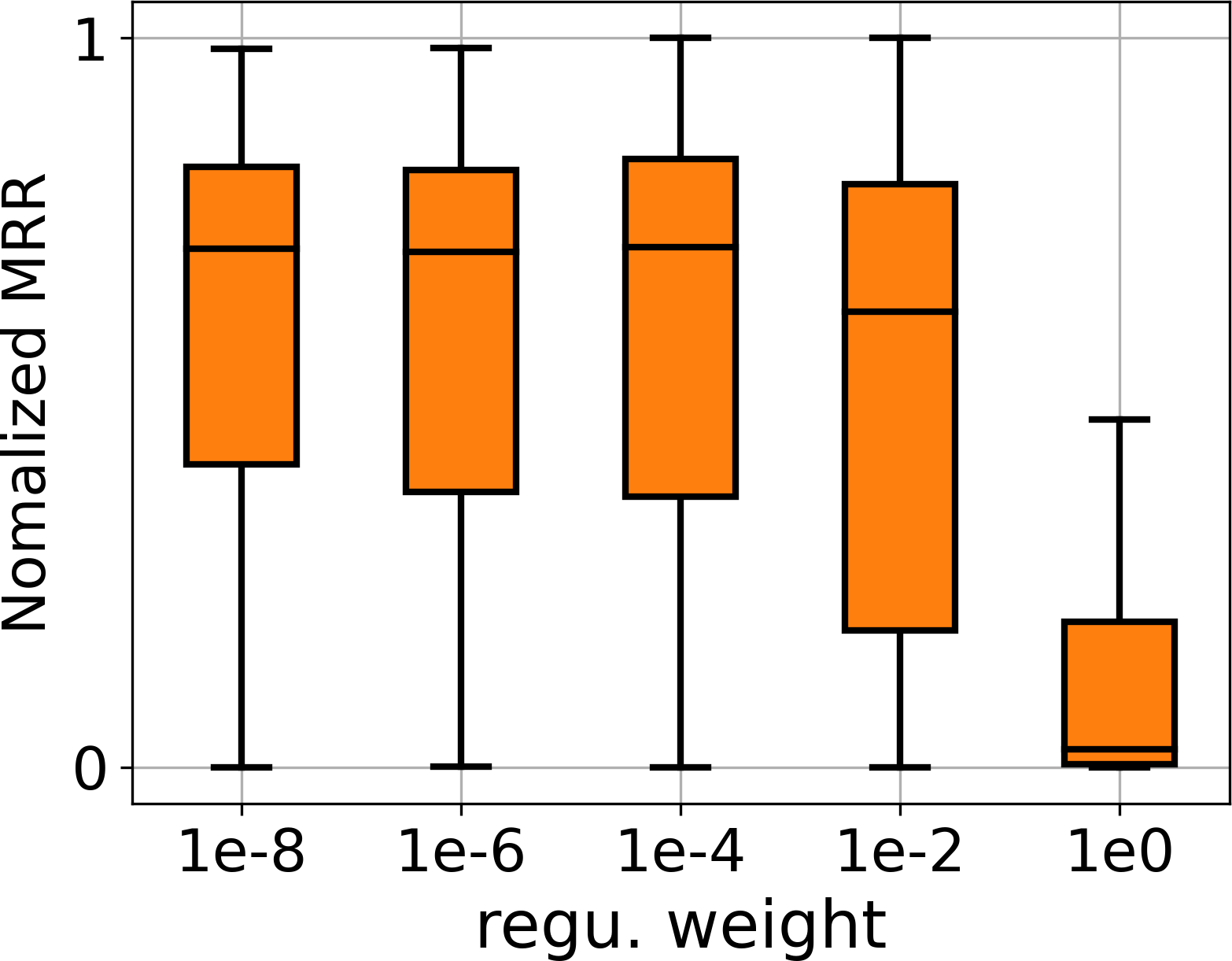}}
	
	\subfigure[dropout rate]{\includegraphics[width=0.24\textwidth]{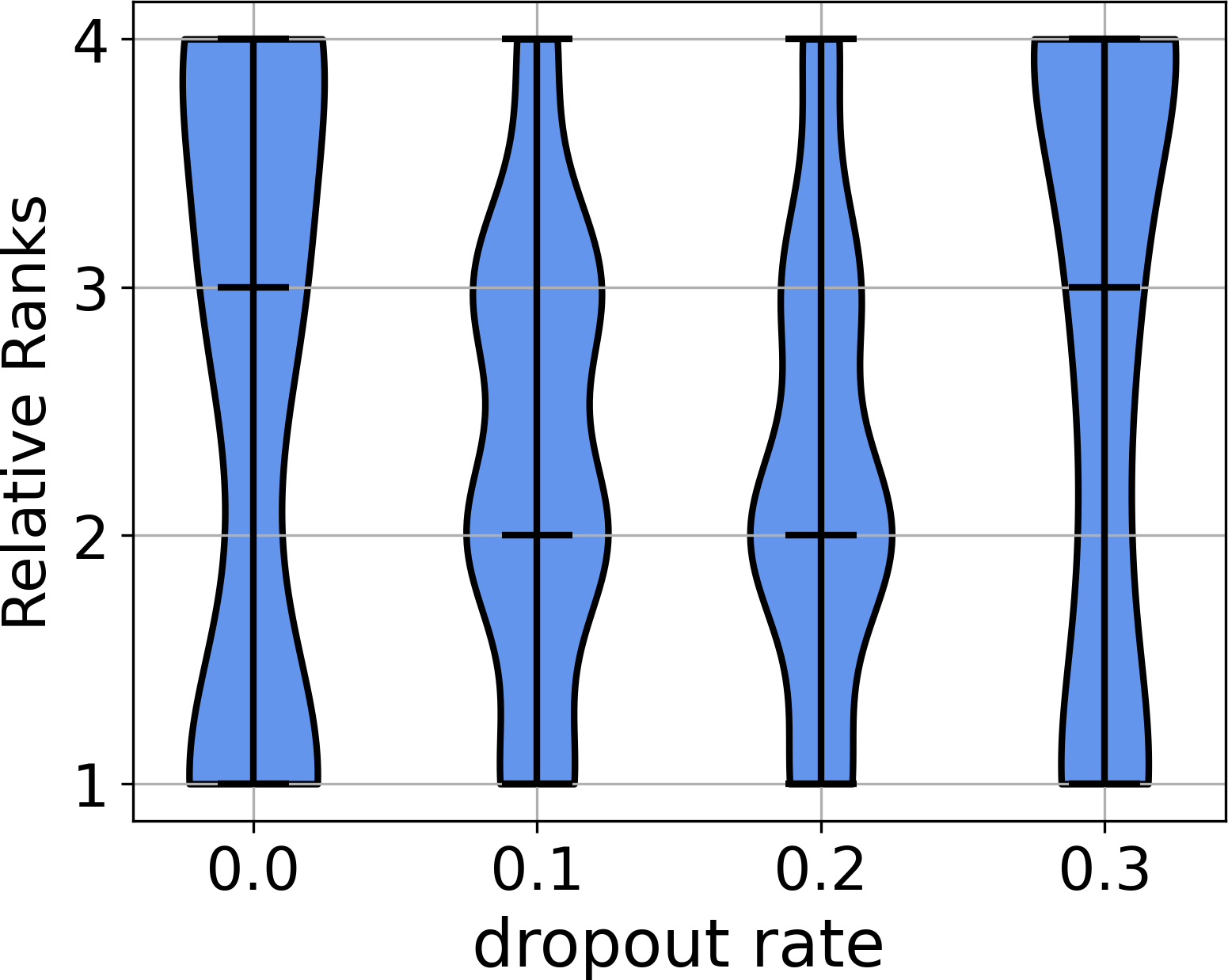}
		\includegraphics[width=0.24\textwidth]{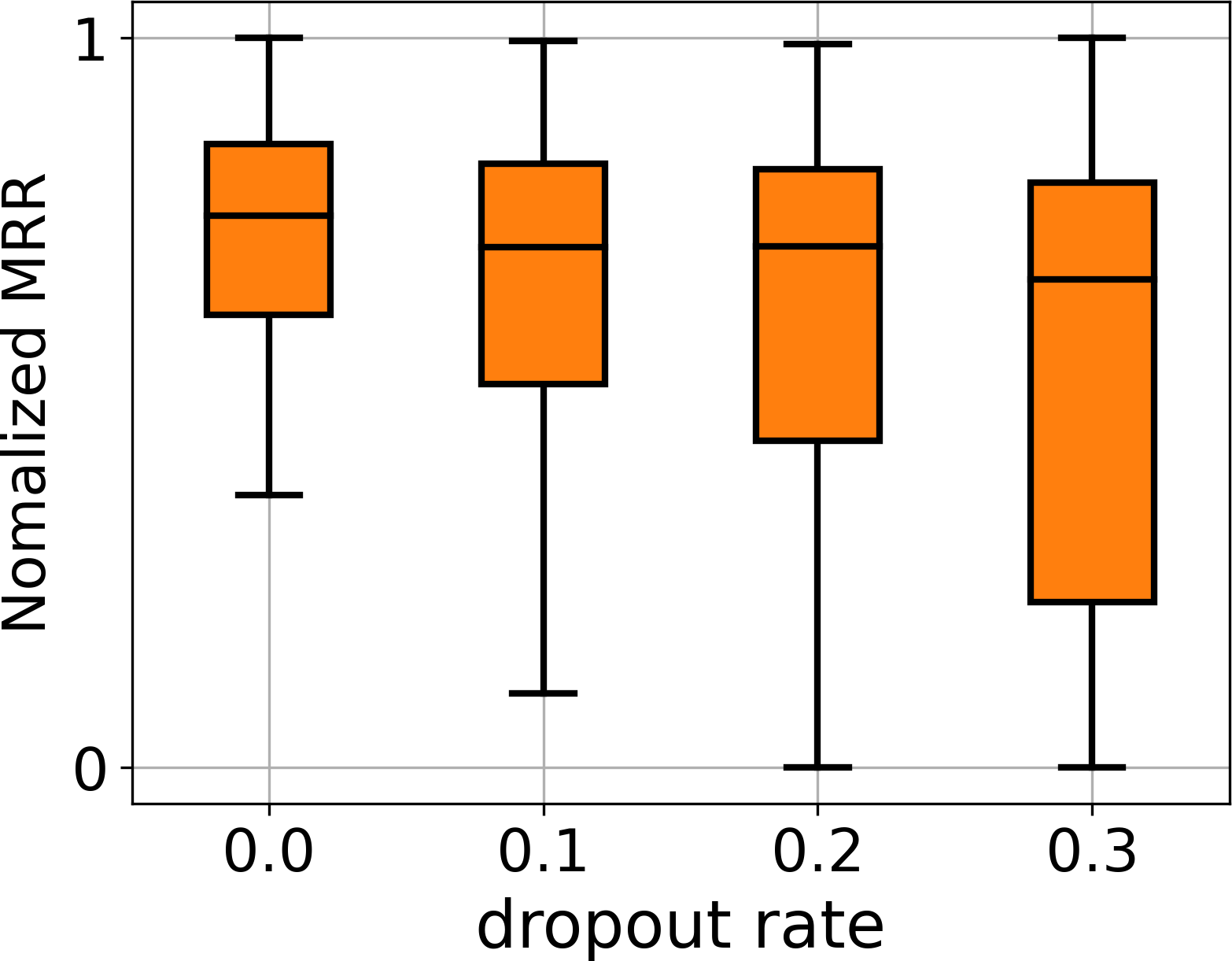}}
	\vspace{-11px}
	\caption{HPs that have limited ranges since they only perform well in certain ranges.}	
	\label{fig:hp:limited}
\end{figure}

\begin{figure}[ht]
		\subfigure[batch\_size]{\includegraphics[width=0.24\textwidth]{figures/HP_understanding_batch_size_rank}
		\includegraphics[width=0.24\textwidth]{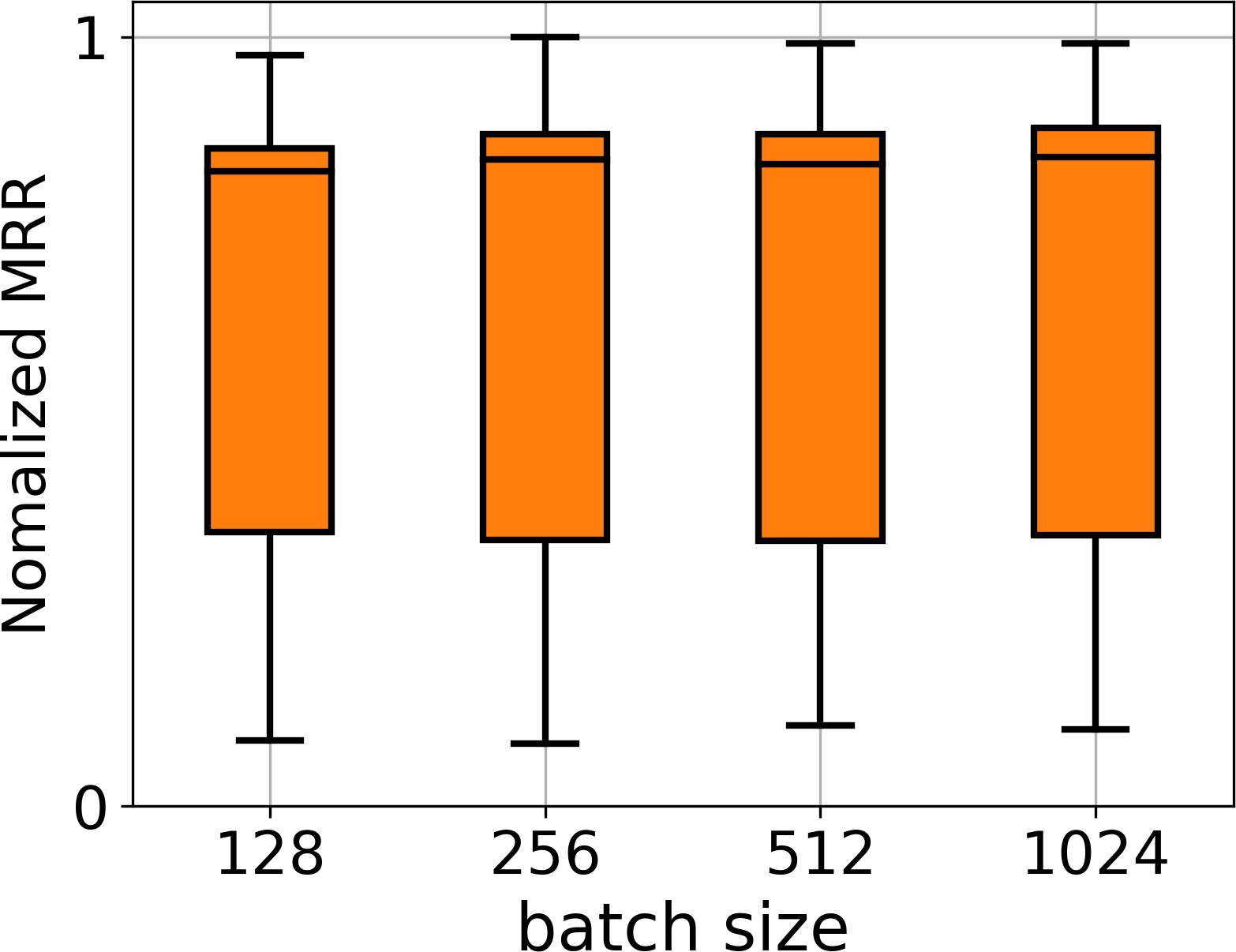}} \hfill
		\subfigure[dimension]{\includegraphics[width=0.24\textwidth]{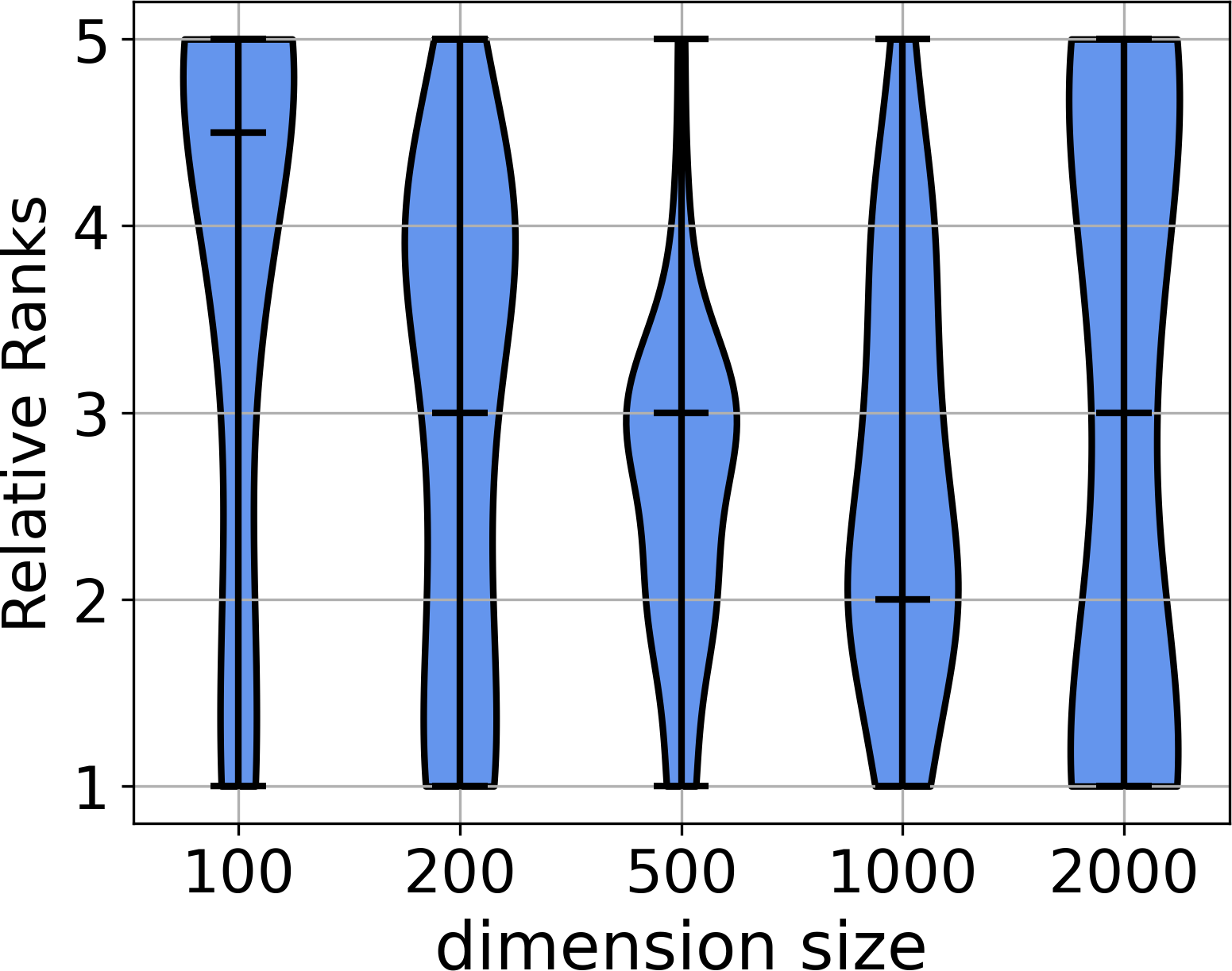}
		\includegraphics[width=0.24\textwidth]{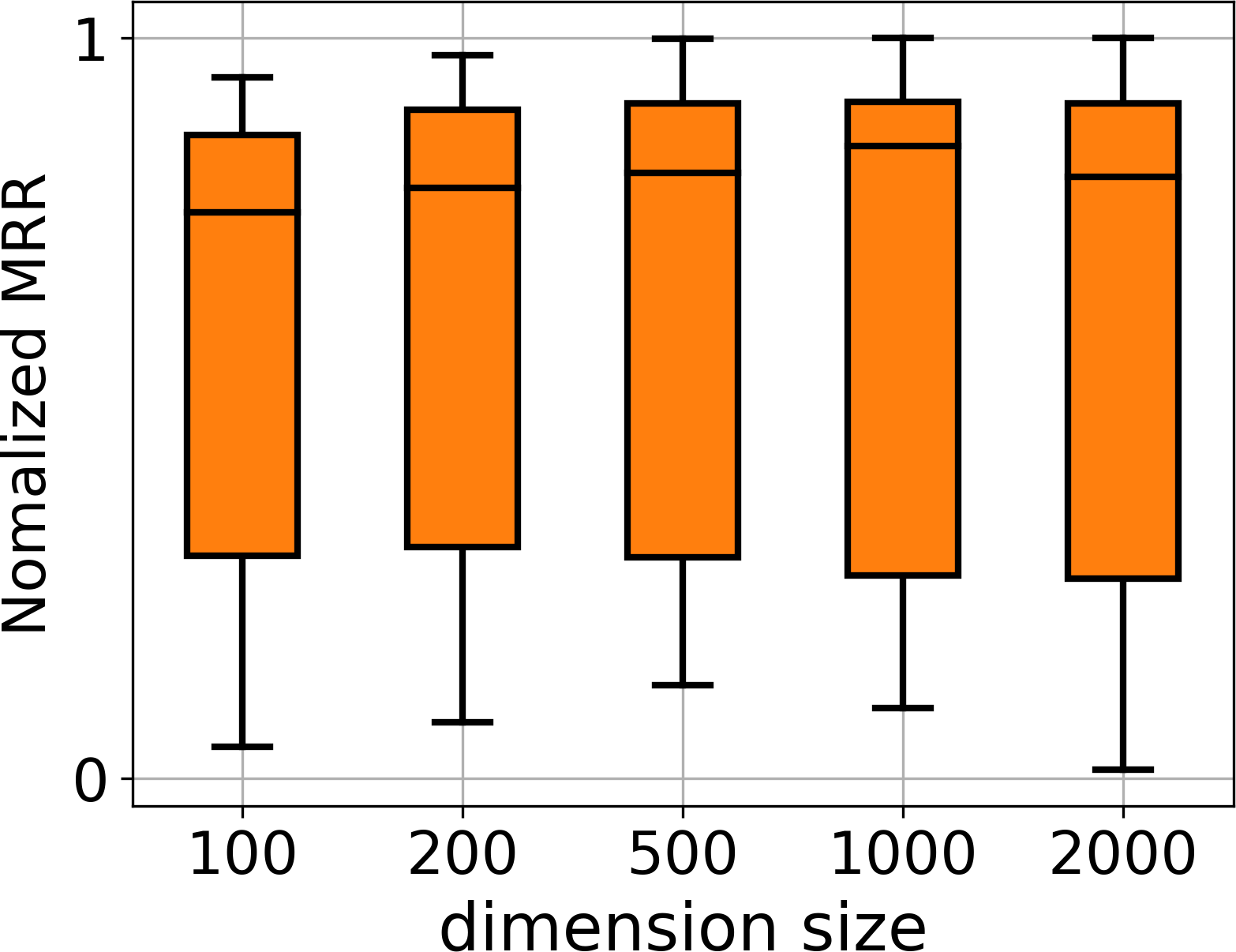}}
	\vspace{-11px}
	\caption{HPs that is monotonic with different choices of values.}	
	\label{fig:hp:monotonic}
\end{figure}

\begin{figure}[ht]
	\subfigure[loss]{\includegraphics[width=0.243\textwidth]{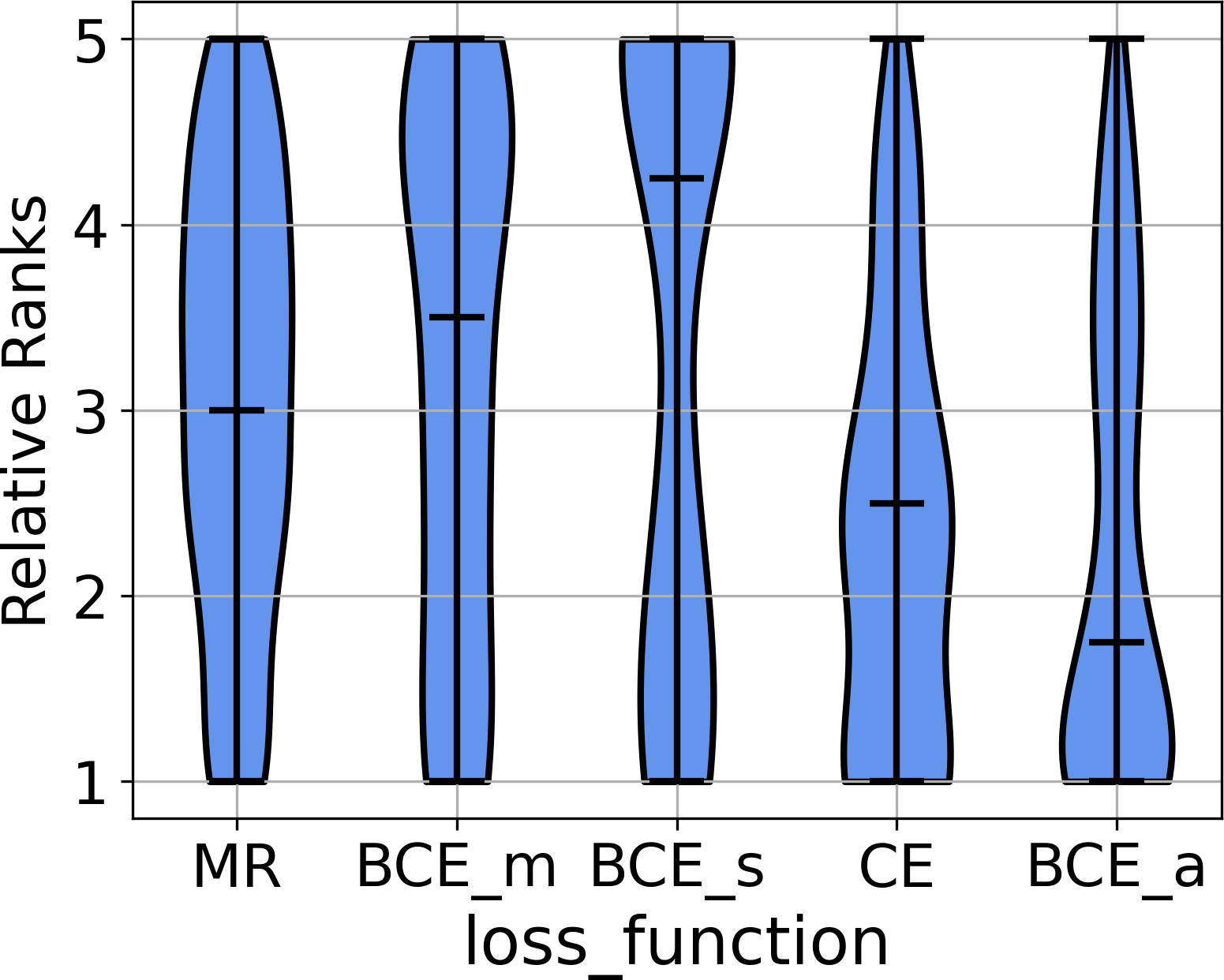}
		\includegraphics[width=0.243\textwidth]{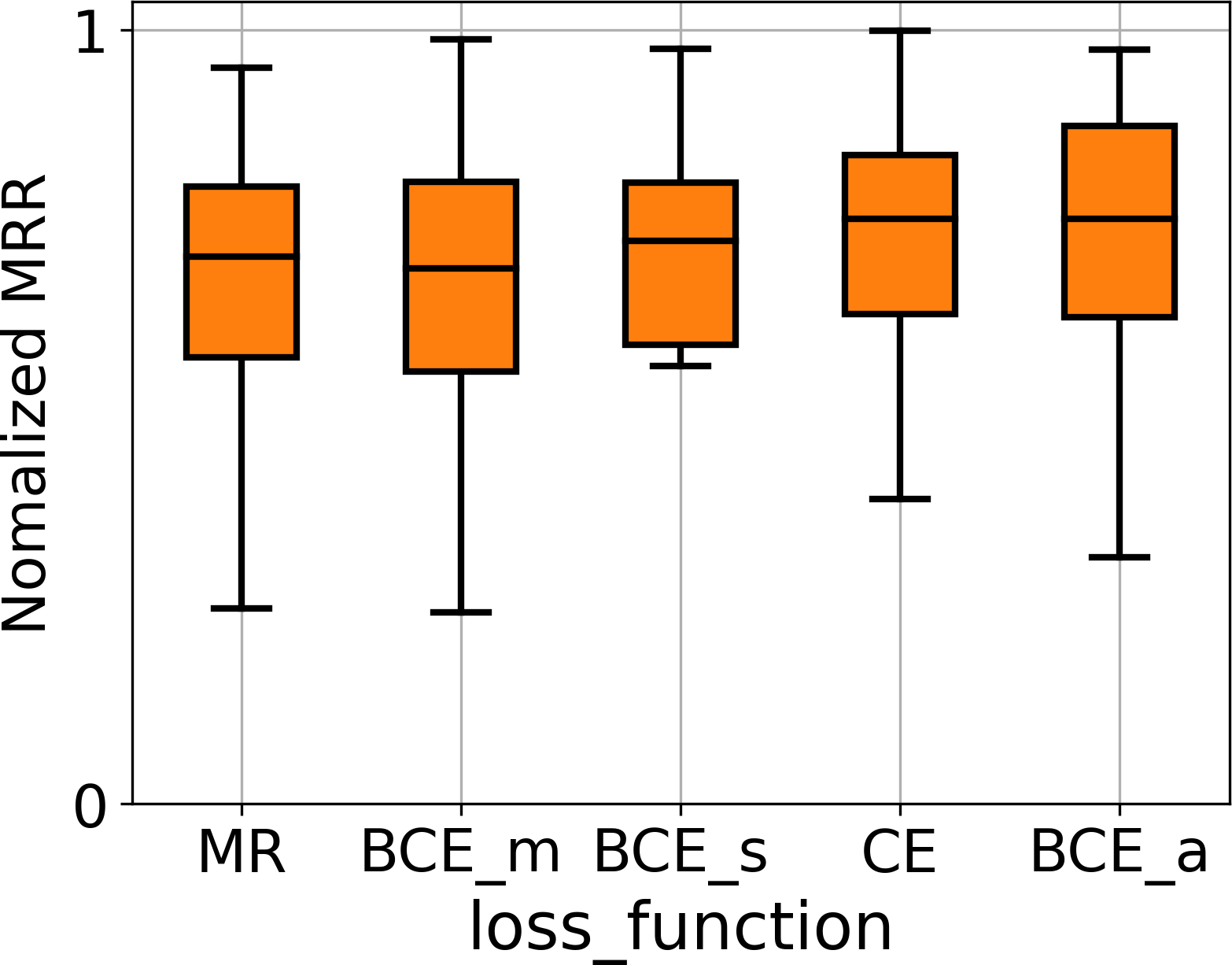}} \hfill
	\subfigure[gamma (MR)]{\includegraphics[width=0.24\textwidth]{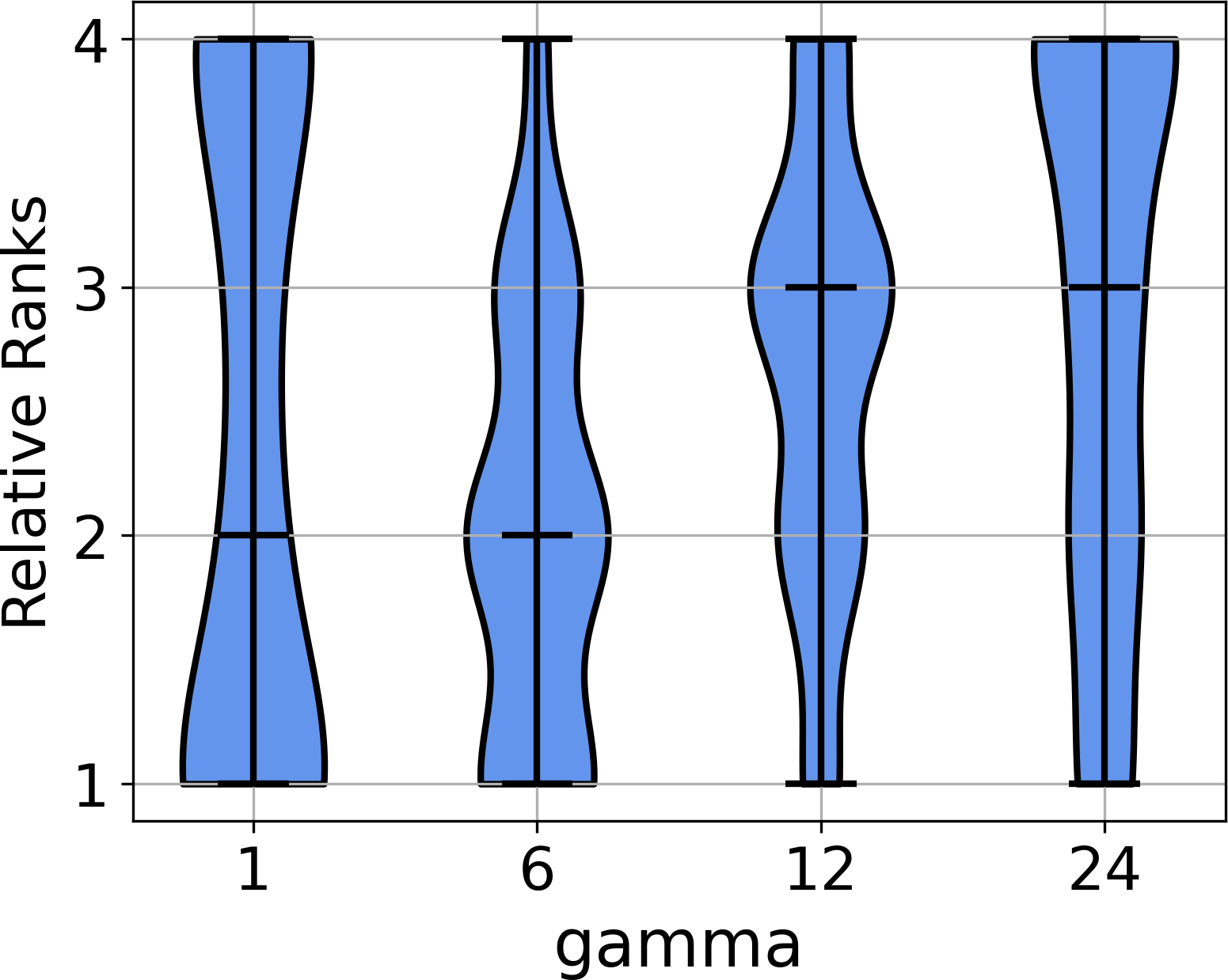}
		\includegraphics[width=0.24\textwidth]{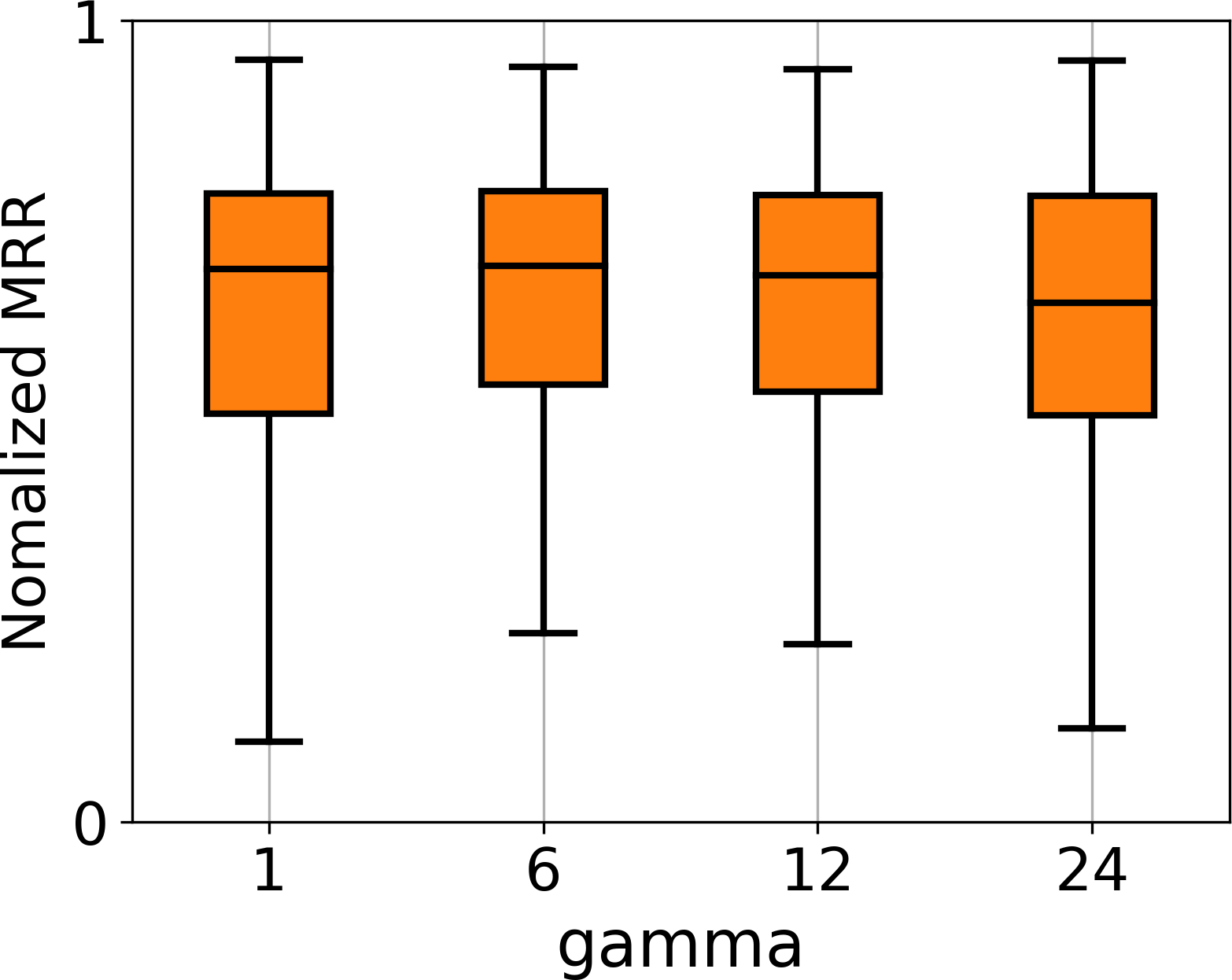}}
	\vspace{-5px}
	
	\subfigure[adv weight (BCE\_adv)]{\includegraphics[width=0.24\textwidth]{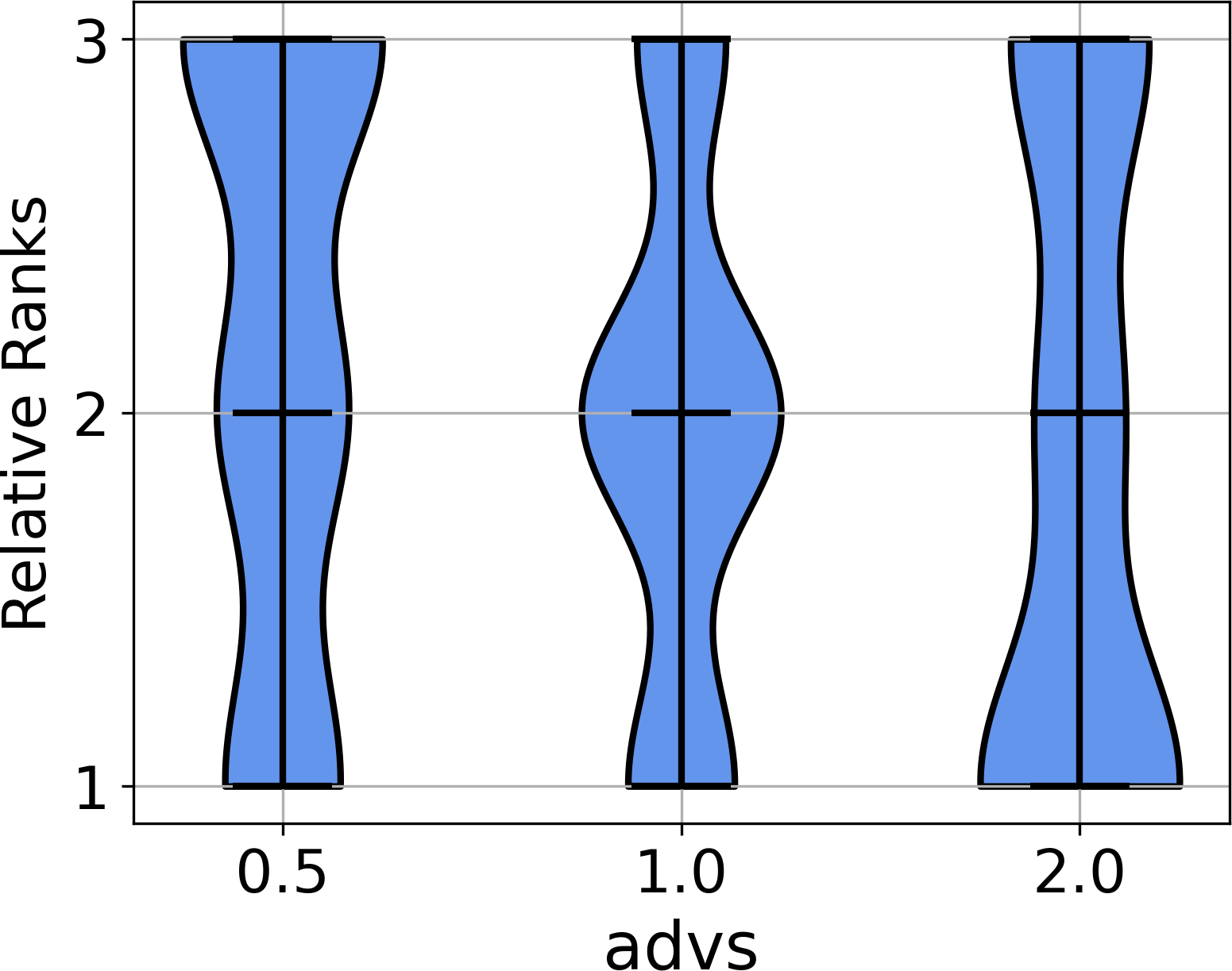}
		\includegraphics[width=0.24\textwidth]{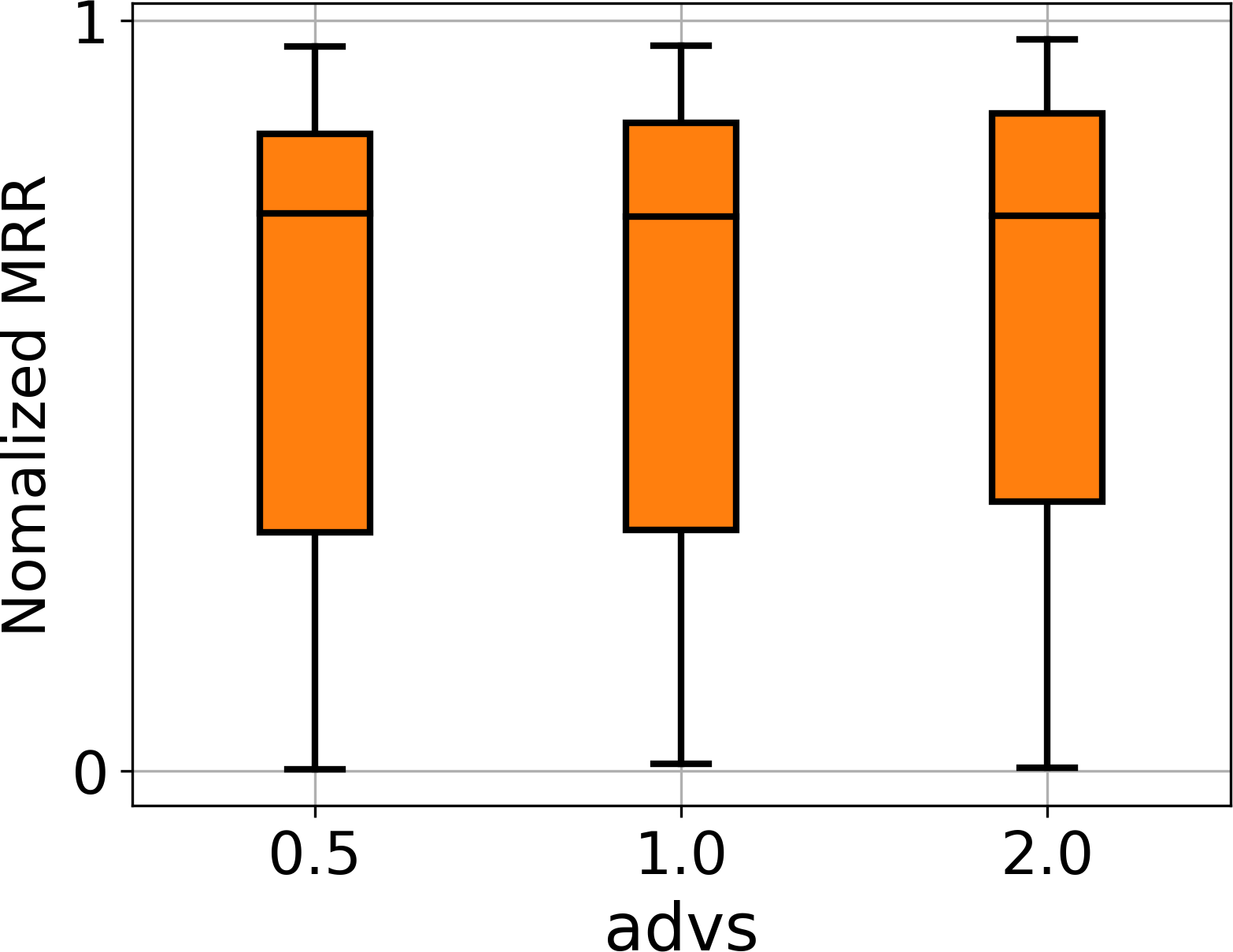}}\hfill
	\subfigure[\# negative sample]{\includegraphics[width=0.243\textwidth]{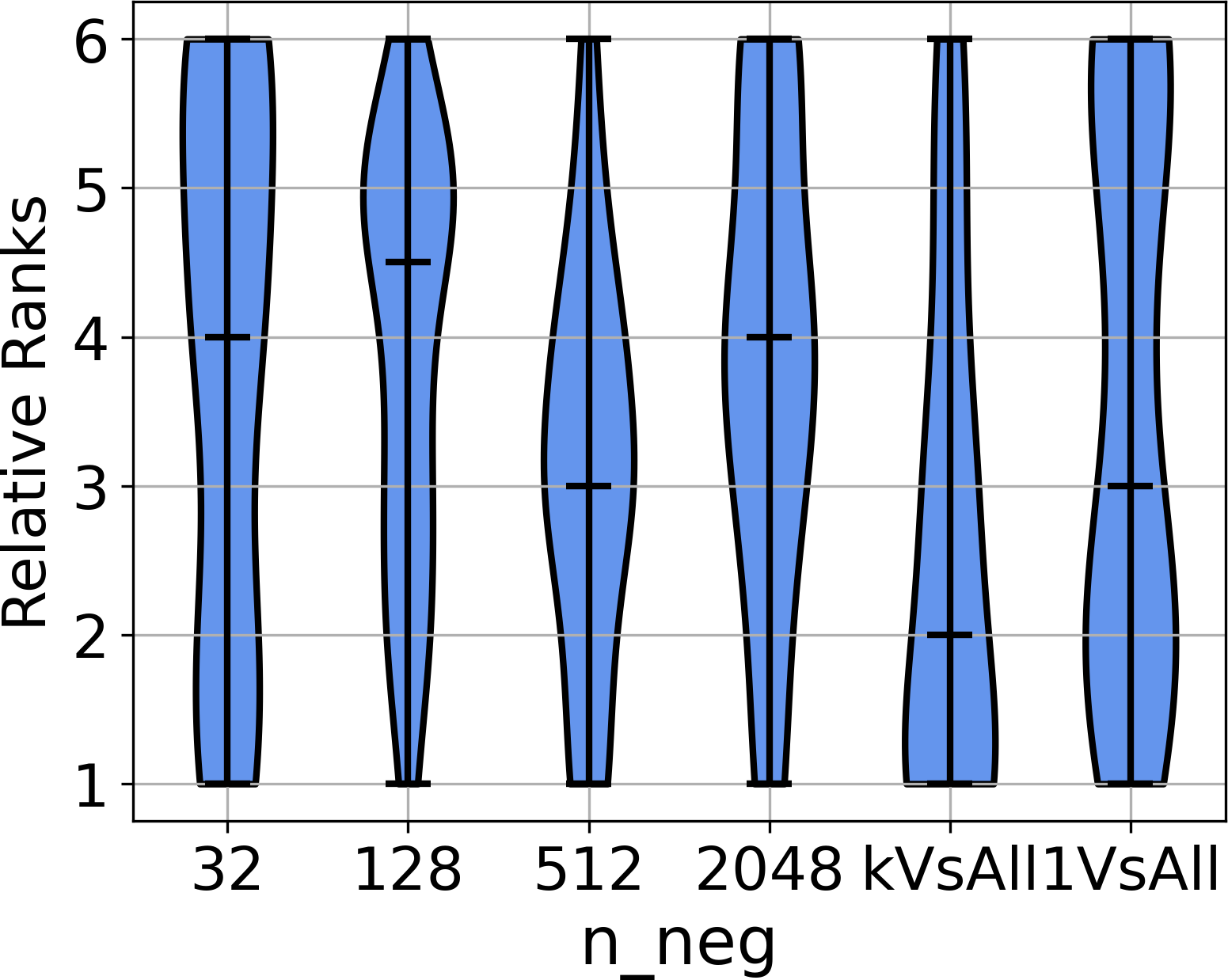}
		\includegraphics[width=0.243\textwidth]{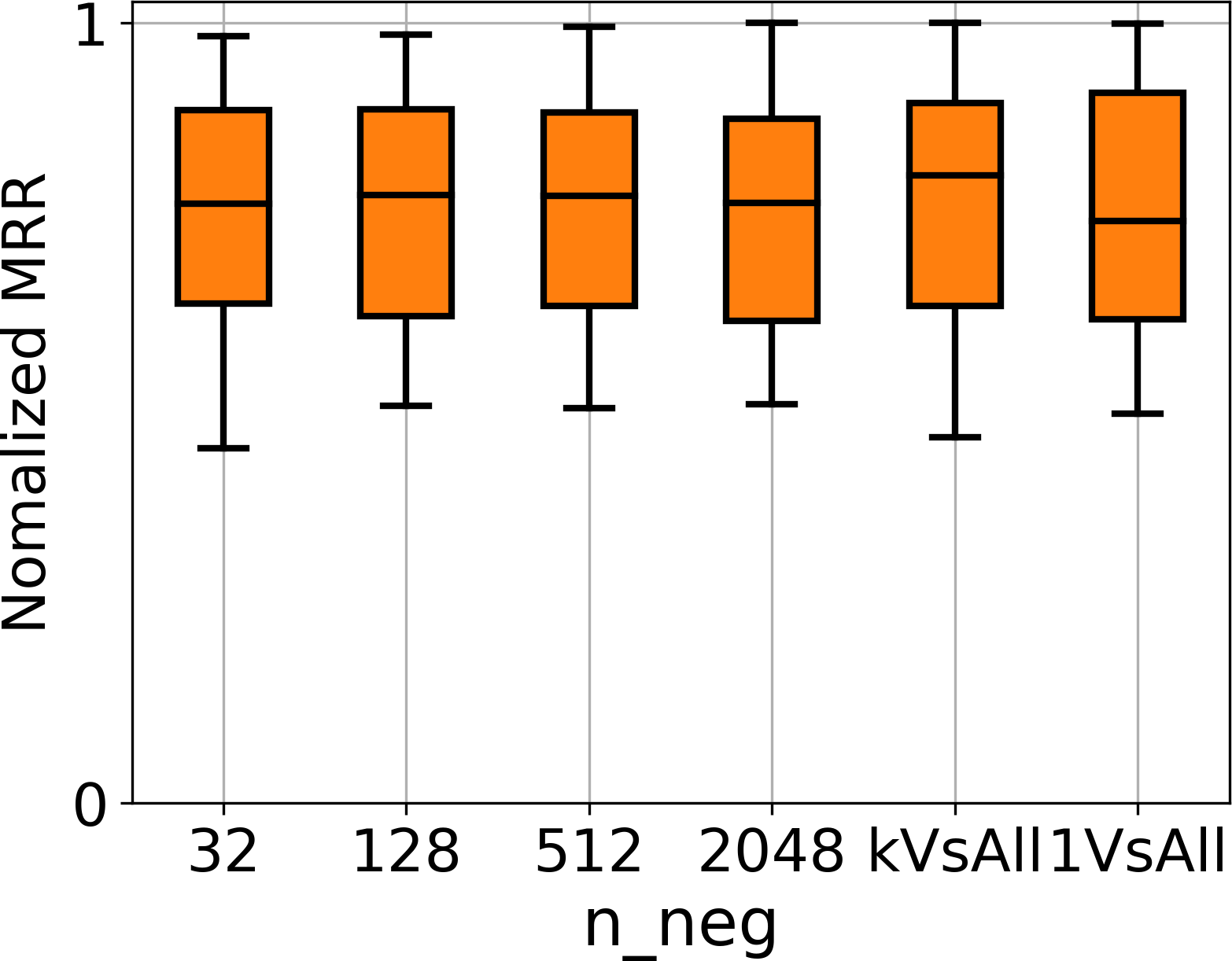}}
	
	\vspace{-5px}	
	\subfigure[regularizer]{\includegraphics[width=0.243\textwidth]{figures/HP_understanding_regularizer_rank}
		\includegraphics[width=0.243\textwidth]{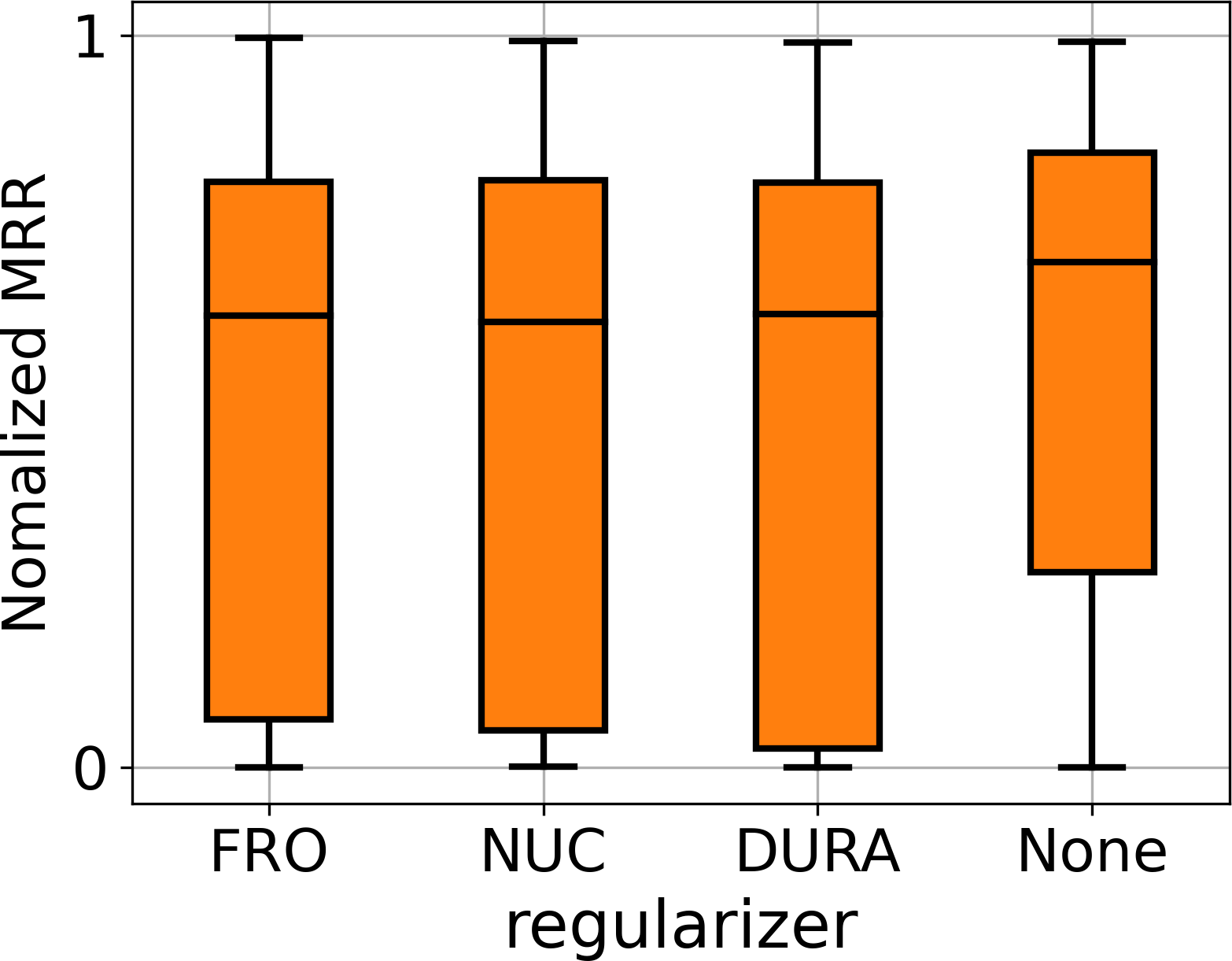}} \hfill
	\subfigure[initializer]{\includegraphics[width=0.24\textwidth]{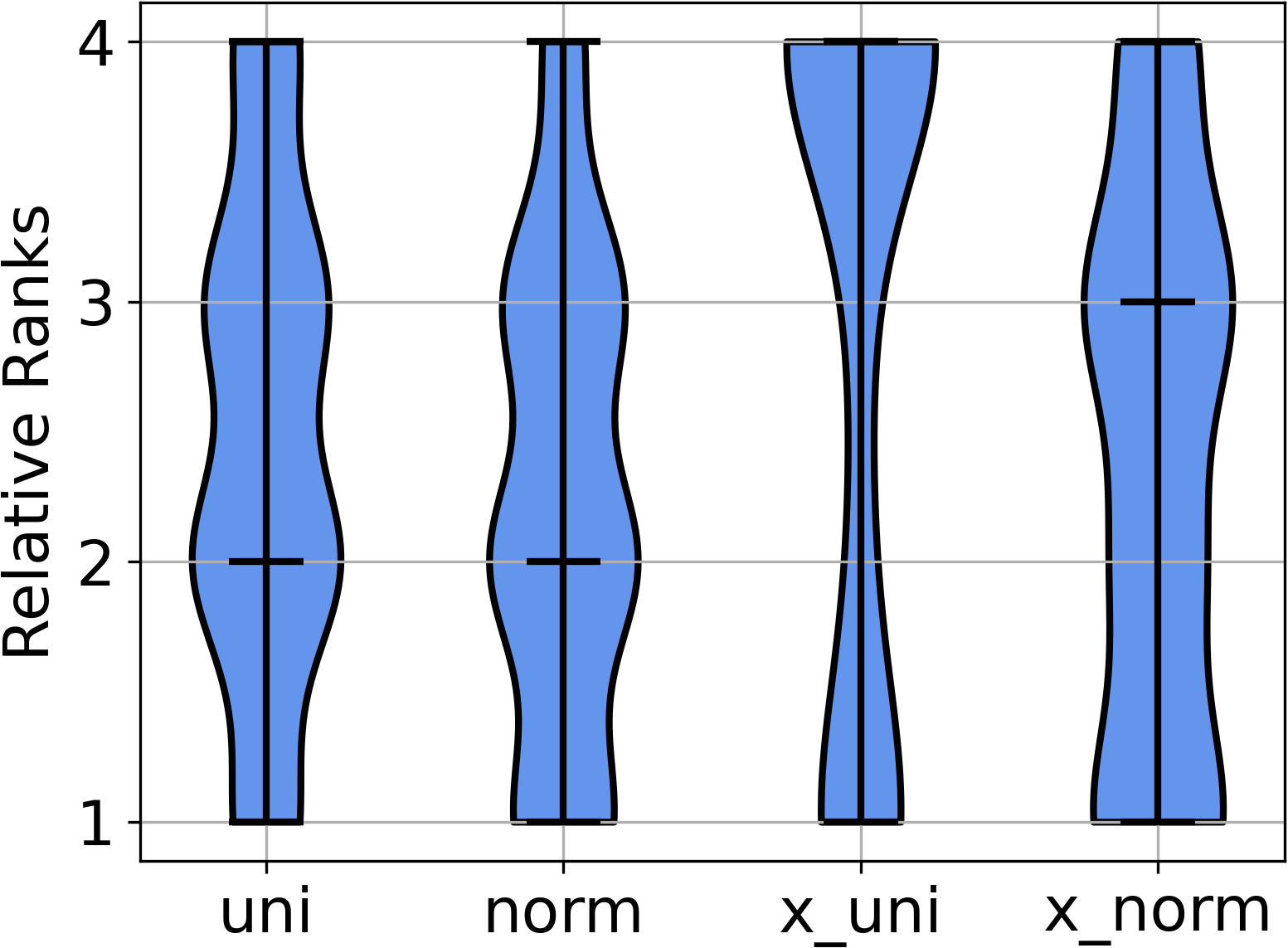}
		\includegraphics[width=0.24\textwidth]{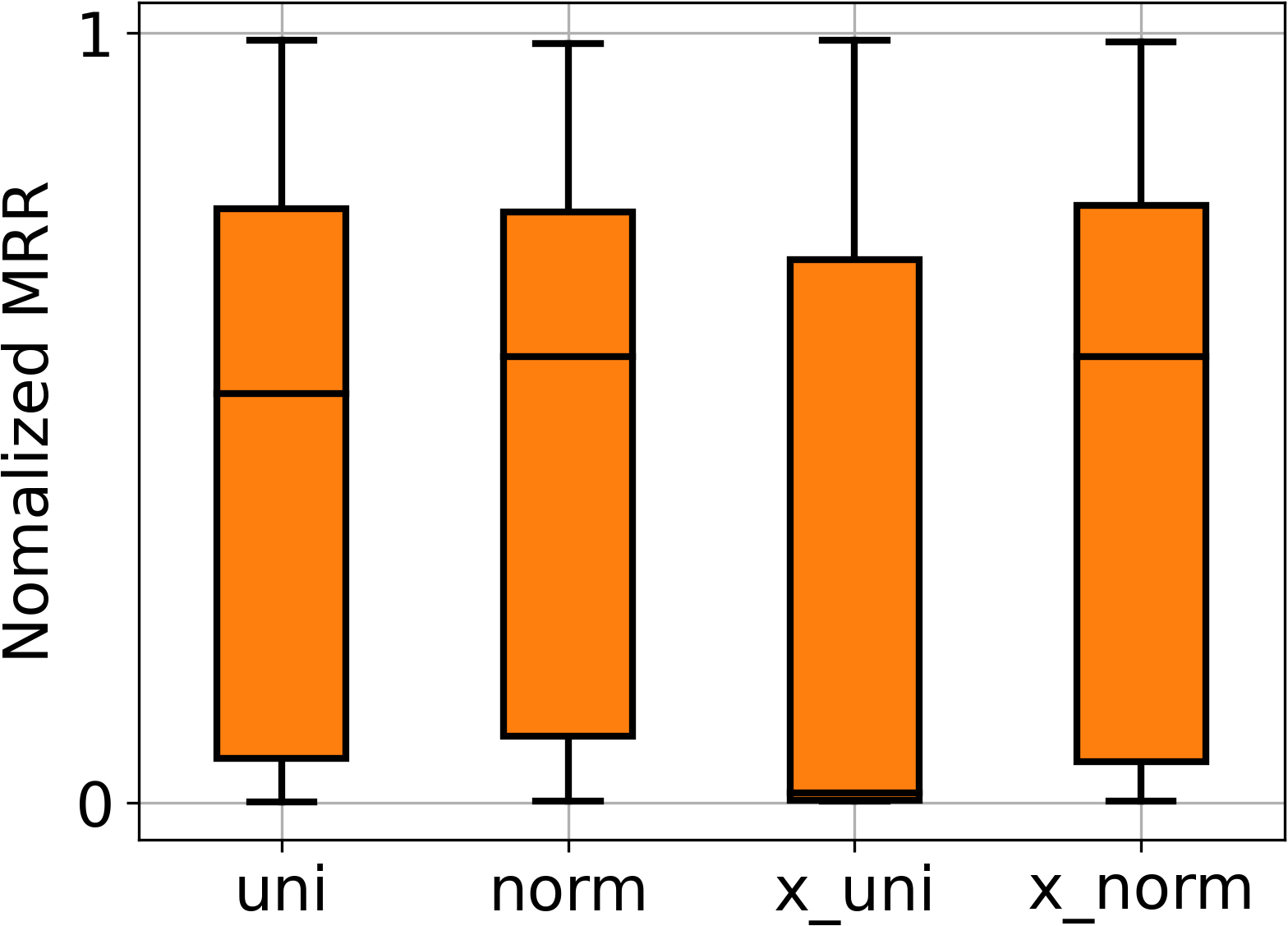}}	

	\vspace{-10px}
	
	\caption{HPs that do not have obvious patterns. All of the values should be searched.}
	\label{fig:hp:nopattern}	
\end{figure}

The full results for the four types of HPs in Section~\ref{ssec:indivudial}
are provided in Figures~\ref{fig:hp:fixed}-\ref{fig:hp:nopattern}. 
The larger area in the bottom in the voilin plots 
and the top area in the box plots indicate better performance.
The HPs can be classified into four types:
\begin{itemize}[leftmargin=20pt]
	\item[(a).] \textit{fixed choices}: Adam is the fixed optimizer, and inverse relation is not preferred. See Figure~\ref{fig:hp:fixed}. 
	\item[(b).] \textit{limited range}: Learning rate, regularization weight and dropout rate should be limited in the ranges
		$[10^{-4}, 10^{-1}]$, $[10^{-12}, 10^{-2}]$ and $[0,0.3]$, respectively.
		See Figure~\ref{fig:hp:limited}
	\item[(c).] \textit{monotonously related}: Batch size and dimension size have monotonic performance. The larger value tends to lead better results.
		See Figure~\ref{fig:hp:monotonic}.
	\item[(d).] \textit{no obvious patterns}: The choice of loss function, value of gamma, adversarial weight, number of negative samples, regularizer, initializer do not have obvious patterns.
		See Figure~\ref{fig:hp:nopattern}.
\end{itemize}

In addition,
we provide the details of Spearman's ranking correlation coefficient (\texttt{SRCC}).
Given a set of anchor configurations $\mathcal X_i$
to analyze the $i$-th HP,
we denote $r(\mathbf x, \theta)$ as the rank of different $\mathbf x\in\mathcal X_i$ with fixed $x_i=\theta$.
Then,
the \texttt{SRCC} between two 
HP values $\theta_1, \theta_2\in X_i$ is 
\begin{equation}
\texttt{SRCC}
(\theta_1, \theta_2)
=
1 - 
\frac{\sum_{\mathbf x\in\mathcal X_i}|r(\mathbf x,\theta_1)-r(\mathbf x,\theta_2)|^2}
{|\mathcal X_i|\cdot (|\mathcal X_i|^2-1)},
\end{equation}
where $|\mathcal X_i|$
means the number of anchor configurations in $\mathcal X_i$.
We evaluate
the consistency of the $i$-th HP 
by averaging the \texttt{SRCC} over 
the different pairs of $(\theta_1, \theta_2) \in X_i\times X_i$,
the different models and datasets.

\subsection{Approximation ability of surrogate models}
\label{app:fitting}

In Section~\ref{ssec:surrogate},
we have shown that the curvature of a learned random forest (RF)
model is more similar with the real curvature of the ground truth.
Here,
we further demonstrate this point through a synthetic experiment.

Specifically,
100 random configurations with evaluated performance are sampled.
We use 10/20/30 
random samples from them to train the surrogates
since only a small number of  HP configurations
are available for the surrogate during searching.
The remaining configurations are used for testing.
Then,
we evaluate the fitting ability of each model
by the mean square error (MSE)
of the estimated prediction to the target prediction.
For GP \citep{rasmussen2003gaussian}, we show the prediction with the Matern kernel used in AutoNE \citep{tu2019autone}.
For RF \citep{breiman2001random}, we build 200 tree estimators to fit the training samples.
The MLP here \citep{gardner1998artificial} is designed as a three-layer feed-forward network 
with 100 hidden units and ReLU activation function in each layer.
The
average value and std of
MSE  over five different groups of configurations
are shown in Table~\ref{tab:surrogate}.
As can been seen,
random forest show much lower prediction error than GP and MLP
with different number of training samples.
This further demonstrates that RF can better fit 
such a complex HP search space.

\begin{table}[ht]
	\centering
	\caption{Comparison of different surrogate models in MSE.}
	\label{tab:surrogate}
	\vspace{-8px}
	\small
	\begin{tabular}{c|ccc}
		\toprule
		\# train configurations & 10 & 20 &30 \\
		\midrule
		GP  & 0.0693$\pm$0.02 &0.029$\pm$0.01  & 0.019$\pm$0.01  \\
		MLP  &  2.121$\pm$0.4 & 2.052$\pm$0.3  & 0.584$\pm$0.1 \\
		RF   & \underline{\textbf{ 0.003$\pm$0.002}}  &  \underline{\textbf{0.002$\pm$0.001}} &  \underline{\textbf{0.001$\pm$0.001}} \\ 
		\bottomrule
	\end{tabular}
\end{table}

\subsection{Results of cost evaluation}
\label{app:under:cost}

\begin{figure}[ht]
	\centering
	\vspace{-3px}
	\subfigure{\includegraphics[width=0.192\textwidth]{figures/HP_cost_batchsize_time}}  
	\subfigure{\includegraphics[width=0.192\textwidth]{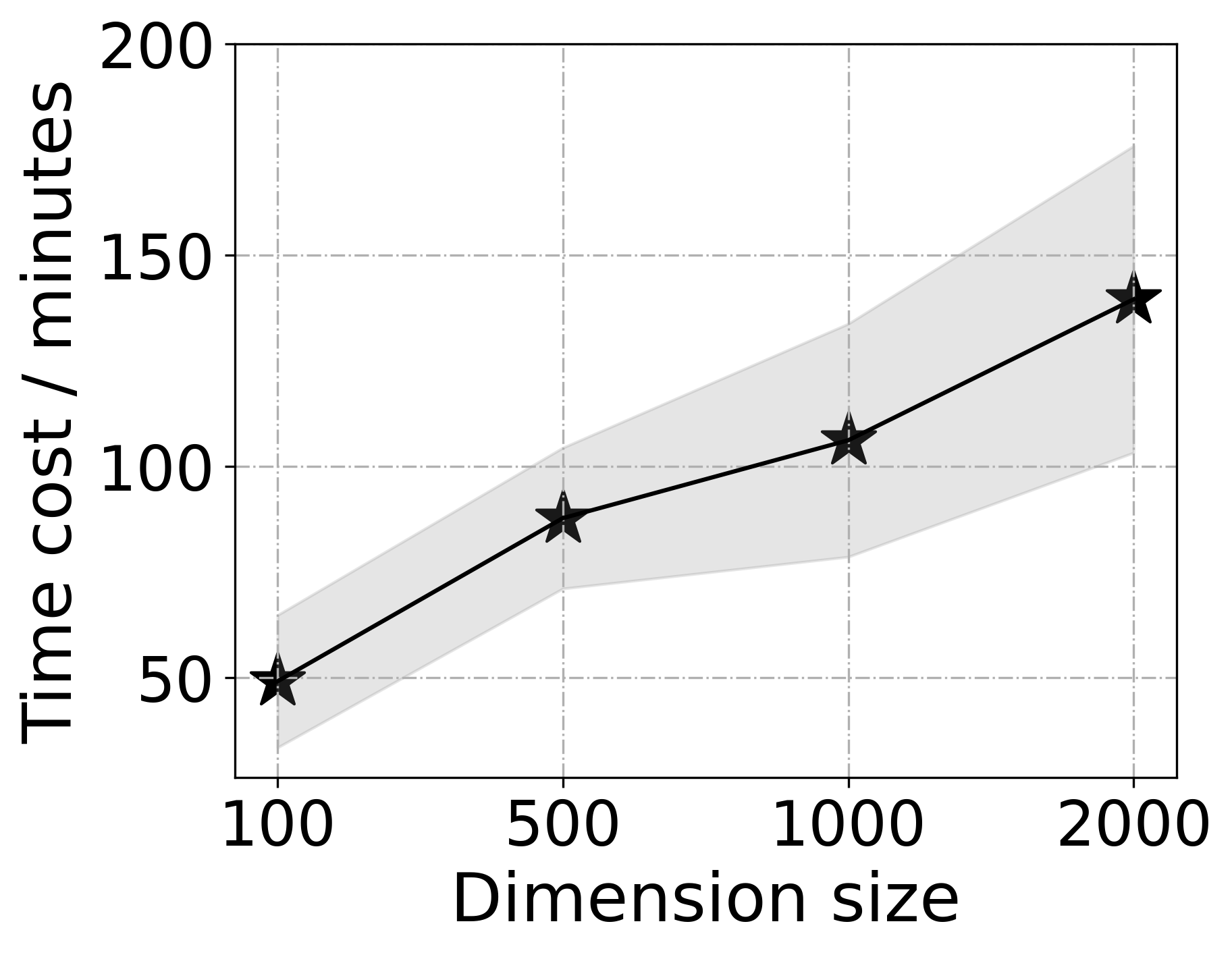}} 
	\subfigure{\includegraphics[width=0.195\textwidth]{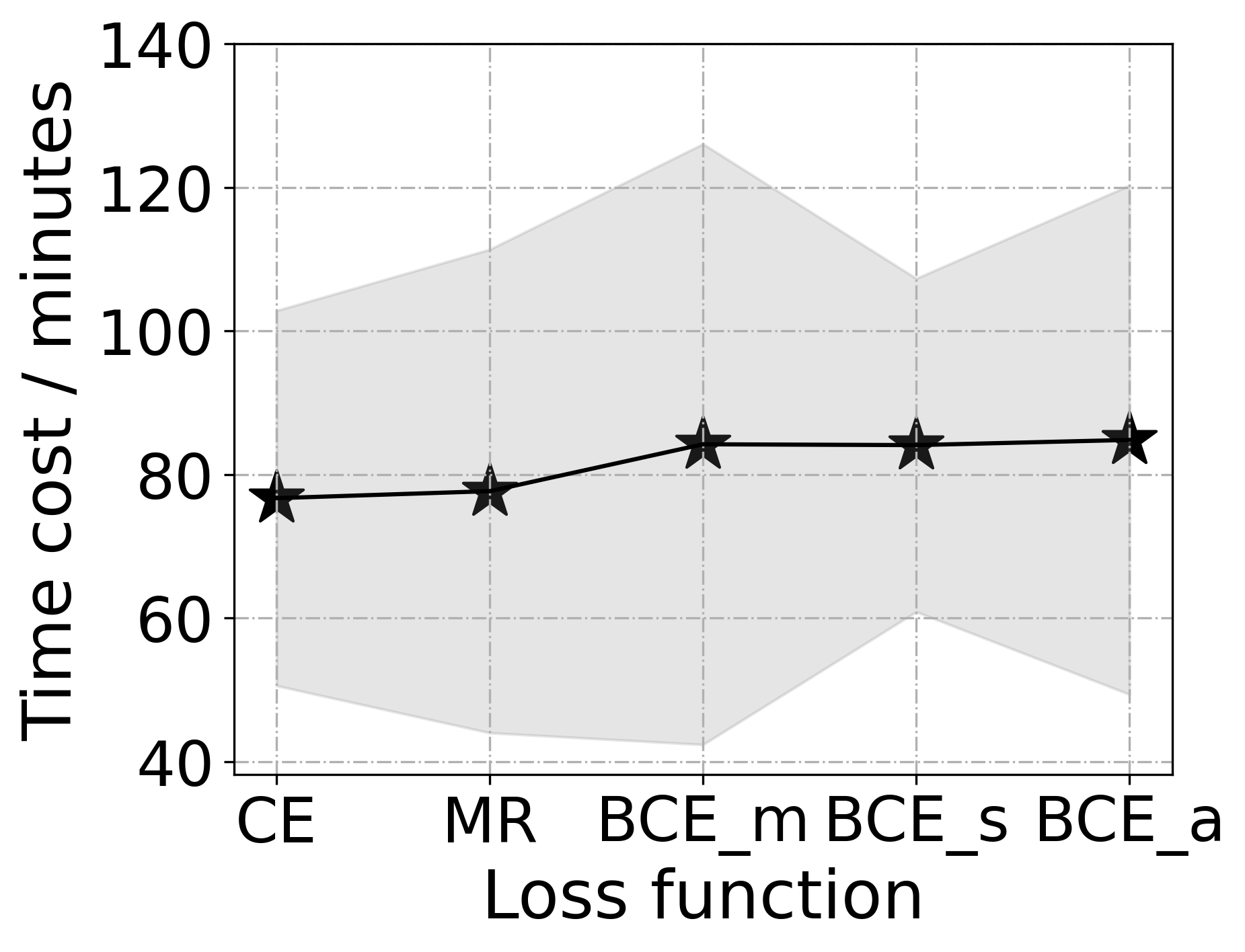}}
	\subfigure{\includegraphics[width=0.191\textwidth]{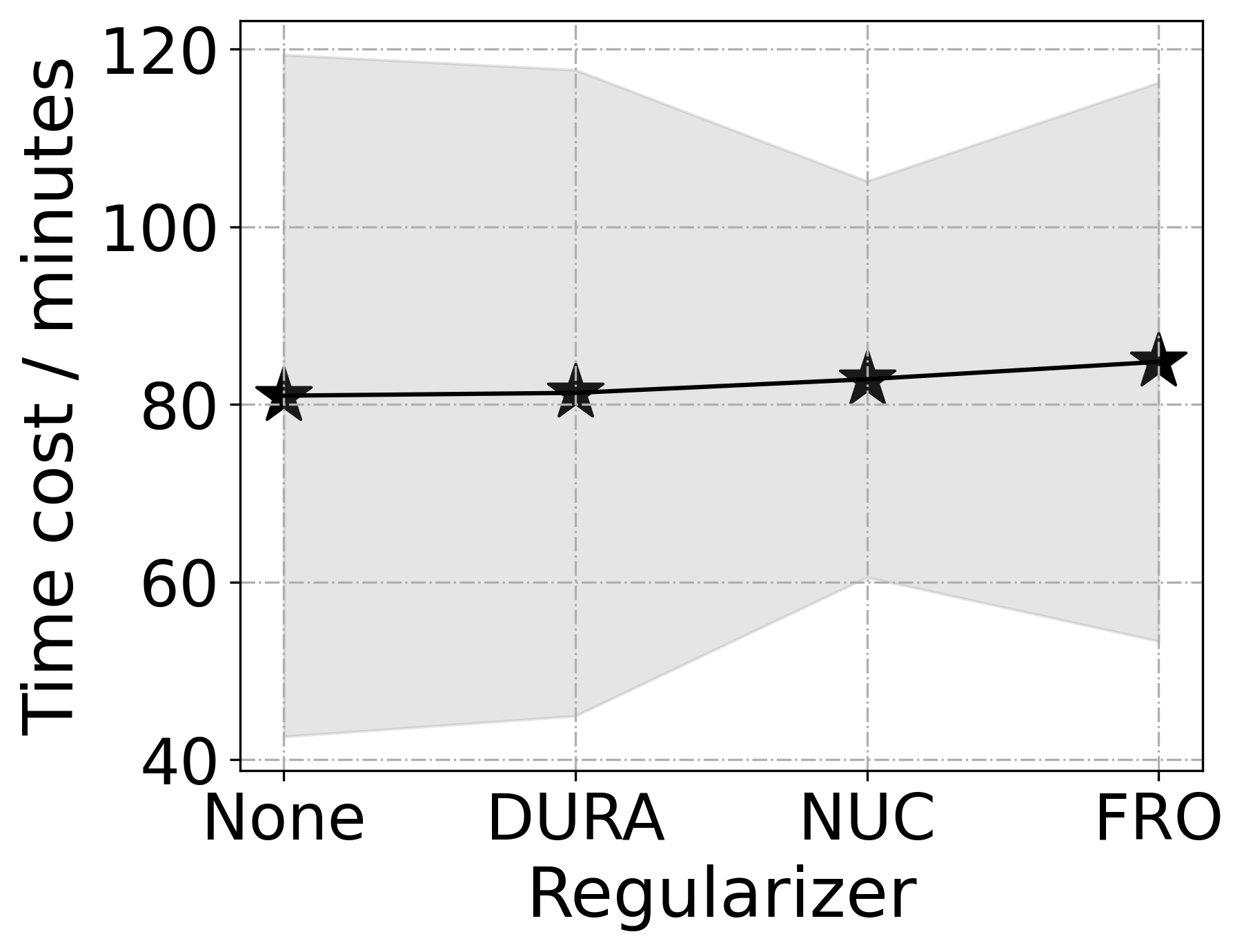}}
	\subfigure{\includegraphics[width=0.1895\textwidth]{figures/HP_cost_negative_samples_time}} 
	\vspace{-8px}
	
	\caption{Computing time cost. The dots are the average and the shades are the standard deviation.}	
	\label{fig:app:hyper-cost}
	\vspace{-5px}
\end{figure}

We show the average cost and standard derivation of 
five HPs,
i.e. batch size, dimension size,
number of negative samples,
 loss functions,
 and regularizer,
in Figure~\ref{fig:app:hyper-cost}.
As can be seen,
the cost of batch size and dimension size increase much 
when the size increases.
But for 
the number of negative samples, 
choices of loss functions
and regularizers,
the influence on cost is not strong
as indicated by the average cost.

\section{Detail for the search algorithm}
\label{app:algorithm}

\subsection{Search space}
We show  
the shrunken and decoupled search space 
compared with the full space in Table~\ref{tab:spacereduce}.
To evaluate the ratio of space change after shrinkage and decoupling,
we measure the learning rate
and regularization weight in log scale.
The size of the whole space $\mathcal X$ compared with the decoupled ${\mathcal X}_{S|D}$
is
\[3\times \frac{14}{6}\times\frac{5}{3}\times\frac{5}{3}\times 2\times 4 \times 5 = 777.8. \]
Hence,
the reduced and decoupled space is hundreds times smaller than the full space.

\begin{table}[ht]
	\centering
	\caption{The revised HP values in the reduced and decoupled search space compared with the full space.}
	\label{tab:spacereduce}
	\setlength\tabcolsep{2pt}
	\small
	\vspace{-10px}
	\begin{tabular}{c|c|c}
		\toprule
		name  & ranges in the whole space & revised ranges \\
		\midrule
		optimizer &  \{Adam, Adagrad, SGD\}           &        Adam       \\
		learning rate & [$10^{-5}$, $10^0$]   &  [$10^{-4}$, $10^{-1}$]  \\  
		reg. weight &   [$10^{-12}$, $10^{2}$]  & [$10^{-8}$, $10^{-2}$]  \\
		dropout rate & [0, 0.5]  & [0, 0.3] \\ 
		inverse relation   & \{True, False\}  & \{False\} \\
		\midrule
		batch size & \{128, 256, 512, 1024\} & 128 \\
		dimension size   &   \{100, 200, 500, 1000, 2000\}  &   100 \\
		\bottomrule
	\end{tabular}
\end{table}

\subsection{Search algorithm}

\begin{figure*}[ht]
	\centering
	\vspace{-5px}
	\includegraphics[width=0.73\columnwidth]{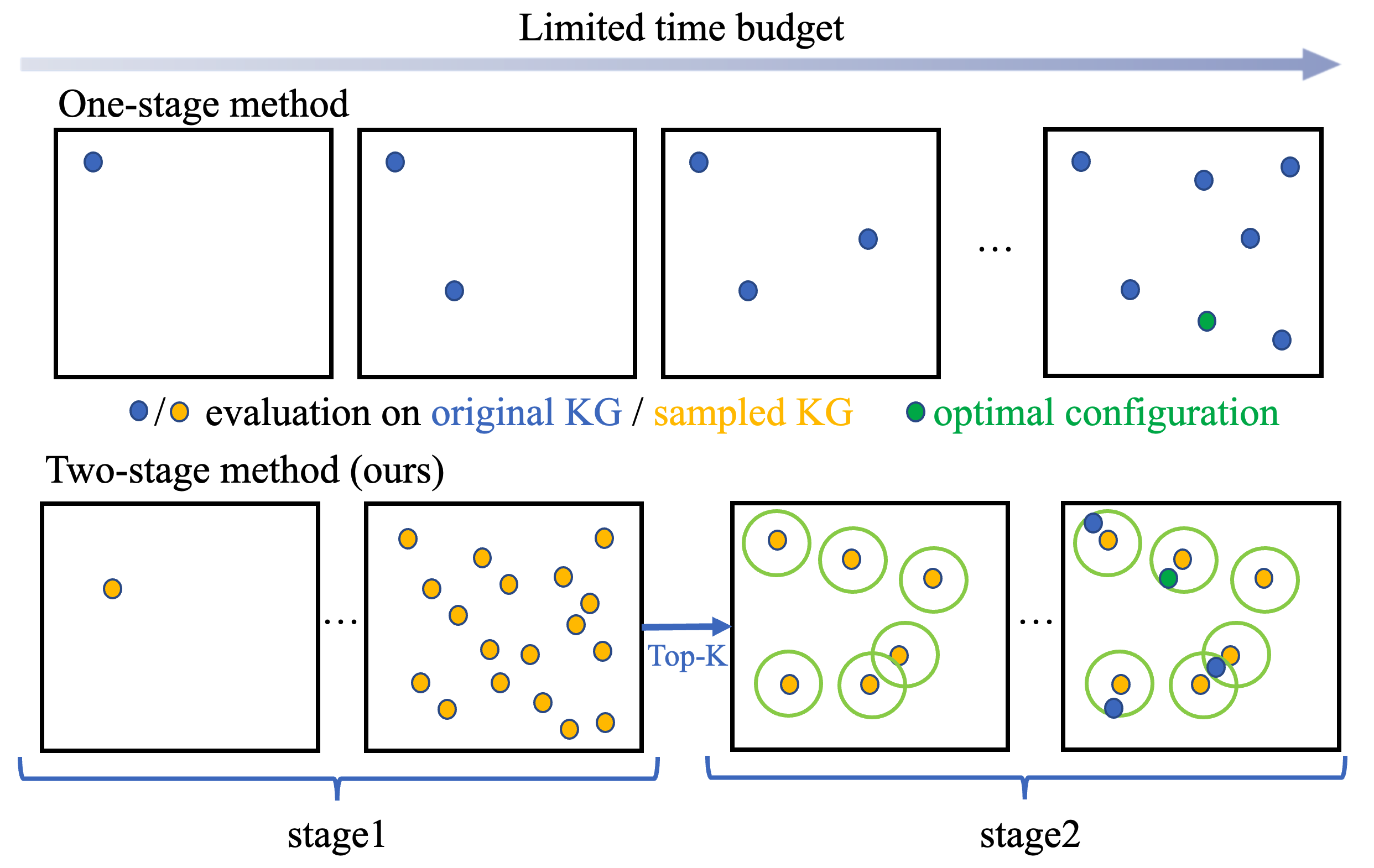}
	\vspace{-10px}
	\caption{Diagram of one-stage search method and the proposed two-stage method.
	}	
	\label{fig:one_two_stage_comparison}
	\vspace{-2px}
\end{figure*}

We visualize the
searching process of
the traditional one-stage method
and the proposed two-stage method in 
Figure~\ref{fig:one_two_stage_comparison}.
Since the evaluation cost on the full graph
is rather high, 
the one-stage method 
can only take a few optimization trials.
Thus the search space remains unexplored 
for a large proportion,
and the performance of 
the optimal configuration 
is hard to be guaranteed.
As for the proposed two-stage method KGTuner,
it efficiently explores the search space
on the sampled subgraph
at the first stage,
and then fine-tunes the top-K 
configurations
on the full graph.

In Algorithm~\ref{alg:search},
we increase the batch size and dimension size in stage two.
We set the searched range for batch size in stage two 
as $[512, 1024]$
and dimension size as $[1000,2000]$.
There are some exceptions due to the memory issues,
i.e.,
dimension size for RESCAL is in $[500, 1000]$;
dimension size for TuckER is in $[200, 500]$.
For ogbl-wikikg2,
since the used GPU only has 24GB memory,
we cannot run models with 500 dimensions
which requires much more memory 
in the OGB board.
Instead,
we set the dimension as 100 to be consistent with the smaller models
in OGB board with 100 dimensions,
and increase the batch size in $[512, 1024]$
in the second stage.
In addition,
we show the details for the search procedure
by RF+BORE
in Algorithm \ref{alg:full}.

\begin{algorithm}[ht]
	\caption{Full procedure of HP search with RF+BORE (in stage one)}
	\label{alg:full}
	\small
	\begin{algorithmic}[1]
		\REQUIRE KG embedding $F$, dataset $G$, search space ${\mathcal X}_{S|D}$, budget $\nicefrac{B}{2}$, RF model $y=c(\mathbf x)$, threshold $\tau=0.8$.
		\STATE initialize the RF model and $\mathcal H = \emptyset$;
		\STATE split triplets in $G$ with ratio $9:1$ into $G_{\text{tra}}$ and $G_{\text{val}}$;
		\REPEAT
			\STATE randomly sample a set of configurations ${{\mathcal X}_{\overline{S|D}}}\subset{\mathcal X}_{S|D}$;
			\STATE select $\hat{\mathbf x} = \arg\max_{\mathbf x\in {{\mathcal X}_{\overline{S|D}}}} y(\mathbf x)$;
			\STATE train embedding model into converge \\
				$\bm P^* = \arg\min_{\bm P}\mathcal L\big(F(\bm P, \hat{\mathbf x}), G_{\text{tra}}\big)$;
			\STATE evaluate the performance $\hat{y}_{\hat{\mathbf x}} = \mathcal M\big(F(\bm P^*, \hat{\mathbf x}), G_{\text{val}}\big)$;
			\STATE record $\mathcal H\leftarrow \mathcal H\cup \{(\hat{\mathbf x}, \hat{y}_{\hat{\mathbf x}})\}$;\\
			\% \textbf{BORE}:
			\STATE set label $0$ for configuration in  $\mathcal H$ with $\hat{y}_{\hat{\mathbf x}}<\tau$,
			and label $1$ for $\hat{y}_{\hat{\mathbf x}}\geq\tau$;
			\STATE update RF model $y=c(\mathbf x)$ to classify the two labels;
		\UNTIL{$\nicefrac{B}{2}$ exhausted.}
	\end{algorithmic}
\end{algorithm}

\section{Additional experimental results}

\subsection{Implementation details}
\label{app:implementation}



\noindent
\textbf{Evaluation metrics.}
We follow \cite{bordes2013translating,wang2017knowledge,ruffinelli2019you}
to use the filtered ranking-based metrics for evaluation.
For each triplet $(h,r,t)$ in the validation or testing set,
we take the head prediction $(?,r,t)$ and tail prediction $(h,r,?)$
as the link prediction task.
The filtered rankings on the head and tail are computed as 
\[
\text{rank}_h = 
\Big|\big\{e\in\mathcal E: 
\big(f(e,r,t)\geq f(h,r,t) \big)
\wedge 
\big((e,r,t)\notin D_{\text{tra}}\cup D_{\text{val}}\cup D_{\text{tst}}) \big) 
\big\}\Big| + 1,
\]
\[
\text{rank}_t = 
\Big|\big\{e\in\mathcal E: 
\big(f(h,r,t)\geq f(h,r,e) \big)
\wedge 
\big((h,r,e)\notin D_{\text{tra}}\cup D_{\text{val}}\cup D_{\text{tst}}) \big) 
\big\}\Big| + 1,
\]
respectively, where $|\cdot|$ is the number of elements in the set.
The the two metrics used are:
\begin{itemize}
	\item Mean reciprocal ranking (MRR): the average of reciprocal of all the obtained rankings.
	\item Hit@$k$: the ratio of ranks no larger than $k$.
\end{itemize}
For both the metrics,
the large value indicates the better performance.


\vspace{8px}

\noindent
\textbf{Dataset statistics.}
We summarize the statistics of different
benchmark datasets
in Table~\ref{tab:dataset}.
As shown,
ogbl-biokg and ogbl-wikikg2
have much larger size compared with
WN18RR and FB15k-237.

\begin{table}[ht]
	\centering
	\caption{Statistics of the KG completion datasets.}
	\vspace{-10px}
	\small
	\label{tab:dataset}
	\setlength\tabcolsep{3.3pt}
	\begin{tabular}{c|cc|ccc}
		\toprule
		dataset                 & \#entity & \#relation &  \#train & \#validate &
		\#test  \\ \midrule
		WN18RR ~\cite{dettmers2017convolutional} &  41k  &     11     &  87k   &  3k  & 3k   \\
		FB15k-237~\cite{toutanova2015observed}   &  15k  &    237     &  272k  & 18k  & 20k  \\
		ogbl-biokg~\cite{hu2020open} & 94k & 51 & 4,763k & 163k  & 163k 	\\
		ogbl-wikikg2~\cite{hu2020open} & 2,500k & 535 & 16,109k & 429k  & 598k 	\\
		\bottomrule
	\end{tabular}
\end{table}

\noindent
\textbf{Baseline implementation.}
All the baselines compared in this paper
are based on their own original open-source implementations.
Here we list the source links:
\begin{itemize}
	\item Hyperopt \citep{bergstra2013hyperopt}, \url{https://github.com/hyperopt/hyperopt};
	\item Ax, \url{https://github.com/facebook/Ax};
	\item SMAC \citep{hutter2011sequential}, \url{https://github.com/automl/SMAC3};
	\item BORE \citep{tiao2021bore}, \url{https://github.com/ltiao/bore};
	\item AutoNE \citep{tu2019autone}, \url{https://github.com/tadpole/AutoNE}.
\end{itemize}

\clearpage
\noindent
\textbf{Searched hyperparameters.} 
We list the searched hyperparameters for each embedding model
on the different datasets 
in Tables~\ref{tab:searched_HP_WN18RR}-\ref{tab:searched_HP_wikikg2}
for reproduction.

\begin{table}[ht]
	\centering
	\caption{Searched optimal hyperparameters for the WN18RR dataset.}
	\label{tab:searched_HP_WN18RR}
	\small
	\vspace{-5px}
	\renewcommand{\arraystretch}{1.2}
	\setlength\tabcolsep{3pt}
	\begin{tabular}{c|c|c|c|c|c|c|c }
		\toprule
		HP/Model  & ComplEx  &  DistMult  & RESCAL  &  ConvE & TransE &  RotatE & TuckER   \\ \midrule
		\# negative samples 	&  32 & 128 &  128 &  512 &  128  &  2048  &  128   \\   \midrule
		loss function 					& BCE\_mean & BCE\_adv & BCE\_mean  &  BCE\_adv  & BCE\_adv   &  BCE\_adv  &  BCE\_adv \\   
		gamma 							  & 2.29 & 12.88 & 2.41 & 12.16  &  3.50  &  3.78  &  12.97 \\   
		adv. weight 					& 0.00 & 1.41 & 0.00 &  0.78 &  1.14  &  1.66  &  1.94 \\ 
		\midrule
		regularizer 		  			 & NUC & NUC & DURA & DURA  & FRO   &  FRO  & DURA  \\ 
		reg. weight 					 & $1.21 \times 10^{-3}$ & $9.58 \times 10^{-3}$ & $1.76 \times 10^{-3}$ & $ 9.79 \times 10^{-3}$   & $ 4.19 \times 10^{-4}$    &    $5.13 \times 10^{-8}$ &  $2.22  \times 10^{-3}$  \\  
		dropout rate 					& 0.28 & 0.29 & 0.00 &  0.02 &  0.00  &  0.00  & 0.00  \\
		\midrule
		optimizer 						   &  Adam  &  Adam&   Adam &   Adam  &    Adam  &  Adam & Adam  \\  
		learning rate  					& $6.08 \times 10^{-4}$ & $4.58 \times 10^{-3}$  & $ 1.73\times 10^{-3}$ &  $ 6.88 \times 10^{-4}$  &  $ 1.02 \times 10^{-4}$   &  $ 1.24 \times 10^{-3}$   &  $2.60 \times 10^{-3}$ \\  
		initializer  						 & x\_uni & norm & uni  & x\_uni  &  norm  &  norm  & x\_uni  \\   
		\midrule
		batch size   				      & 1024 & 1024 & 512 & 512  &  512  &  512  &  512 \\ 
		dimension size   			  & 2000 & 2000  & 1000 & 1000  & 1000   & 1000   & 200 \\ 
		inverse relation    		  &  False & False &   False&  False&  False&  False&  False \\   
		\bottomrule
	\end{tabular}
\end{table}

\begin{table}[ht]
	\centering
	\caption{Searched optimal hyperparameters for the FB15k-237 dataset.}
	\label{tab:searched_HP_FB15k237}
	\small
	\vspace{-5px}
	\renewcommand{\arraystretch}{1.2}
	\setlength\tabcolsep{3pt}
	\begin{tabular}{c|c|c|c|c|c|c|c }
		\toprule
		HP/Model  & ComplEx  &  DistMult  & RESCAL  &  ConvE & TransE &  RotatE & TuckER   \\ \midrule
		\# negative samples 	& 512 & \texttt{kVsAll} & 2048 & 512  &  512  &  128  & 2048 \\    \midrule
		loss function 					& BCE\_adv & CE & CE & BCE\_sum  &  BCE\_adv  &  BCE\_adv  & BCE\_adv \\  
		gamma 							  & 13.05 & 2.90 & 4.17 & 14.52  &  6.76  &  14.46  & 13.51 \\   
		adv. weight 					& 1.93 & 0.00 & 0.00 &  0.00 &  1.99  &  1.12  & 1.95 \\   
		\midrule
		regularizer 		  			 & DURA & NUC & DURA & DURA &  FRO  &  NUC   & DURA \\
		reg. weight 					 & $9.75 \times 10^{-3}$ & $2.13 \times 10^{-3}$ & $8.34 \times 10^{-3}$ & $6.42 \times 10^{-3}$  &  $2.16 \times 10^{-4}$  & $2.99 \times 10^{-4}$   & $2.66 \times 10^{-4}$ \\  
		dropout rate 					& 0.22 & 0.29 & 0.01  & 0.07  &  0.02  &  0.01  &  0.01\\   
		\midrule
		optimizer 						   &  Adam  &  Adam&   Adam &   Adam  &    Adam  &  Adam & Adam  \\  
		learning rate  					& $9.70 \times 10^{-4}$ & $4.91\times 10^{-4}$ & $ 9.30\times 10^{-4}$ & $2.09\times 10^{-4}$  &  $ 2.66\times 10^{-4}$  & $5.89\times 10^{-4}$   & $ 3.35\times 10^{-4}$  \\   
		initializer  						 & uni & x\_uni & x\_uni  & norm  &  x\_norm  &  norm  & norm  \\   
		\midrule
		batch size   				      & 1024 & 1024 & 2048 &  1024  &  512  & 1024   & 1024 \\  
		dimension size   			  & 2000 & 1000 & 500  & 500  &  1000  & 2000   & 500 \\  
		inverse relation    		  &  False & False &   False&  False&  False&  False&  False \\     
		\bottomrule
	\end{tabular}
\end{table}

\clearpage

\begin{table}[ht]
	\centering
	\caption{Searched optimal hyperparameters for the ogbl-biokg dataset.}
	\label{tab:searched_HP_biokg}
	\small
	\vspace{-5px}
	\renewcommand{\arraystretch}{1.2}
	\begin{tabular}{c|c|c|c|c|c }
		\toprule
		HP/Model  & ComplEx  &  DistMult  &  TransE &  RotatE  &  AutoSF    \\ \midrule
		\# negative samples 	& 512  & 512  & 128  & 128  &  512 \\   \midrule
		loss function 					& CE  &  CE &  CE  & BCE\_adv & CE \\   
		gamma 							  &  12.90 &  11.82 &  7.60 & 18.34 & 12.90 \\   
		adv. weight 					& 0.00  & 0.00  &  0.00 & 1.94 &  0.00 \\   \midrule
		regularizer 		  			 &  NUC & NUC  &  NUC & DURA & NUC \\   
		reg. weight 					 & $1.38 \times 10^{-3}$  & $1.20 \times 10^{-6}$  & $6.99 \times 10^{-3}$  & $1.09 \times 10^{-6}$ & $ 1.38 \times 10^{-4}$ \\   
		dropout rate 					&  0.01 & 0.00  & 0.00  & 0.00 & 0.01  \\   \midrule
		optimizer 						   & Adam  & Adam & Adam  & Adam  & Adam \\   
		learning rate  					& $1.89 \times 10^{-3}$  & $1.25 \times 10^{-3}$  & $1.24 \times 10^{-4}$  & $1.11 \times 10^{-4}$ & $1.89 \times 10^{-3}$ \\    
		initializer  						 & uni  & x\_uni  & x\_uni  & norm & uni \\   
		batch size   				      & 1024  & 1024  & 1024  & 1024  &  1024 \\   
		dimension size   			  &  2000 & 1000  &  2000 & 2000  &   2000 \\   
		inverse relation    		  & False  & False  & False  & False  & False \\   
		\bottomrule
	\end{tabular}
\end{table}

\vspace{40px}

\begin{table}[ht]
	\centering
	\caption{Searched optimal hyperparameters for the ogbl-wikikg2 dataset}
	\label{tab:searched_HP_wikikg2}
	\small
	\vspace{-5px}
	\renewcommand{\arraystretch}{1.2}
	\begin{tabular}{c|c|c|c|c|c}
		\toprule
		HP/Model  & ComplEx  &  DistMult  &  TransE &  RotatE  & AutoSF   \\ \midrule
		\# negative samples 	& 32  & 32  & 128  & 32  &  2048 \\   \midrule
		loss function 					& CE  & CE  & CE  & CE & CE \\   
		gamma 							  & 6.00  & 6.00  & 21.05  & 23.94 & 18.91 \\   
		adv. weight 					&  0.00 & 0.00  &  0.00 & 0.00 & 0.00 \\   \midrule
		regularizer 		  			 & DURA  & DURA  &  FRO & DURA & DURA \\   
		reg. weight 					 & $9.58 \times 10^{-7}$   & $ 1.98 \times 10^{-4}$  &  $1.56 \times 10^{-5}$ & $ 8.10\times 10^{-3}$ & $ 1.38 \times 10^{-4}$  \\   
		dropout rate 					&  0.00 &  0.00 & 0.01  & 0.07 & 0.07 \\   \midrule
		optimizer 						   & Adam  & Adam & Adam  & Adam & Adam \\   
		learning rate  					& $1.34 \times 10^{-4}$   & $ 1.98 \times 10^{-4}$  &  $ 6.05 \times 10^{-4}$ & $ 4.07 \times 10^{-2}$ & $ 1.04 \times 10^{-2}$ \\   
		initializer  						 & x\_norm  & x\_norm  & x\_norm  & x\_norm & x\_norm  \\   
		batch size   				      & 1024  &  1024 & 1024  & 1024  &  1024 \\   
		dimension size   			  &  100  & 100 & 100  & 100 & 100 \\   
		inverse relation    		  & False  & False  & False  & False  & False \\   
		\bottomrule
	\end{tabular}
	\vspace{-5px}
\end{table}

\clearpage
\subsection{Results on general benchmarks}
\label{app:general-benchmark}

We compare the types of results on WN18RR and FB15k-237 in Table~\ref{tab:perf_wn18rr_fb15k237}.
In the first part,
we show the results reported in the original papers.
In the second part,
we show the reproduced results in \citep{ruffinelli2019you}.
And in the third part,
we show the results of the HPs searched by KGTuner.

\begin{table*}[ht]
	\centering
	\caption{Performance on WN18RR and FB15k-237 dataset. 
		The \textbf{bold numbers} mean the best performances of the same model, 
		and the \underline{underlines} mean the second best.}
	\label{tab:perf_wn18rr_fb15k237}
	\vspace{-6px}
	\renewcommand{\arraystretch}{1.2}
	\small
	\begin{tabular}{cc|cccc|cccc}
		\toprule
		&          & \multicolumn{4}{c}{WN18RR}     & \multicolumn{4}{c}{FB15k-237}  \\
		&          & MRR   & Hit@1 & Hit@3 & Hit@10 & MRR   & Hit@1 & Hit@3 & Hit@10 \\ \midrule
		\multirow{6}{*}{Original} 
		& ComplEx  & 0.440 &  0.410     &   0.460     &  0.510      & 0.247 &  0.158     &  0.275     &  0.428      \\
		& DistMult & 0.430 &   0.390  &   0.440    &   0.490     & 0.241 &   0.155    &  0.263       &  0.419      \\
		& RESCAL   & 0.420 &    -  &   -  & 0.447       & 0.270 &   -    &   -    &   0.427     \\
		& ConvE    & 0.430 &  0.400  & 0.440    &  \textbf{0.520}  & 0.325 &    \underline{0.237} &   0.356     &   0.501     \\
		& TransE   & 0.226 &  -     &  -     &   0.501     & 0.294 &    -   &   -    &  0.465      \\
		& RotatE   & \underline{0.476} &  \textbf{0.428}    &  \underline{0.492}     &   \underline{0.571}     &  \textbf{0.338}  &   \underline{0.241}    &   \textbf{0.375}    &   \textbf{0.533}     \\ 
		& TuckER   & \underline{0.470} &  \textbf{0.443}    &  \underline{0.482}    &   \underline{0.526}     & \textbf{0.358}  &   \textbf{0.266}    &   \textbf{0.394}    &   \textbf{0.544}    \\ \midrule

		\multirow{6}{*}{\begin{tabular}[c]{@{}l@{}} LibKGE \\  \citep{ruffinelli2019you}\end{tabular}}
		& ComplEx  & \underline{0.475}  &  \underline{0.438}     &    \underline{0.490}     &   \underline{0.547}     &  \underline{0.348} &   \underline{0.253}     &   \underline{0.384}     &  \textbf{0.536}      \\
		& DistMult &  \underline{0.452} &   \textbf{0.413}  &    \underline{0.466}    &   \underline{0.530}     &  \underline{0.343} &    \underline{0.250}   &  \textbf{0.378}       &  \textbf{0.531}      \\
		& RESCAL   &  \underline{0.467} &    \textbf{0.439}  &    \underline{0.480}  &   \underline{0.517}       &   \underline{0.356}  &    \underline{0.263}    &    \textbf{0.393}    &   \textbf{0.541}     \\
		& ConvE    & \textbf{0.442} &   \textbf{0.411}  & \textbf{0.451}    &  \underline{0.504}  & \textbf{0.339}  &   \textbf{0.248}  &   \textbf{0.369}     &   \underline{0.521}     \\
		& TransE   & \underline{0.228} &  \textbf{0.053}     &  \underline{0.368}    &   \underline{0.520}     & \underline{0.313} &    \underline{0.221}   &   \underline{0.347}    &  \underline{0.497}      \\ \midrule

		\multirow{6}{*}{KGTuner (ours)} 
		& ComplEx  & \textbf{0.484} &  \textbf{0.440}     &  \textbf{0.506}     &    \textbf{0.562}    &   \textbf{0.352}   &  \textbf{0.263}     &  \textbf{0.387}    &   \underline{0.530}      \\
		& DistMult &  \textbf{0.453}    &   \underline{0.407}     &   \textbf{0.468}    &  \textbf{0.548}   &   \textbf{0.345}    &  \textbf{0.254}   &  \underline{0.377}     &   \underline{0.527}       \\
		& RESCAL   &  \textbf{0.479}  &   \underline{0.436}    &    \textbf{0.496}    &     \textbf{0.557}   &  \textbf{0.357}  &  \textbf{0.268}    &  \underline{0.390}     &   \underline{0.535}     \\
		& ConvE    &  \underline{0.437}   &   \underline{0.399}     &    \underline{0.449}   &     0.515   &  \underline{0.335}  &    \underline{0.242}   &   \underline{0.368}     &    \textbf{0.523}    \\
		& TransE   & \textbf{0.233}  &  \underline{0.032}     &    \textbf{0.399}    &      \textbf{0.542}   &   \textbf{0.327}   &     \textbf{0.228}   &    \textbf{0.369}    &   \textbf{0.522}      \\
		& RotatE   &  \textbf{0.480} &  \underline{0.427}     &   \textbf{0.501}     &    \textbf{0.582}     &   \textbf{0.338}  &   \textbf{0.243}      &   \underline{0.373}     &    \underline{0.527}     \\ 
		&TuckER & \textbf{0.480} & \underline{0.437} &  \textbf{0.500} &   \textbf{0.557}  &  \underline{0.347}   &  \underline{0.255}    &   \underline{0.382}    &    \underline{0.534}    \\  \bottomrule
		
	\end{tabular}
\end{table*}

\subsection{Full results for OGB}
\label{app:ogb}


\begin{table}[ht]
	\centering
	\caption{Full results on ogbl-biokg and ogbl-wikikg2 dataset.}
	\label{tab:perf_ogb_full}
	\setlength\tabcolsep{3pt}
	\renewcommand{\arraystretch}{1.3}
	\small
	\vspace{-6px}
	\begin{tabular}{cc|ccc|ccc}
		\toprule
		&      & \multicolumn{3}{c|}{ogbl-biokg}                        & \multicolumn{3}{c}{ogbl-wikikg2} \\
		&  & Test MRR & Val MRR & {\#parameters} & Test MRR & Val MRR & \#parameters \\ \midrule
	 	& {ComplEx}  & 0.8095$\pm$0.0007   &  0.8105$\pm$0.0001  & {187,648,000}  &     0.4027$\pm$0.0027	   &    0.3759$\pm$0.0016     &  1,250,569,500  \\
		OGB & {DistMult} &  0.8043$\pm$0.0003  & 0.8055$\pm$0.0003   & {187,648,000}  &    0.3729$\pm$0.0045	      &     0.3506$\pm$0.0042    & 1,250,569,500  \\
		board & {RotatE}   &   0.7989$\pm$0.0004       &   0.7997$\pm$0.0002      & {187,597,000}  &     0.2530$\pm$0.0034     &     0.2250$\pm$0.0035     & 250,087,150   \\
		& {TransE}   &     0.7452$\pm$0.0004     &    0.7456$\pm$0.0003     & {187,648,000}  &    0.4256$\pm$0.0030     &   0.4272$\pm$0.0030      &  1,250,569,500  \\ 
		& {AutoSF}   &     0.8309$\pm$0.0008     &    0.8317$\pm$0.0007     &  187,648,000  &    0.5186$\pm$0.0065     &   0.5239$\pm$0.0074      &  250,113,900  \\ 
		\midrule
		\multirow{5}{*}{KGTuner}                                                & {ComplEx}  &   0.8385$\pm$0.0009    &   0.8394$\pm$0.0007    & {187,648,000}  &   0.4942$\pm$0.0017    &    0.5099$\pm$0.0023   & 250,113,900   \\
		& {DistMult} &   0.8241$\pm$0.0008       &    0.8245$\pm$0.0009     &   {93,824,000}  &    0.4837$\pm$0.0078      &    0.5004$\pm$0.0075     & 250,113,900  \\
		& {RotatE}   &   0.8013$\pm$0.0015       &    0.8024$\pm$0.0012     & {187,597,000}  &  0.2948$\pm$0.0026  &   0.2650$\pm$0.0034     & 250,087,150     \\
		& {TransE}   &  0.7781$\pm$0.0009       &   0.7787$\pm$0.0008    &   {187,648,000}  &     0.4739$\pm$0.0021    &   0.4932$\pm$0.0013   & 250,113,900    \\ 
		& {AutoSF}   &     0.8354$\pm$0.0013     &    0.8361$\pm$0.0012     &  187,648,000 &    0.5222$\pm$0.0021     &   0.5397$\pm$0.0023      &  250,113,900  \\ 
		\bottomrule
	\end{tabular}
\end{table}

\end{document}